\pdfoutput=1                              %
\RequirePackage{amsmath}                  %
\documentclass{fairmeta}
\usepackage{amssymb,amsthm,wrapfig,float}
\usepackage{lmodern}                      %
\usepackage[export]{adjustbox}            %
\let\fullincludegraphics\includegraphics  %
\renewcommand{\includegraphics}[2][]{\fullincludegraphics[#1,max height=0.82\textheight]{#2}}

\newtheorem{theorem}{Theorem}
\numberwithin{equation}{section}
\newcommand{\TPP}{\textup{TPP}}
\newcommand{\Lstar}{L_*}
\newcommand{\E}{\mathbb{E}}
\newcommand{\w}{\mathbf{w}}
\newcommand{\g}{\mathbf{g}}
\makeatletter                             %
\g@addto@macro\appendix{\crefalias{section}{appendix}\crefalias{subsection}{appendix}}
\makeatother
\graphicspath{{./}}

\title{Universal scaling law}

\author[1]{Zhiqi Bu, Junlin Han, Zeyuan Allen-Zhu, Ashok Cutkosky, Aaron Defazio, Mike Lewis, Lin Xiao}                      %
\affiliation[1]{FAIR, Meta Superintelligence Labs}             %

\abstract{Scaling laws are the bedrock for foundation model training, yet standard scaling laws are largely restricted to evaluating cross-entropy loss at terminal checkpoints -- leaving the continuous dynamics of general observables under-explored. In this work, we establish universal scaling laws across large language models (LLM) and multi-modal language models (MMLM) that accurately characterize the any-time-any-scale trajectories of losses, evaluation metrics, and parameter norms. To establish this unified framework, we integrate three key pillars: conventional end-point scaling laws, convex optimization theory, and an enhanced collapse transformation. We systematically validate this framework across diverse model architectures, optimizers, training hyperparameters, compute regimes, and modalities, providing a principled recipe for universal scaling.
}

\correspondence{\email{zhiqibu@meta.com}} %

\begin{document}

\maketitle

\vspace{-0.1cm}
\section{Introduction}
\vspace{-0.1cm}
Scaling laws are central to foundation model training, precisely characterizing the loss by simple functions
\citep{kaplan2020scaling,hoffmann2022training} across wide ranges of model size, dataset size, and compute.
These laws enable extrapolation from small-scale experiments to expensive large-scale ones, and provide quantitative guidance for allocating compute across model and data scales. %

A conventional scaling law, however, provides only a compressed description of training: it predicts a single quantity, the loss value at a single time, the terminal checkpoint.
A broad line of work has therefore sought to enrich scaling laws along two complementary axes.
The first broadens the prediction from the final loss alone to the entire training trajectory, though so far largely restricted to the loss itself.
The second broadens the predicted observables: from pretraining loss to downstream evaluation and emergent capabilities.

In this work, we advance both axes jointly by characterizing the full training trajectory for generalized observables, including evaluation metrics and parameter norms. We show that end-point scaling laws and scaling collapse of loss curves are complementary rather than competing:
a scaling law determines the absolute loss scale, while scaling collapse gives how loss evolves relative to that scale.
Their combination therefore converts an end-point prediction into an absolute prediction of the full training trajectory:
\begin{equation}
    \text{scaling law}
    +
    \text{scaling collapse}
    \quad\Longrightarrow\quad
    \text{universal scaling law}.
\end{equation}
We further show convex optimization can be used as a tractable model of training dynamics, not only characterizing the trajectory in closed form, but also identifying the scaling of hyperparameters for trajectories at different training scales to collapse.
Together, these connections link scaling laws, scaling collapse, and convex optimization theory into a unified description of training dynamics.

\vspace{-0.1cm}
\subsection{Scaling Laws --- absolute prediction of end-points}
\label{sec:scaling-law}
\vspace{-0.1cm}
The conventional scaling laws \citep{kaplan2020scaling,henighan2020scaling,hoffmann2022training,achiam2023gpt,dey2023cerebras,grattafiori2024llama,bi2024deepseek} characterize the last-iteration losses: with $N$ denoting the model size, $T$ the total number of training iterations, and $B$ the batch size (tokens per iteration), the scaling law under fixed token-per-parameter ratio ($\TPP$) has a common form as
\begin{equation}
    L_{T,N}(T)-L_{*}=\alpha\cdot C(N,T)^{\beta}
    =\alpha\cdot(6BNT)^{\beta}
\label{eq:scaling-law}
\end{equation}
Here $L_{T,N}(t)$ measures the intermediate losses after consuming $t/T$ of training budget for a model of size $N$, $C(N,T)=6BNT$ is the training compute, $L_*$ is the irreducible loss, and $\alpha,\beta$ are the coefficients to be determined by multiple small-scale runs.

Scaling laws have been generally observed across data qualities and mixtures \citep{ye2025data,shukor2026scaling}, model architectures \citep{ludziejewski2024scaling,wang2024scaling,liang2026scaling}, and optimizer choices \citep{liu2025muon,volkova2026towards,wen2026fantastic}.%

Furthermore, scaling laws extend beyond the cross-entropy loss to downstream and discrete evaluation metrics, including the log-scale pass rate \citep{achiam2023gpt,khatri2026art} and the log-scale task accuracy \citep{gadre2025language,grattafiori2024llama,krajewski2026revisiting}.
These extensions significantly enhance the capability of scaling laws, allowing robust and broader prediction of model performance at large scale.

However, these scaling laws are limited to the terminal loss at $t=T$ and do not characterize intermediate $L_{T,N}(t)$ at $t<T$ for unseen model scale $N$.
We establish that the scaling-law structure itself extends from terminal checkpoints to the entire training trajectory: for a fixed-TPP scale family,
$$L_{T,N}(xT)\approx(1-\Phi(x))\,L_\text{offset}+\Phi(x)\bigl(L_*+\alpha\cdot C(N,T)^\beta\bigr),$$
where $\Phi(x)$ is the universal trajectory to be introduced in the next subsection and $L_\text{offset}$ is the observable offset of the scaling collapse \eqref{eq:new collapse}. Thus, the end-point law is the $x=1$ slice of our universal scaling law, and every fixed-$x$ slice yields an intermediate-point scaling law with the same compute exponent $\beta$. This shared structure also allows all checkpoints (not only the terminal ones) across scale and training progress to jointly constrain the scaling coefficients.

\vspace{-0.1cm}
\subsection{Scaling collapse --- relative prediction of entire trajectory}
\label{sec:collapse}
\vspace{-0.1cm}
The scaling collapse phenomenon \citep{qiu2025scaling,bergsma2025scaling} shows that, under fixed $\TPP=BT/N$ and appropriate training configurations, the loss curves of different model sizes and training budgets can collapse onto a universal trajectory. To be specific, one can obtain the relative
trajectory $\Phi:[0,1]\to [1,\infty)$ by normalizing the training progress and the loss value, which is approximately scale-invariant:
\begin{align}
\frac{L_{T,N}(xT; \eta^*)-\Lstar}{L_{T,N}(T; \eta^*)-\Lstar}\approx \Phi(x;\TPP)
\label{eq:collapse_old}
\end{align}
Here $x:=t/T$ is the normalized training progress, $\eta^*=\text{argmin}_\eta L_{T,N}(T; \eta)$ is the optimal learning rate, and $L_{T,N}(t; \eta)$ is the loss obtained after $t$ out of $T$ iterations with:
\begin{itemize}
    \item the peak learning rate is equal to $\eta$ and learning rate schedule shape is fixed;
    \item the weight decay is dependent on learning rate via $\lambda\propto\frac{1}{\eta T}$, i.e. fixing the optimization timescale $\frac{1}{\lambda\eta T}$ as suggested by \citep{wang2025set}\footnote{\cite{qiu2025scaling} fixes $\lambda=0$ as a special case by not using the weight decay.}.
\end{itemize}

However, prior scaling collapse results leave $\Phi(x)$ as an unparameterized empirical curve and lack a principled recipe for preserving scaling collapse across scales. We address these limitations from the convex-optimization perspective developed in the next subsection, which motivates $1/\sqrt{T}$-scaling of the learning rate and weight decay to preserve scaling collapse, together with an offset $T_\text{offset}$ and a schedule-dependent parameterization of $\Phi(x)$.

\vspace{-0.1cm}
\subsection{Convex dominance --- an optimization principle of entire trajectory}
\label{sec:optimization-theory}
\vspace{-0.1cm}
Convex dominance \citep{buconvex,li2026towards} shows that convex optimization theory by \citep{defazio2023optimal} can accurately characterize the training dynamics and loss curves for deep learning (cf. Figure 2-5 in \citep{buconvex}, Figure 2 in \citep{li2026towards}, and \cite{schaipp2025surprising}), indicating that deep learning is convex-like although the optimization of neural networks is highly non-convex and non-smooth.

Concretely, for any learning rate sequence and optimizers (e.g. AdamW and Muon) following
\begin{align}
w_{t+1}=(1-\lambda\eta_t)w_t-\eta_{t} u_t,
\label{eq:sgd}
\end{align}
the any-iteration loss can be characterized as
\begin{align}
L_{T,N}(t;\eta)-L_{N}\approx\frac{D_N^2}{2\sum_{\tau=1}^t \eta_\tau}+ \frac{1}{2}\sum_{\tau=1}^{t-1}\frac{\eta_\tau^2 G_N^2}{\sum_{k=\tau+1}^t \eta_k}+\frac{1}{2}\eta_t G_N^2
\label{eq:loss seq2seq}
\end{align}
where $(L_N,D_N,G_N)$ are coefficients that depend on model size. Moreover, by optimizing the last-iteration loss in \eqref{eq:loss seq2seq}, the asymptotically optimal convergence can be achieved when
\begin{itemize}
    \item The peak learning rate is $1/\sqrt{T}$-scaled: $\eta(T)=\frac{\eta_\textit{ref}}{\sqrt{T}}$ where $\eta_\textit{ref}$ is tuned at small scale.
    \item The learning rate schedule shape is qualified, which includes cosine, warmup-stable-decay, and linear schedules, but not constant learning rate (cf. Condition 2.5 of \citep{buconvex}).
    \item The weight decay $\lambda$ is $1/\sqrt{T}$-scaled: $\lambda(T)=\frac{\lambda_\textit{ref}}{\sqrt{T}}$ where $\lambda_\textit{ref}$ is tuned at small scale.
\end{itemize}
The optimality of $1/\sqrt{T}$-scaled hyperparameters is theoretically derived in \citep{li2026towards} and empirically verified in \citep{microsoftai2026frontier} up to 2.8T-parameter MoE.
We leverage this convex perspective as a principle for scaling collapse: $1/\sqrt{T}$-scaling keeps the optimization timescale $1/(\lambda\eta T)$ invariant and empirically preserves scaling collapse across scales. Moreover, convex analysis yields an explicit parameterized form for $\Phi(x)$ in \Cref{eq:closed form}, and further motivates a training-progress offset $T_\text{offset}$.

\vspace{-0.1cm}
\subsection{Contributions}
\label{sec:contributions}
\vspace{-0.1cm}

\paragraph{Universal scaling of entire loss trajectory.}
We combine the end-point scaling law and universal trajectory into an any-time, any-scale law, indicating that all intermediate-point scaling laws share the same compute exponent. This structure enables joint estimation from all checkpoints for robust trajectory forecasting at unseen scales.

\paragraph{Universal scaling beyond loss.}
We generalize universal scaling to evaluation metrics -- such as continuous LAMBADA perplexity and discrete HellaSwag accuracy -- and parameter norms at different granularities.

\paragraph{Convexity-informed scaling collapse.}
We advance scaling collapse through the lens of convex dominance. Specifically, we employ $1/\sqrt{T}$-scaling of hyperparameters to preserve scaling collapse, introduce a new transformation with offset to sharpen the scaling collapse, and derive a closed-form parameterization for the universal trajectory.

\paragraph{Extensive experiments across settings and scales.}
We validate universal scaling on a variety of optimizers (AdamW and Muon), model architectures (LLaMA, Qwen3-MoE, Transfusion), TPP notions, and modalities (text and image).
We train up to 7B MoE and 14B MMLM with trillions of tokens, and evaluate over 20 vision and language metrics.

\vspace{-0.1cm}
\section{Convexity-informed scaling collapse}
\vspace{-0.1cm}
We connect the convex optimization to the scaling collapse phenomenon, transferring explicit hyperparameter scaling and loss characterization to the deep learning regime. This connection allows scaling collapse to be more principled: it provides guidance on how to induce scaling collapse, improve the quality of scaling collapse, and explicitly characterize the resulting trajectory.

\vspace{-0.1cm}
\subsection{Scaled hyperparameters preserve scaling collapse}
\vspace{-0.1cm}
Prior work establishes scaling collapse under a fixed AdamW timescale~\citep{qiu2025scaling,bergsma2025scaling}, and does not uniquely determine how the hyperparameters should scale to preserve such timescale. We extend this phenomenon to general optimizers, including Muon, and specifically derive the $1/\sqrt{T}$-scaling for hyperparameters, which attains the optimal asymptotic convergence rate in convex optimization analysis. We show this in \Cref{thm:scaling hyperparam}, with a proof and an alternative derivation via Robbins-Monro conditions  in \Cref{app:theory}.
\begin{theorem}
\label{thm:scaling hyperparam}
For convex loss $L_{T,N}$, bounded gradient $\|g_t\|\leq G_N$ and bounded parameters $\|w_t\|\leq W$, the stochastic gradient descent $w_{t+1}= (1-\lambda\eta_t)w_t-\eta_t g_t$ satisfies
$$\E L_{T,N}(T)-L_N=O(1/\sqrt{T})$$
under $\lambda=\lambda_\text{ref}/\sqrt{T}, \eta_t=s(t/T)\cdot \eta_\text{ref}/\sqrt{T}$ for any choice of $(\lambda_\text{ref},\eta_\text{ref})$ and any qualified learning rate schedule $s:[0,1]\to [0,1]$ (cf. condition 2.5 of \cite{buconvex}).
\end{theorem}

\begin{figure}[!htb]
    \centering
    \includegraphics[width=\linewidth]{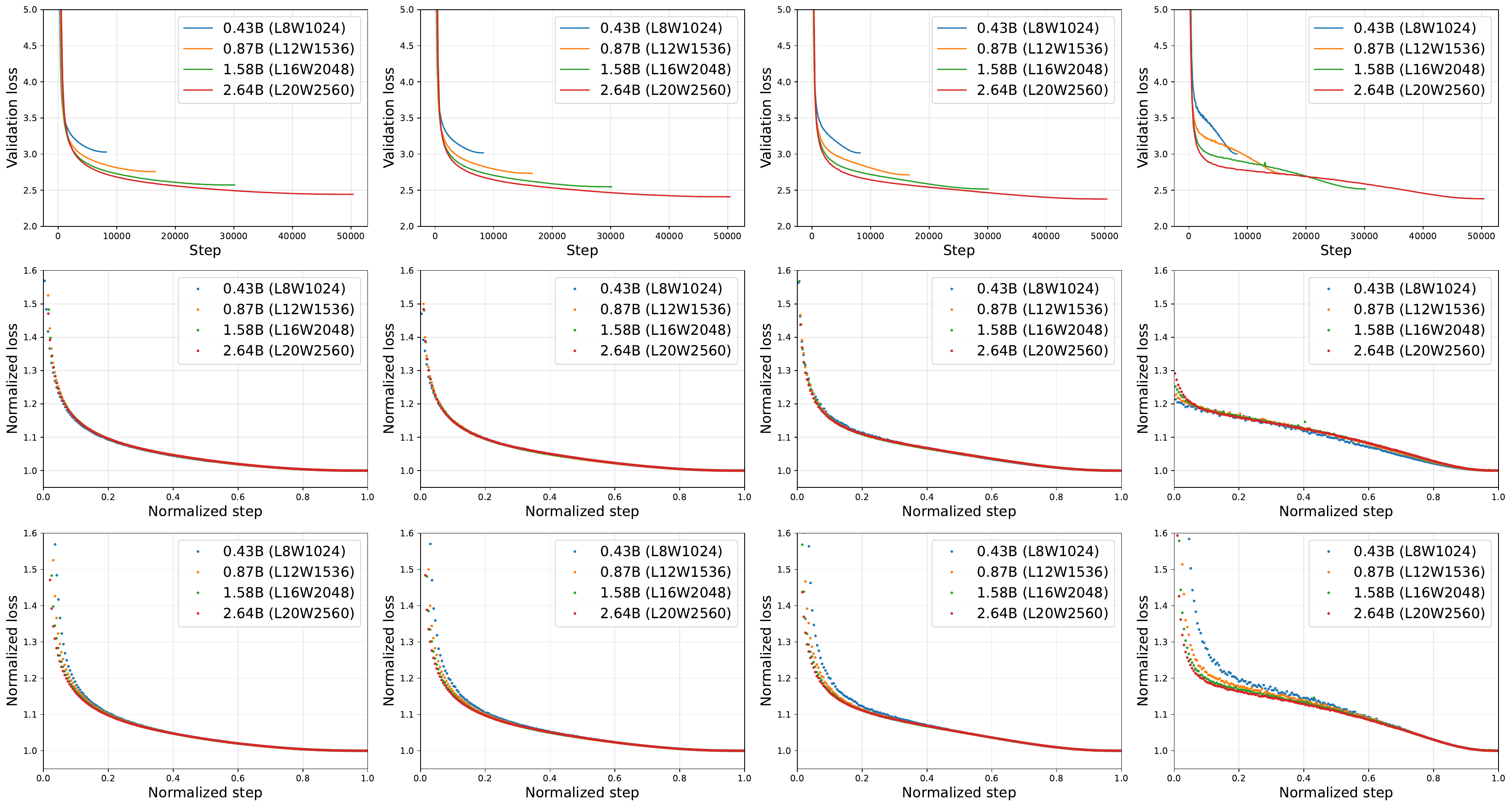}
    \vspace{-0.6cm}
    \caption{Validation loss for varying model sizes, raw (top row) and normalized (middle row with fitted $T_\text{offset}$; bottom row with $T_\text{offset}=0$). We train Qwen3-MoE models with Muon-NSGD optimizer under $\TPP=20$. Each column is a different $(\lambda_\text{ref},\eta_{\rm ref})$ pair: $\eta_\text{ref}=\lambda_\text{ref}=\{0.5,1,2,5\}$ from left to right. The value of \eqref{eq:sequential-collapse} is $\{4.2,1.6,0.31,0.82\}\times10^{-5}$ at the fitted $T_\text{offset}=\{266,266,266,1399\}$ and $\{113,82,68,91\}\times10^{-5}$ at $T_\text{offset}=0$.
    }
    \label{fig:sqrtT scaling collapse}
\end{figure}

We highlight that $1/\sqrt{T}$-scaling preserves scaling collapse for \textbf{any} $(\lambda_\text{ref},\eta_\text{ref})$, including the optimal and non-optimal choices. This can be observed by comparing first and second row in \Cref{fig:sqrtT scaling collapse}, where the optimal $\eta_\text{ref}=\lambda_\text{ref}=2$.

\vspace{-0.1cm}
\subsection{Training-progress offset sharpens scaling collapse}
\vspace{-0.1cm}
Convex dominance reveals that convex-like behavior emerges only after an initial burn-in period~\citep{buconvex,li2026towards}. This motivates introducing a training-progress offset $T_\text{offset}$ into the scaling collapse transformation:
\begin{align}
    t(x;T_\text{offset})&=T_\text{offset}+x(T-T_\text{offset}),
    \label{eq:progress}
\end{align}
we obtain a new transformation that sharpens scaling collapse in \Cref{fig:sqrtT scaling collapse}:
\begin{equation}
    \frac{
    L\!\left(t(x);
    \eta \right)-L_\text{offset}
    }{
    L\!\left(T;\eta \right)-L_\text{offset}
    } := F(x; T_\text{offset}, L_\text{offset})
    \approx \Phi(x; \TPP),
    \label{eq:new collapse}
\end{equation}
where $L_\text{offset}$ is the loss offset.

To estimate $(T_\text{offset},L_\text{offset})$ in \eqref{eq:new collapse}, we minimize the discrepancy among the transformed trajectories so that all trajectories are as close as possible to one unknown $\Phi(x)$: denoting $F_i(x;T_\text{offset},L_\text{offset})$ as the transformed loss curve of run $i\in\mathcal I$, we solve \eqref{eq:find offset}:
\begin{equation} \min_{T_\text{offset},L_\text{offset}} \sum_{i<j} \int_0^1 \left| F_i(x;T_\text{offset},L_\text{offset})-F_j(x;T_\text{offset},L_\text{offset}) \right|^2 \,dx.
\label{eq:find offset}
\end{equation}
Equivalently, let $\bar\Phi(x)=\frac{1}{|\mathcal I|}\sum_{i\in\mathcal I}F_i(x;T_\text{offset},L_\text{offset})$ be the pointwise mean of the normalized curves \eqref{eq:new collapse}. Since $\bar\Phi$ solves the functional optimization of $\Phi$ for fixed offsets, we profile out $\Phi$ and estimate the offsets by
\begin{equation}
    \min_{T_\text{offset},L_\text{offset}}
    \frac{1}{|\mathcal I|}\sum_{i\in\mathcal I}\int_0^1
    \bigl[F_i(x;T_\text{offset},L_\text{offset})-\bar\Phi(x)\bigr]^2\,dx.
    \label{eq:sequential-collapse}
\end{equation}

\vspace{-0.1cm}
\subsection{Closed-form structure of universal trajectory}
\begin{figure}[!htb]
    \centering
    \includegraphics[width=\linewidth]{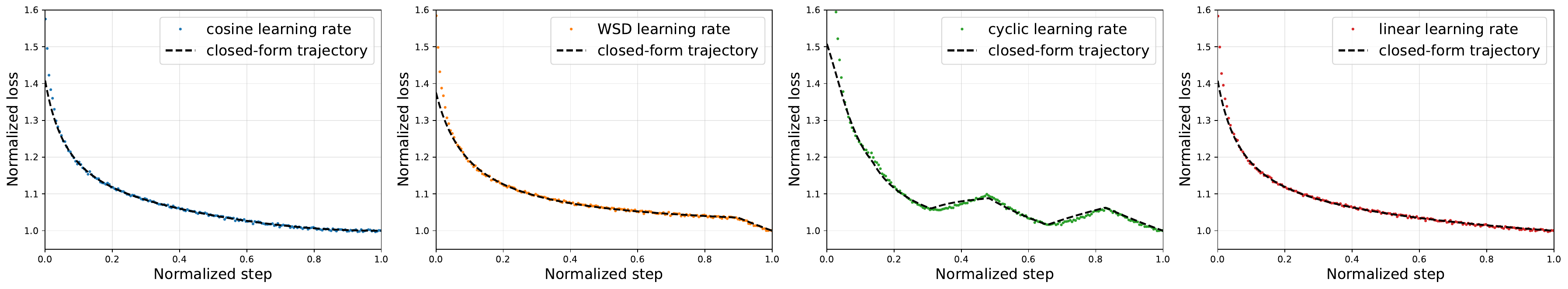}
    \vspace{-0.6cm}
    \caption{Empirical normalized loss curves (colored dots) compared to analytically derived closed-form trajectories (dashed black lines) for different learning rate schedules.
    }
    \label{fig:lr schedules}
\end{figure}

\vspace{-0.1cm}
Scaling collapse establishes that normalized loss trajectories converge to a universal trajectory $\Phi(x)$, but leaves its functional form unspecified\footnote{Eq. (21) of \citep{qiu2025scaling} derives a schedule-transfer relation for the universal trajectory that predicts a target trajectory from a reference trajectory, together with an empirically measured function $h(x)\approx \operatorname{Tr}(\widetilde{\Sigma}(xT))/L(xT)$, where $\widetilde{\Sigma}$ is the preconditioned gradient-noise covariance. Thus, their formulation is not fully explicit or readily applicable in practice.}. We now leverage the convex loss characterization in \eqref{eq:loss seq2seq} to unravel the structure of this universal trajectory: as derived in \Cref{app:closed form derivation}, we have
\begin{align*}
L_{T,N}(xT;\eta)
\approx
L_N
+\frac{D_N^2}{2\eta T \int_0^{x} s(u)\,du}
+\frac{\eta G_N^2}{2}s(x)\log N,
\end{align*}
which informs a closed-form parameterization as visualized in \Cref{fig:lr schedules}:
\begin{align}
\Phi(x)
\approx
\frac{
A+\frac{1}{\int_0^{x}s(u)\,du+Q}+R\,s(x)
}{
A+\frac{1}{\int_0^{1}s(u)\,du+Q}+R\,s(1)
}.
\label{eq:closed form}
\end{align}
Here $A$, $R$, and $Q$ are fitted coefficients and $s(x)$ is the normalized learning rate schedule, e.g. linear schedule is $s(x)=1-x$.
Importantly, scaling collapse is refined from an empirical statement that a universal trajectory exists to an explicit characterization of the trajectory, whose shape strongly depends on the shape of learning rate schedule.

\vspace{-0.1cm}
\section{Universal scaling of loss trajectory}
\vspace{-0.1cm}
We connect scaling laws with the scaling collapse phenomenon, extending scaling from end-point performance to the entire training trajectory. Remarkably, the two are perfectly complementary and indispensable: scaling collapse reveals a universal but normalized trajectory, providing the relative shape; the end-point scaling law predicts the terminal loss, providing the absolute magnitude. As a result, we propose the universal scaling law, to upgrade point-wise scaling to trajectory-wise scaling that enables \textbf{any-time, any-scale} loss prediction:
\begin{align}
    L_{T,N}(xT+(1-x)T_\text{offset})
&=\bigl(1-\Phi(x)\bigr)\,L_\text{offset}+\Phi(x)\bigl(L_*+\alpha\cdot C(N,T)^{\beta}\bigr)
\label{thm:frac scaling law}
\end{align}
where $L_\text{offset}$ is the observable offset of the scaling collapse \eqref{eq:new collapse}. The law is thus an affine interpolation between $L_\text{offset}$ and the end-point law \eqref{eq:scaling-law}, weighted by the universal trajectory $\Phi(x)$.

\vspace{-0.1cm}
\subsection{Scaling law at every intermediate point}
\begin{figure}[!htb]
    \centering
    \includegraphics[width=\linewidth]{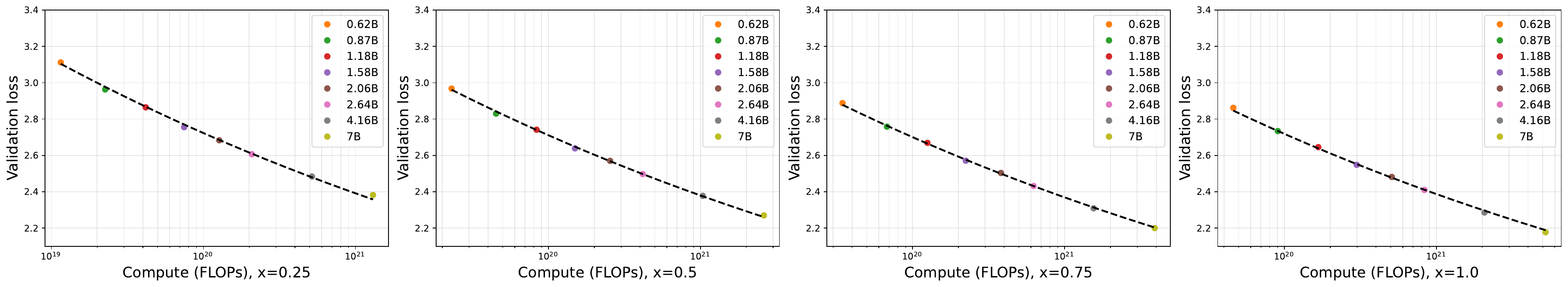}
    \vspace{-0.6cm}
    \caption{Intermediate-point scaling laws have the common formula $L_{T,N}(xT)\approx [(1-\Phi(x))\,L_\text{offset}+\Phi(x)L_*]+\alpha\Phi(x)\cdot C(N,T)^{\beta}$.
    We derive $\alpha=116.5,\beta=-0.091$. All fits have $R^2>0.997$.
    }
    \label{fig:intermediate scaling law}
\end{figure}

\vspace{-0.1cm}
Even without knowing the functional form of $\Phi(x)$ nor estimating $(T_\text{offset},L_\text{offset})$, \Cref{thm:frac scaling law} already yields an important consequence. For every fixed $0<x\leq1$, $\alpha\Phi(x)$ and $L_\text{offset}$ are constants with respect to model and data scale. Hence, every intermediate-point scaling law has exactly the same exponent $\beta$; only its amplitude $\alpha\Phi(x)$ and floor $(1-\Phi(x))L_\text{offset}+\Phi(x)L_*$ change with $x$. We visualize these intermediate-point scaling laws in \Cref{fig:intermediate scaling law}. The conventional end-point law is the special case $x=1$.

\vspace{-0.1cm}
\subsection{Joint estimation of scaling coefficients}
\label{sec:sequential-estimation}
\vspace{-0.1cm}
\noindent\begin{minipage}[t]{0.52\linewidth}
\vspace{0pt}
To apply the universal scaling law \eqref{thm:frac scaling law}, we estimate 5~scalars -- $(\alpha,\beta,T_\text{offset},L_\text{offset},L_*)$, and 1~function -- the universal trajectory $\Phi$. We use loss trajectories at different scales to obtain more precise estimates than relying on end-points alone, allowing intermediate checkpoints to jointly determine the coefficients.

\medskip
Let $\mathcal I$ be the set of fitting runs. We estimate the offsets and $\bar\Phi$ by \eqref{eq:sequential-collapse}, and the scaling coefficients by
\end{minipage}\hfill
\begin{minipage}[t]{0.45\linewidth}
    \vspace{-1cm}
    \centering
    \includegraphics[width=\linewidth]{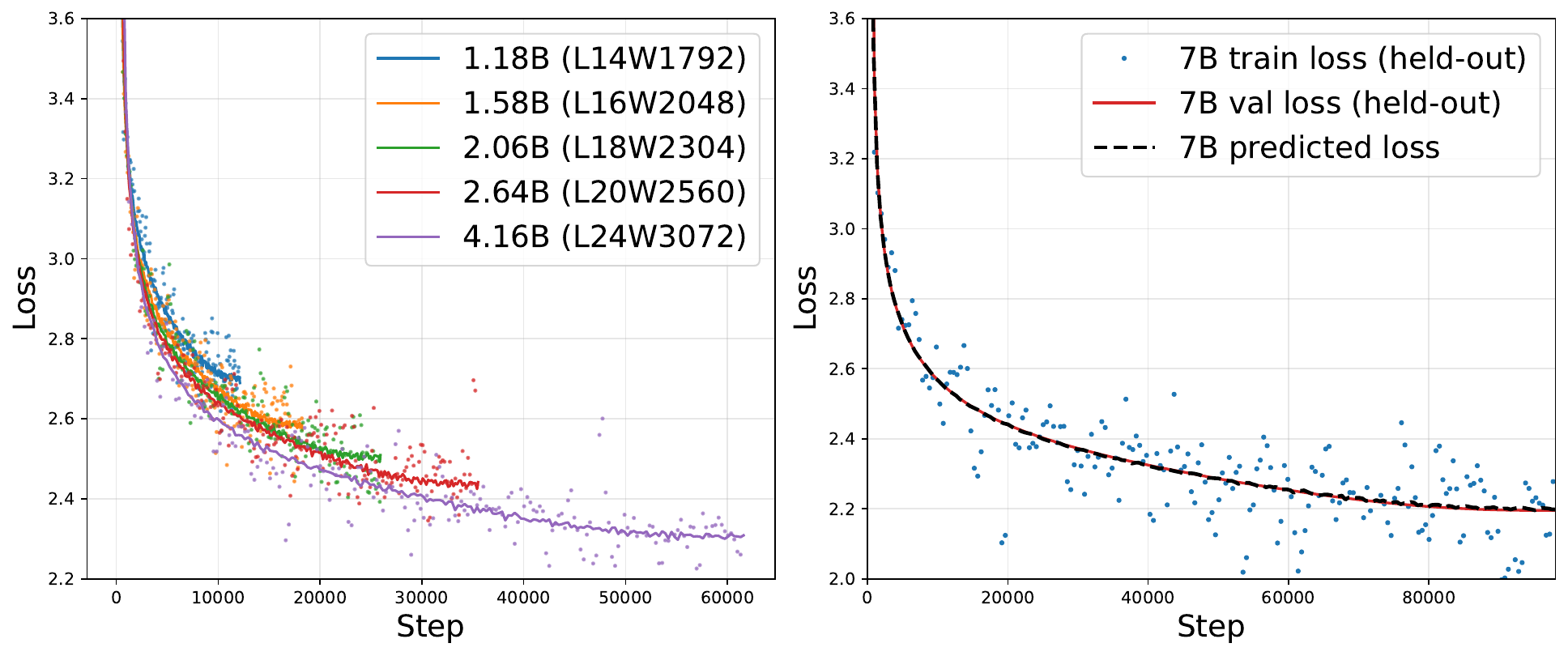}
    \vspace{-0.6cm}
    \captionof{figure}{Estimation of scaling coefficients on smaller runs can forecast the loss trajectory of larger runs. The fitted $T_\text{offset}=372$.
    }
    \label{fig:forecast}
\end{minipage}
\begin{equation}
    \min_{L_*,\alpha,\,\beta}
    \sum_{i\in\mathcal I}\int_0^1
    \bigl[L_{T_i,N_i}(t(x))-(1-\bar\Phi(x))L_\text{offset}-\bar\Phi(x)(L_*+\alpha C_i^\beta)\bigr]^2\,dx,
    \label{eq:sequential-endpoint}
\end{equation}
where $C_i=6BN_iT_i$ and $t(x)$ is the progress coordinate \eqref{eq:progress} with $T=T_i$. With the offsets and trajectory fixed, $L_*$ and $\alpha$ have closed-form least-squares solutions for each $\beta$, leaving a one-dimensional search over $\beta$; \Cref{app:power-law} gives the derivation.

In \Cref{fig:forecast}, our joint estimation on smaller runs from 1.18B to 4.16B models precisely forecasts the entire loss trajectory of the 7B model. The relative difference between validation loss and predicted loss is $<0.5\%$ along the trajectory\footnote{In contrast, the average difference rises from $0.16\%$ to $1.06\%$ for $L_\text{offset}=L_*$ and $T_\text{offset}=0$ \citep{qiu2025scaling}, and to $0.71\%$ for fitted $L_\text{offset}$ and $T_\text{offset}=0$ \citep{bergsma2025scaling}.}.

\vspace{-0.1cm}
\section{Enhanced recipe for universal scaling}
\vspace{-0.1cm}
We now illustrate that universal scaling is robust to modeling and optimization choices. \Cref{tab:collapse} summarizes the scope of our validation relative to prior scaling collapse studies. We focus on TPP notions, optimizers, and model architectures, deferring the complete recommended configuration, including all prescribed choices, to \Cref{app:recipe}.

\begin{table}[!htb]
    \centering
\setlength{\tabcolsep}{2.5pt}
    \caption{Summary of scaling collapse studies; \Cref{app:table1} explains the columns.}
    \vspace{-0.4cm}
\resizebox{\textwidth}{!}{
    \begin{tabular}{c|c|c|c|c|c|c|c}
         &TPP&modality& model arch & optimizer & timescale $\eta\lambda T$ & $L_\text{offset}, T_\text{offset}$ &observable \\\hline
        \cite{qiu2025scaling} &total &text & dense &Adam & 0 & $L_*$, none & loss
\\
        \cite{bergsma2025scaling} &total&text & dense & AdamW & fixed & fitted, none & loss
\\
        This work & general& text/vision  & general & Muon/AdamW & fixed & both fitted & loss/metric/norm
\end{tabular}
}
    \label{tab:collapse}
\end{table}

\vspace{-0.1cm}
\subsection{TPP notions}
\vspace{-0.1cm}
To preserve scaling collapse, we require a fixed TPP and explore its notions. Decomposing model parameters into \{\texttt{embedding}, \texttt{body(active)}, \texttt{body(stale)}, \texttt{head}\}, we define $\TPP_\text{active}$ on \texttt{body(active)}, $\TPP_\text{body}$ on $\texttt{body(active)}+\texttt{body(stale)}$, and $\TPP_\text{total}$ over the entire parameter set. We note that $\TPP_\text{active}$ is proportional to $\TPP_\text{body}$ -- a condition that is always true for dense models and readily preserved in standard MoE scaling suites\footnote{$\TPP_\text{body}/\TPP_\text{active}$ ratio is scale-invariant when expert-intermediate/hidden dimension is fixed, number of experts and top-k routing are fixed, and attention parameters scale as width$^2$. Our setup keeps this ratio $\approx 4$.}; in addition, $\TPP_\text{total}$ is asymptotically equal to $\TPP_\text{body}$ as parameter count scales and non-body parameter overhead becomes negligible.

Consequently, we can preserve scaling collapse using any of these TPP notions, without strictly adhering to $\text{TPP}_\text{total}$ as in prior work \citep{qiu2025scaling,bergsma2025scaling}. In particular, one can naturally adopt $\text{TPP}_\text{body}$ or $\text{TPP}_\text{active}$ for MoEs.
\begin{figure}[!htb]
    \centering
    \includegraphics[width=0.48\linewidth]{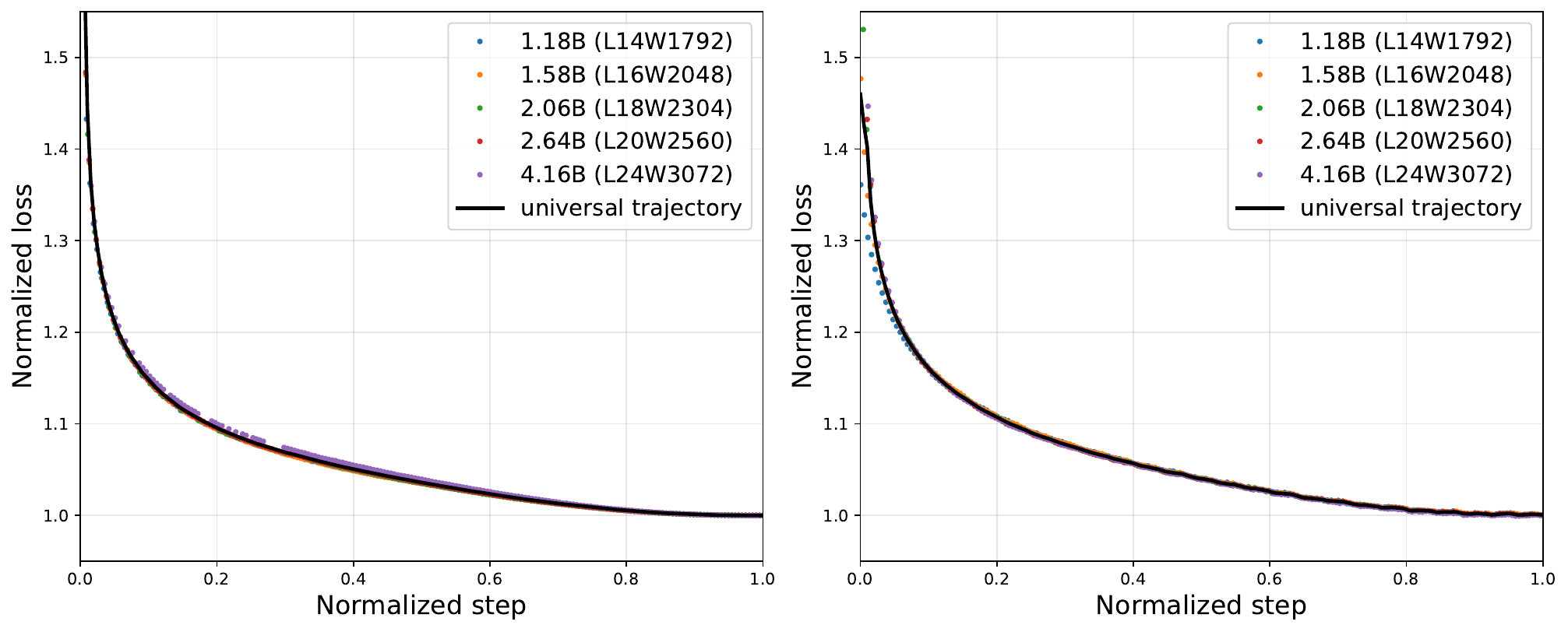}
    \includegraphics[width=0.48\linewidth]{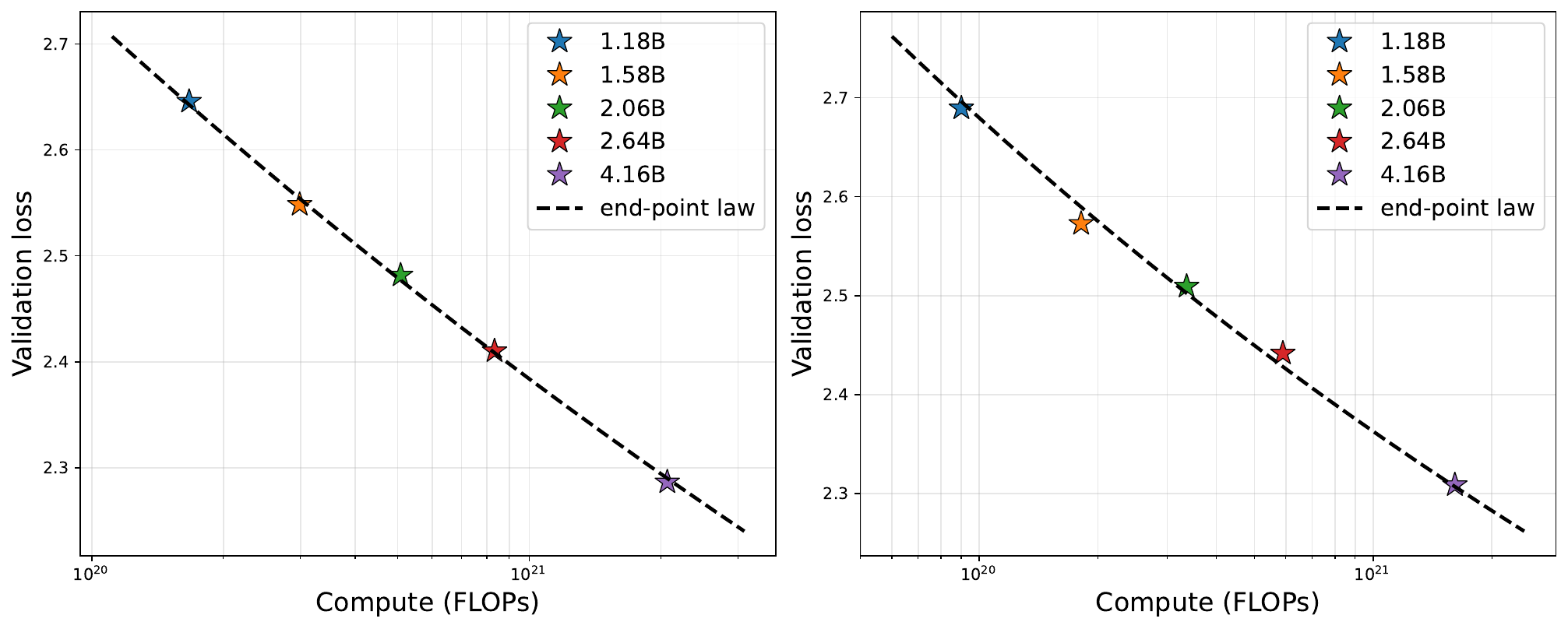}
    \captionof{figure}{Universal scaling holds under $\TPP_\text{total}=20$ (left) and $\TPP_\text{body}=20$ (right) in sufficiently scaled regimes.}
    \label{fig:placeholder}
\end{figure}

\vspace{-0.1cm}
\subsection{Optimizers and hyperparameters}
\vspace{-0.1cm}
We observe scaling collapse using various optimizers under various hyperparameters
: (1) Muon-NSGD is our default optimizer for LLM training and preserves scaling collapse across sections. (2) AdamW preserves scaling collapse in \cite{qiu2025scaling,bergsma2025scaling} as well as in \Cref{sec:mmlm}.%

\vspace{-0.1cm}
\subsection{Model architectures}
\vspace{-0.1cm}

\begin{figure}[!htb]
    \centering
    \includegraphics[width=\linewidth]{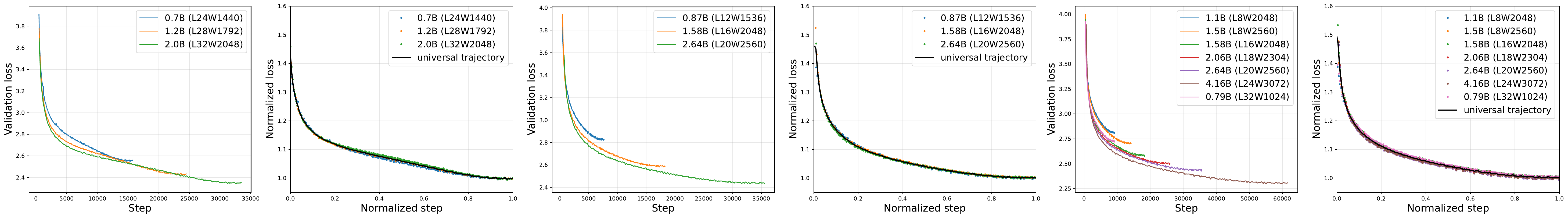}
    \vspace{-0.6cm}
    \caption{Universal trajectory holds under $\TPP_\text{body}=20$. Left two: LLaMA-3 model, no weight tying, aspect ratio$\approx 64$. Middle two: Qwen3-MoE model, weight tying, aspect ratio$= 128$. Right two: Qwen3-MoE model, no weight tying, various aspect ratios.}
    \label{fig:model arch}
\end{figure}

We observe scaling collapse under various model architectures. While our previous experiments are conducted on Qwen3-MoE without weight tying, we consistently observe the same pattern on  LLaMA-3 dense models (left two plots, no weight tying) and Qwen3-MoE with \textbf{weight tying} (middle two plots) in \Cref{fig:model arch}. Furthermore, we validate the universal scaling on Transfusion architecture in \Cref{sec:mmlm}.
In addition, we vary the \textbf{aspect ratio} from 32 to 320 in \Cref{fig:model arch} (right two plots), and observe a universal trajectory when the model size ranges from 0.8B to 4B.

\vspace{-0.1cm}
\section{Generalized universal scaling}
\vspace{-0.1cm}
We generalize universal scaling to evaluation metrics and parameter norms, beyond the optimization loss. We use $M$ to denote a metric or norm, and the end-point scaling law as $g(N,T;\theta):=M_{T,N}(T)$. The generalized universal scaling law follows
\begin{equation}
    M_{T,N}\!\left(xT\right)
    =(1-\Phi(x))
    M_\text{offset}+\Phi(x)g(N,T;\theta).
\label{eq:generalized universal}
\end{equation}
\vspace{-0.1cm}
\subsection{Universal scaling of evaluation metrics}
\vspace{-0.1cm}

\begin{figure}[!htb]
    \centering
    \includegraphics[width=\linewidth]{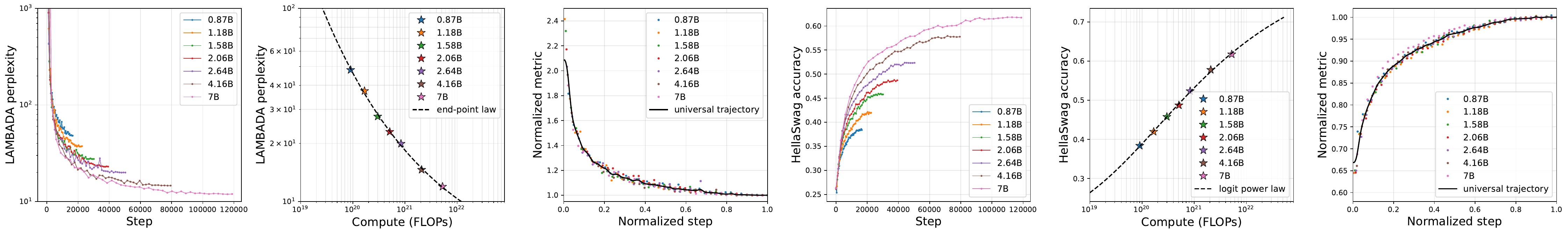}
    \vspace{-0.6cm}
    \caption{Universal scaling of continuous and discrete evaluation metrics on Qwen3-MoE, under $\TPP=20$. Left three: LAMBADA. Right three: HellaSwag. Each group shows raw metric, end-point scaling law (a power law on log perplexity for LAMBADA and on logit accuracy for HellaSwag), and generalized scaling collapse. More metrics are presented in \Cref{app:general universal}.}
    \label{fig:lambada hellaswag}
\end{figure}

We analyze a range of continuous and discrete metrics, including but not limited to LAMBADA  \citep{paperno2016lambada} and HellaSwag \citep{zellers2019hellaswag}. We fit $g$ with a power law to the logarithm of LAMBADA perplexity and to the logit of HellaSwag accuracy (see \Cref{app:eval metrics}) in \Cref{fig:lambada hellaswag}. For these metrics, the end-point laws are precise and trajectories are universal, allowing the construction of entire metric trajectory across scale. Additionally, we observe the scaling collapse of continuous and discrete metrics on multi-modal training in \Cref{sec:mmlm}.

\vspace{-0.1cm}
\subsection{Universal scaling of parameter norms}
\vspace{-0.1cm}

For universal scaling of parameter norm, we apply an initialization-aware transformation:
\begin{equation}
    \frac{
        M_{T,N}\!\left(xT\right)-M_{T,N}(0)
    }{
        M_{T,N}(T)-M_{T,N}(0)
    }
    \approx\Phi(x;\TPP).
\end{equation}
We analyze Frobenius norms $\sqrt{\sum_l \|\mathbf{W}_l\|^2}$ at various levels, agnostic to model depth and width: (1) \textit{Model-level} parameter norm. (2) \textit{Module-level} parameter norms, including the embedding layer, output layer, and hidden layers. (3) \textit{Matrix-level} parameter norms, including Q/K/V matrices, SWiGLU, experts, router, and normalization.
\begin{figure}[!htb]
    \centering
    \includegraphics[width=\linewidth]{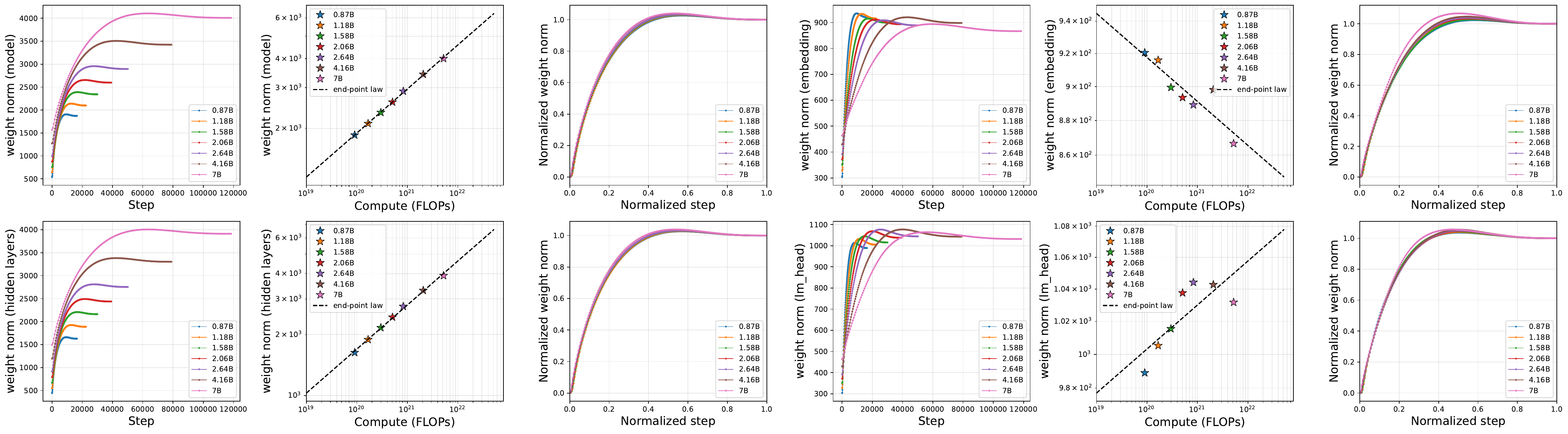}
    \vspace{-0.6cm}
    \caption{Universal scaling of parameter norms on Qwen3-MoE, under $\TPP=20$. Top-left: whole model norm. Top-right: embedding norm. Bottom-left: hidden-layer norm. Bottom-right: language model head norm. Each group shows raw norm, end-point scaling law, and generalized scaling collapse. More norms are presented in \Cref{app:general universal}.}
    \label{fig:wnorm}
\end{figure}

We fit a power law of parameter norms in \Cref{fig:wnorm}, which is sufficiently precise for the whole model and hidden layers (attention, MLP, and layer-norm), with an outlier of MoE router (\texttt{mlp.gate}). In particular, compared to the model norm in \Cref{fig:wnorm}, different norms can have significantly different universal trajectories, e.g. \texttt{self\_attn.q\_norm/k\_norm}.

\vspace{-0.1cm}
\section{Universal scaling in native multi-modal training}
\label{sec:mmlm}
\vspace{-0.1cm}
We further generalize universal scaling to multi-modal training and multi-task optimization, strictly following \citep{tong2026beyond}. Details on model architecture, data, and optimization are listed in \Cref{app:MMLM}, as well as fine-grained evaluation for individual tasks.

\paragraph{Models.} We use the Transfusion framework \citep{zhou2025transfusion} to natively unify discrete next-token prediction for text and continuous flow matching for visual generation. We adopt LLaMA-3 architecture \citep{grattafiori2024llama} and Representation Autoencoder (RAE; \citep{zheng2026diffusion,tong2026scaling}) in this framework. We employ both dense networks and modality-specific split feed-forward networks (split-FFNs).

\paragraph{Training.} We train models with 1.5B, 2.5B, 3.6B, and 8B active parameters, i.e. up to 14B total parameters for split-FFN. We fix $\TPP=100$ based on active parameters. We primarily use DCLM \citep{li2024datacomp} as text data, and Shutterstock-Image as text-image paired
data for both image-to-text understanding and text-to-image conditional generation.

\paragraph{Optimization.} Our optimization is multi-task in a single pass:
\begin{equation}
    \mathcal{L}_\text{total} = \mathcal{L}_{\text{text}} + 3 \mathcal{L}_{\text{image}}
\end{equation}
where $\mathcal{L}_{\text{text}}$ is the cross-entropy loss and $\mathcal{L}_{\text{image}}$ is the diffusion flow-matching loss. We optimize using AdamW with $1/\sqrt{T}$ hyperparameters scaling, anchored at learning rate=2e-4 and weight decay=0.1 for 1.5B model.

\begin{figure}[!htb]
    \centering
    \includegraphics[width=\linewidth]{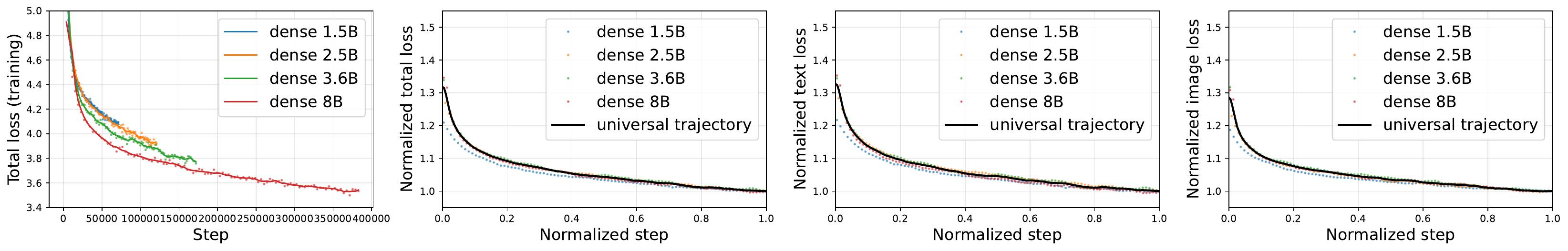}
    \includegraphics[width=\linewidth]{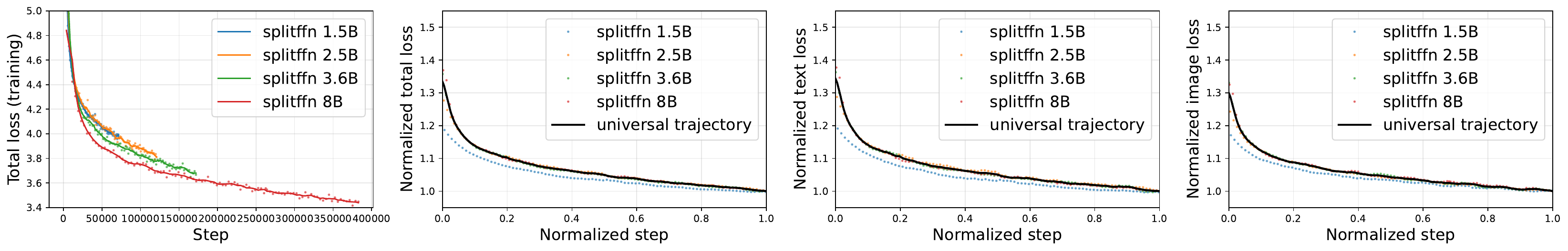}
    \vspace{-0.6cm}
    \caption{Universal scaling for multi-modal training in terms of losses. The pattern is consistent across loss type (total loss and component losses), scale (1.5-8B), modality (text and image), and architecture (dense and split-FFN).} %
\label{fig:MMLM loss}
\end{figure}
\begin{figure}[!htb]
    \centering
    \includegraphics[width=\linewidth]{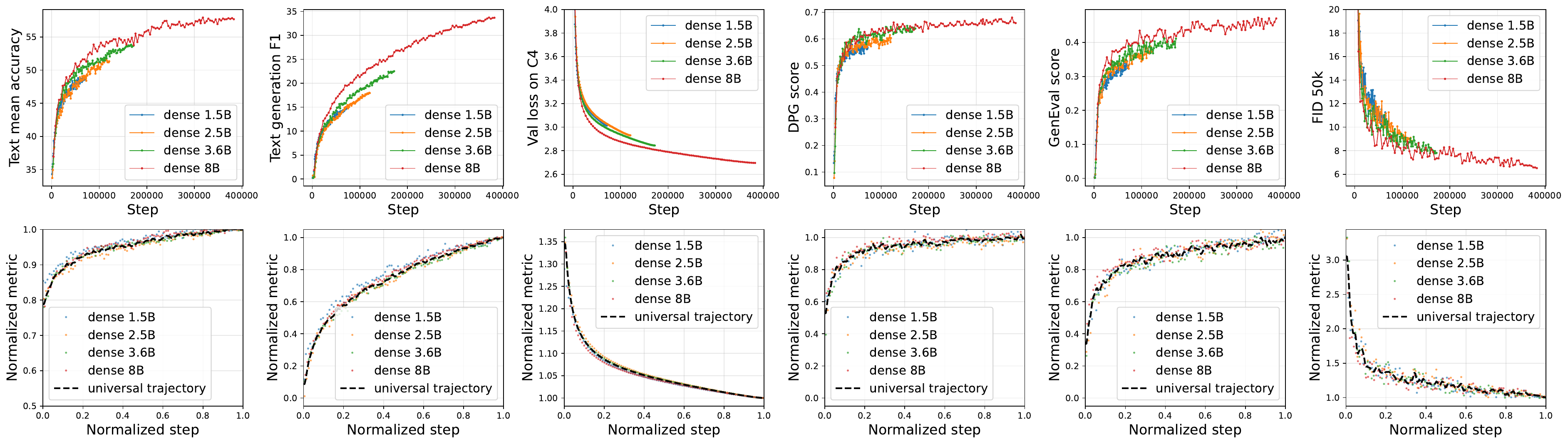}
    \vspace{-0.6cm}
    \caption{Universal scaling for multi-modal training in terms of metrics. The pattern is consistent across metric type, scale (1.5-8B), modality (text and image), and architecture (dense and split-FFN; the latter is visualized in \Cref{fig:MMLM splitffn eval}). For text mean accuracy, the generalized scaling collapse uses logit accuracy (\Cref{app:eval metrics}).}
    \label{fig:MMLM dense eval}
\end{figure}

Universal scaling consistently holds for dense and split-FFN models across losses and evaluation metrics. To be specific, universal scaling is observed not only for the directly optimized loss $\mathcal{L}_\text{total}$ but also for component losses $\mathcal{L}_\text{text}$ and $\mathcal{L}_\text{image}$ in \Cref{fig:MMLM loss}. This holds across scale in each sub-plot, and across modality by comparing columns. Furthermore, it is observed in \Cref{fig:MMLM dense eval} and \Cref{fig:MMLM splitffn eval} for multiple types of evaluation metrics: (1) \textbf{Discriminative text accuracy} -- multiple-choice accuracy on 8 commonsense and knowledge benchmarks. (2) \textbf{Generative text score} -- F1 on the NQ \citep{kwiatkowski2019natural} and TriviaQA \citep{joshi2017triviaqa}. (3) \textbf{Continuous text perplexity} on C4 dataset. (4) \textbf{Generative image score}, such as DPG \citep{hu2024ella}, GenEval \citep{ghosh2023geneval}, and FID \citep{heusel2017gans}.

\vspace{-0.1cm}
\section{Meta-learning on open scaling ladders}
\label{sec:meta ladders}
\vspace{-0.1cm}
In this section, we conduct meta-learning without training any models, thanks to the community efforts on open scaling experiments and releases of intermediate checkpoints. Following our recipe in \Cref{app:recipe}, we analyze
\begin{itemize}
    \item MAI-Base-2 \citep{microsoftai2026frontier}, which extends our study to a 2.8T-parameter model (80B active) on 50T tokens. TPP fixed at 625 per active parameter. Peak learning rate scales as $T^{-0.573}$ and weight decay scales so that $\eta\lambda T$ is fixed. Aspect ratio is fixed.
    \item DataDecide \citep{magnusson2025datadecide}, whose $\approx 20k$ checkpoints span 25 pretraining datasets, 10 model sizes, and 11 perplexity metrics. TPP fixed at 100 per parameter. Effective peak learning rate scales as $T^{-1}$ and weight decay is constant, so that $\eta\lambda T$ is fixed. Aspect ratio varies from 8 to 128.
\end{itemize}

\vspace{-0.1cm}
\subsection{Universal scaling towards trillion-parameter models}
\label{sec:trillion ladder}
\vspace{-0.1cm}
\citet{microsoftai2026frontier} trains a scaling ladder of MoE models with \emph{multi-epoch} pre-training, matching data repetition across training horizons. Each rung is determined by its number of layers $L$: L18, L24, L30, L36, and L42 (estimated at 0.7B--7B active parameters), and L102 (MAI-Base-2). MAI-Base-2's trajectory covers only the last $\sim$80\% of its compute, and the end-point scaling law is provided explicitly as $L(C)=0.696+0.428\,(C/10^{22})^{-0.177}$.

\begin{figure}[!htb]
    \centering
    \includegraphics[width=\linewidth]{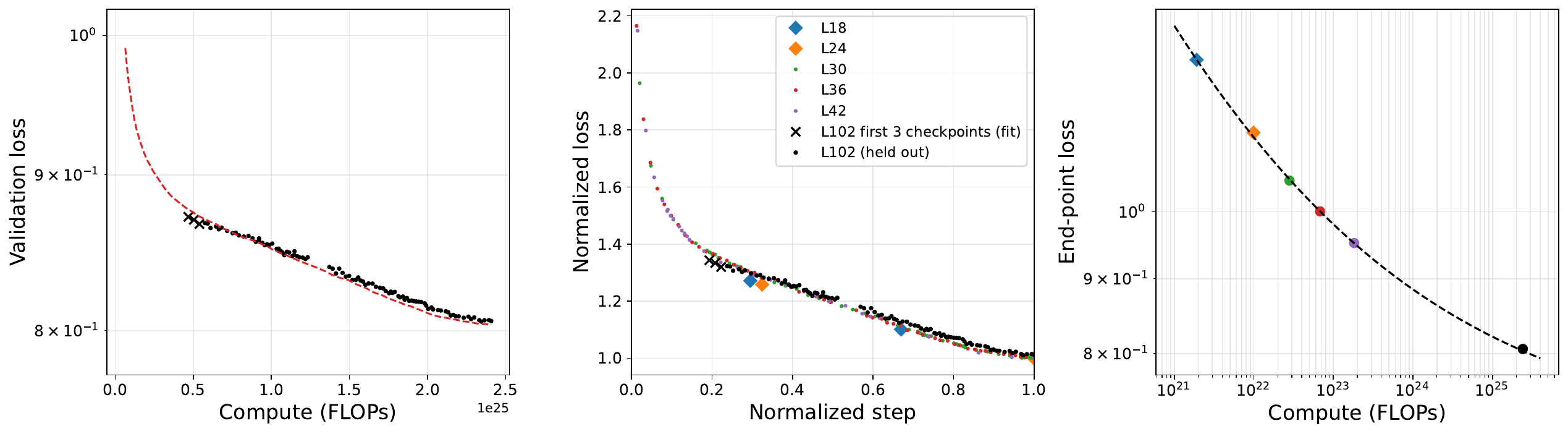}
    \vspace{-0.6cm}
    \caption{Forecasting MAI-Base-2 (2.8T param, L102) from its scaling ladder with the universal scaling law. Left: Weighted validation loss of MAI-Base-2 and the forecast (dashed). Only the first 3 out of 95 checkpoints (crosses) are used for fitting; the remaining checkpoints (dots), including the final one, are held out. Middle: normalized loss with the fitted $T_\text{offset}=676$. Right: end-point power law, fitted by \citet{microsoftai2026frontier}.}
    \label{fig:mai universal}
\end{figure}

It therefore suffices to fit only the universal trajectory to obtain the universal scaling law, which we fit on the ladder's trajectories together with the first 3 of MAI-Base-2's checkpoints (all before 23\% of training). In \Cref{fig:mai universal}, our forecast extrapolates about $130\times$ in compute: the relative deviation from the held-out checkpoints is at most $0.68\%$ and on average $0.34\%$, and the final loss is within $0.3\%$.

\vspace{-0.1cm}
\subsection{Universal scaling across thousands of trajectories}
\label{sec:multi ladder}
\vspace{-0.1cm}
\citep{magnusson2025datadecide} trains a suite of OLMo models \citep{groeneveld2024olmo,bhagia2024establishing} on different pre-training datasets and model sizes, where the hyperparameter scaling follows \citep{porian2024resolving}. We filter out less complete or smallest runs, keeping model sizes between 10M and 1B parameters and ensuring each trajectory covers at least 88\% of training.

\begin{figure}[!htbp]
    \centering
    \includegraphics[width=0.93\linewidth]{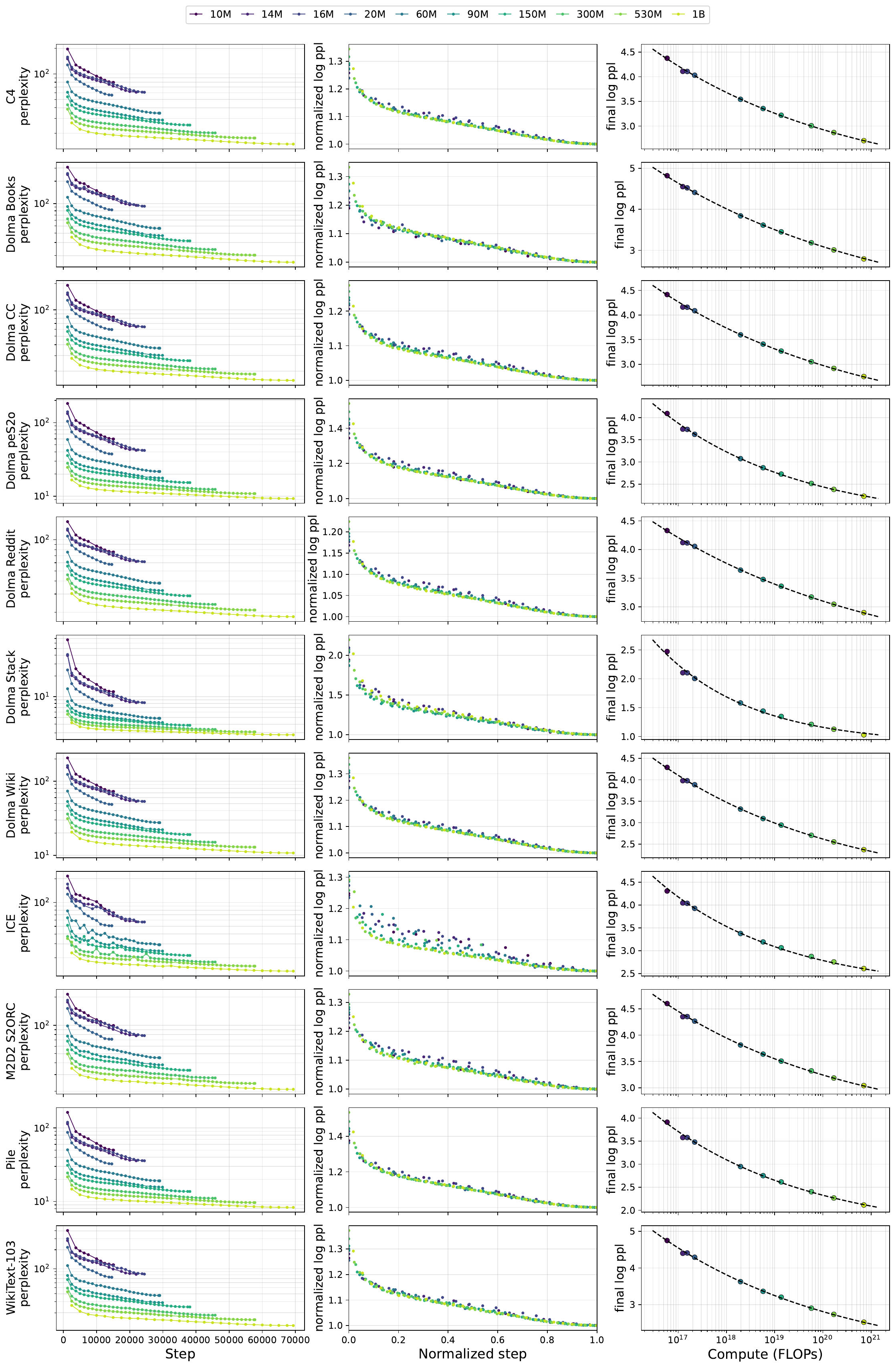}
    \caption{Universal scaling across 10 model sizes (10M--1B) trained on Dolma 1.7 pretraining dataset, for perplexity on 11 validation sets (rows). Left: raw perplexity. Middle: normalized log perplexity, with $T_\text{offset}=1249$ shared by all rows and $M_\text{offset}$ fitted per row by the scaling collapse on $x\in(0,1]$. Right: end-point power law fitted to all checkpoints at $x>0.5$ projected onto the universal trajectory.}
    \label{fig:datadecide ppl}
\end{figure}

In \Cref{fig:datadecide ppl}, we apply the universal scaling law to Dolma 1.7 dataset. The trajectories of 10 model sizes spanning $100\times$ in parameters collapse onto one curve, and the end-point power laws enjoy $R^2$ between $0.9952$ and $0.9997$. The relative deviation from the observed checkpoints is at most $3.49$--$6.87\%$ and $0.31$--$1.32\%$ on average across the 11 validation sets. We repeat the visualization in \Cref{app:meta learning} on the other 24 pretraining datasets, covering 2,750 trajectories (25 datasets $\times$ 10 sizes $\times$ 11 sets).

\vspace{-0.1cm}
\section{Conclusion}
\vspace{-0.1cm}
In this work, we study universal scaling laws across LLM and MMLM to enable the precise prediction of observable trajectories. To establish our unified framework, we advance the scaling collapse transformation with a closed-form parameterization of the universal trajectory, and leverage convex optimization theory to consolidate the hyperparameter choices. We demonstrate that universal scaling extends the standard end-point scaling to encompass full dynamics across loss values, parameter norms, and evaluation metrics, which is validated through a series of ablations of model architectures, optimizers, and training configurations. Together, these contributions yield a practical recipe towards universal scaling.

\clearpage
\bibliographystyle{assets/plainnat}
\bibliography{references}

\clearpage
\beginappendix

\appendix
\section{Recipe of universal scaling}
\label{app:recipe}
We summarize the recipe for multiple runs to follow the universal scaling.
\begin{enumerate}
    \item TPP is fixed. The notion of TPP can be $\TPP_\text{active}$, $\TPP_\text{body}$, or $\TPP_\text{total}$.
    \item Choice of model family is flexible, e.g. dense model and sparse MoE, or LLM and MMLM.
    \item Model size or TPP is sufficiently large, e.g. model $\geq 2B$ parameters for $\TPP=20$, or model $\geq 1B$ parameters for $\TPP=100$.
    \item Learning rate schedule shape is fixed, e.g. cosine/WSD/linear schedule that decays to a fixed fraction of peak learning rate. We can use fixed-ratio warmup or fixed-step warmup, because the learning rate schedule shape is asymptotically identical.
    \item The optimizer is fixed with decoupled weight decay or no weight decay.
    \item $1/\sqrt{T}$-scaling on learning rate and weight decay, though fixing $\eta\lambda T$ (e.g.\ $\eta=1,\lambda=1/T$) is sufficient to preserve universal trajectory.
    \item We recommend that the aspect ratio be fixed, though universal scaling holds when it varies moderately. We also recommend that the head dimension be fixed.
    \item We recommend that, for MoE, the number of experts and top-k routing be fixed, which is necessary to lock the ratio $\TPP_\text{body}/\TPP_\text{active}$. Consequently, sweeping models at a fixed $\TPP_\text{active} = 80$ is mathematically equivalent to sweeping them at a fixed $\TPP_\text{body} = 20$, if $\TPP_\text{body}/\TPP_\text{active}=4$.
    \item Weight tying can be used or not.
\end{enumerate}

\subsection{Columns of \Cref{tab:collapse}}
\label{app:table1}
\begin{itemize}
    \item \textbf{TPP}: the tokens-per-parameter notion held fixed across scales. ``total'' counts all parameters; ``general'' allows $\TPP_\text{total}$, $\TPP_\text{body}$, or $\TPP_\text{active}$ (\Cref{fig:placeholder}).
    \item \textbf{modality}: text refers to language modeling; vision refers to the text-image training of \Cref{sec:mmlm}.
    \item \textbf{model arch}: model architectures; ``general'' covers dense, MoE (\Cref{fig:model arch}), and Transfusion (\Cref{sec:mmlm}) models. \citet{bergsma2025scaling} also experiment with MoE in their Figure 13, but the scaling collapse is weak.
    \item \textbf{optimizer}: the training optimizer.
    \item \textbf{timescale $\eta\lambda T$}: the product of the peak learning rate $\eta$, the weight decay $\lambda$, and the training horizon $T$, the inverse of the optimization timescale. ``0'': \citet{qiu2025scaling} use no weight decay ($\lambda=0$), and the learning rate scales as $T^{-1/2}$ only in the hidden layers, through the width scaling of $\mu$P. ``fixed'': \citet{bergsma2025scaling} scale $\eta\propto T^{-1/3}$ through CompleteP and $\lambda\propto T^{-2/3}$; we scale both as $T^{-1/2}$ (\Cref{thm:scaling hyperparam}).
    \item \textbf{$L_\text{offset}, T_\text{offset}$}: the loss offset in \eqref{eq:new collapse} and the training-progress offset in \eqref{eq:progress}; ``$L_*$'' means $L_\text{offset}$ is the irreducible loss of the end-point law, ``none'' means $T_\text{offset}=0$ ($x=t/T$), and ``fitted'' means estimated.
    \item \textbf{observable}: the quantities shown to collapse; metric and norm refer to \Cref{fig:lambada hellaswag} and \Cref{fig:wnorm}.
\end{itemize}

\section{Theoretical analysis}
\label{app:theory}
\subsection{Proof of \Cref{thm:scaling hyperparam}}
Denote $L_{T,N}^\lambda(\w)=L_{T,N}(\w)+\frac{\lambda}{2}\|\w\|^2$, then weight-decayed SGD on $L_{T,N}$ is equivalent to vanilla SGD on $L_{T,N}^\lambda$. It is clear that $\|\nabla L_{T,N}^\lambda\|\leq \|\nabla L_{T,N}\|+\lambda \|\w\|\leq G_N+\lambda W$. Via Theorem 1 in \cite{li2026towards} that holds for vanilla SGD:
$$\E L_{T,N}^\lambda(\w_T) - L_N^{\lambda}\leq \frac{D_N^2}{2\sum_{t=1}^{T} \eta_t}+
\frac{1}{2}\sum_{t=1}^{T-1}\frac{\eta_t^2 \|\nabla L_{T,N}^\lambda\|^2}{\sum_{k=t+1}^{T} \eta_k},
$$
as long as the learning rate decays to zero for any fixed $T$.

Using the bounded gradient, we obtain
$$\E L_{T,N}^\lambda(\w_T) - L_{N}^{\lambda}\leq \frac{D_N^2}{2\sum_{t=1}^{T} \eta_t}+
\frac{1}{2}\sum_{t=1}^{T-1}\frac{\eta_t^2  (G_N+\lambda W)^2}{\sum_{k=t+1}^{T} \eta_k}
$$
For $1/\sqrt{T}$-scaled peak learning rate, we get the following upper bound:
$$\E L_{T,N}^\lambda(\w_T) - L_{N}^{\lambda}=O(D_N G_N/\sqrt{T})$$

To translate the upper bound from $L_{T,N}^\lambda$ to $L_{T,N}$, we see that, with $\w_*$ minimizing $L_{T,N}$ not $L_{T,N}^{\lambda}$,
$$L_N^{\lambda}-L_N\leq \frac{\lambda}{2}\|\w_*\|^2\leq \frac{\lambda}{2} W^2$$
All in all, we need $\lambda=O(1/\sqrt{T})$ so that
$$\E L_{T,N}^{\lambda}(\w_T) - L_{N}= O(1/\sqrt{T})$$

\subsection{Robbins-Monro conditions}
Alternatively, we view gradient as signal and $\lambda \w_t$ as noise. We require
$$
\sum_{t=1}^T \eta_t=\Theta(\sqrt{T}),\qquad
\sum_{t=1}^T \eta_t^2=O(1),\qquad
\sum_{t=1}^T \eta_t\lambda\|\w_t\|=O(1),
$$
which are satisfied by $1/\sqrt{T}$-scaling. Therefore, under the standard assumptions for stochastic convex optimization,
the weight-decay perturbation is asymptotically negligible and the optimization error converges to zero.

\subsection{Closed-form structure of $\Phi$}
\label{app:closed form derivation}
We start from Eq.~\eqref{eq:loss seq2seq}:
\begin{align*}
L_{T,N}(t;\eta)-L_N
\approx
\frac{D_N^2}{2\sum_{\tau=1}^t \eta_\tau}
+
\frac{G_N^2}{2}
\sum_{\tau=1}^{t-1}
\frac{\eta_\tau^2}
{\sum_{k=\tau+1}^t \eta_k}
+
\frac{G_N^2}{2}\eta_t .
\label{eq:loss-seq2seq}
\end{align*}
Let
\begin{equation*}
    \eta_\tau=\eta\,s(\tau/T),
    \qquad
    t=\lfloor xT\rfloor,
    \qquad x\in(0,1],
\end{equation*}
where $s$ denotes the normalized learning rate
schedule.

For the first term, Riemann-sum approximation gives
\begin{align*}
\sum_{\tau=1}^{xT}\eta_\tau
=
\eta\sum_{\tau=1}^{xT}s(\tau/T)
=
\eta T\int_0^x s(u)\,du+o(\eta T),
\end{align*}
and hence
\begin{equation*}
\frac{D_N^2}{2\sum_{\tau=1}^{xT}\eta_\tau}
=
\frac{D_N^2}
{2\eta T\int_0^x s(u)\,du}
+o\!\left(\frac{1}{\eta T}\right).
\end{equation*}

For the second term,
\begin{align*}
\sum_{\tau=1}^{xT-1}
\frac{\eta_\tau^2}
{\sum_{k=\tau+1}^{xT}\eta_k}
&=
\eta
\sum_{\tau=1}^{xT-1}
\frac{s^2(\tau/T)}
{\sum_{k=\tau+1}^{xT}s(k/T)}.
\end{align*}
Near $\tau=xT$, the summand behaves as
$s(x)/(xT-\tau)$, producing a harmonic sum. Therefore, under mild
regularity conditions on $s$,
\begin{equation*}
\sum_{\tau=1}^{xT-1}
\frac{s^2(\tau/T)}
{\sum_{k=\tau+1}^{xT}s(k/T)}
=
s(x)\log T+C_s(x)+o(1),
\end{equation*}
where $C_s(x)$ is a finite schedule-dependent correction. In particular,
one may write
\begin{equation*}
C_s(x)
=
s(x)\bigl(\log x+0.577215\bigr)
+
\int_0^x
\left[
\frac{s^2(u)}
{\int_u^x s(v)\,dv}
-
\frac{s(x)}{x-u}
\right]du ,
\end{equation*}
with $0.577215$ denoting the Euler--Mascheroni constant. Thus,
\begin{equation*}
\frac{G_N^2}{2}
\sum_{\tau=1}^{xT-1}
\frac{\eta_\tau^2}
{\sum_{k=\tau+1}^{xT}\eta_k}
=
\frac{\eta G_N^2}{2}
\left[
s(x)\log T+C_s(x)+o(1)
\right].
\end{equation*}

Finally,
\begin{equation*}
\frac{G_N^2}{2}\eta_{xT}
=
\frac{\eta G_N^2}{2}s(x)+o(\eta).
\end{equation*}

Combining the three terms yields
\begin{align*}
L_{T,N}(xT;\eta)-L_N
\approx\;&
\frac{D_N^2}
{2\eta T\int_0^x s(u)\,du}
+
\frac{\eta G_N^2}{2}
\left[
s(x)\log T
+C_s(x)
+s(x)
\right],
\label{eq:trajectory-asymptotic}
\end{align*}
Retaining only the leading dependence on $T$,
\begin{equation}
L_{T,N}(xT;\eta)
\approx
L_N
+
\frac{D_N^2}
{2\eta T\int_0^x s(u)\,du}
+
\frac{\eta G_N^2}{2}s(x)\log T
+
O(\eta).
\label{eq:trajectory-leading}
\end{equation}

Under fixed $\TPP=\frac{BT}{N}$ and fixed batch size $B$, we have
$$    \log T=\log N+\log\TPP-\log B.$$
The last two terms are independent of scale and can be
absorbed into the $O(\eta)$ schedule-dependent correction. Therefore,
$$
L_{T,N}(xT;\eta)
\approx
L_N
+
\frac{D_N^2}
{2\eta T\int_0^x s(u)\,du}
+
\frac{\eta G_N^2}{2}s(x)\log N
$$

After factoring out $D_N^2/(2\eta T)$, the coefficient of $s(x)$ is
\begin{equation*}
R_N=\frac{\eta^2T G_N^2\log N}{D_N^2}.
\end{equation*}
Using a single fitted coefficient $R$ therefore assumes that
$R_N\approx R$ across the model scales considered. This is a
scale-family assumption rather than a consequence of convexity; if it
fails, the normalized trajectories need not collapse to a common
$\Phi$.

\section{Coefficient optimization}
\label{app:sequential-estimation}
We derive the minimizations \eqref{eq:sequential-collapse} and \eqref{eq:sequential-endpoint}. For $i\in\mathcal I$ and $t(x)=T_\text{offset}+x(T_i-T_\text{offset})$,
\begin{equation}
    F_i(x;T_\text{offset},L_\text{offset})=\frac{L_{T_i,N_i}(t(x))-L_\text{offset}}{L_{T_i,N_i}(T_i)-L_\text{offset}}.
\end{equation}

\subsection{Offsets}
\label{app:offsets}
We fit all normalized curves to one trajectory:
\begin{equation}
    \min_{T_\text{offset},L_\text{offset},\Phi}\frac{1}{|\mathcal I|}\sum_{i\in\mathcal I}\int_0^1\bigl[F_i(x;T_\text{offset},L_\text{offset})-\Phi(x)\bigr]^2\,dx.
    \label{eq:offsets-joint}
\end{equation}
For fixed offsets, the functional optimization over $\Phi$ separates across $x$: $\sum_{i\in\mathcal I}[F_i(x)-\Phi(x)]^2$ is a convex quadratic in the scalar $\Phi(x)$, minimized by the pointwise mean
\begin{equation}
    \bar\Phi(x)=\frac{1}{|\mathcal I|}\sum_{i\in\mathcal I}F_i(x;T_\text{offset},L_\text{offset}),
\end{equation}
with $\bar\Phi(1)=1$ since every $F_i(1)=1$. Substituting $\bar\Phi$ profiles out $\Phi$ and turns \eqref{eq:offsets-joint} into \eqref{eq:sequential-collapse}.

This is equivalent to the pairwise objective \eqref{eq:find offset}. At each $x$, writing $F_i$ for $F_i(x;T_\text{offset},L_\text{offset})$ and $\bar\Phi$ for $\bar\Phi(x)$,
\begin{equation}
    \sum_{i<j}(F_i-F_j)^2
    =\frac12\sum_{i,j\in\mathcal I}(F_i-F_j)^2
    =|\mathcal I|\sum_{i\in\mathcal I}F_i^2-\Bigl(\sum_{i\in\mathcal I}F_i\Bigr)^2
    =|\mathcal I|\sum_{i\in\mathcal I}(F_i-\bar\Phi)^2.
\end{equation}
Integrating over $x$, \eqref{eq:find offset} equals $|\mathcal I|^2$ times \eqref{eq:sequential-collapse}, so the two have the same minimizers.

\subsection{Power-law coefficients}
\label{app:power-law}
With the offsets and $\bar\Phi$ fixed, let $y_i(x)=L_{T_i,N_i}(t(x))-(1-\bar\Phi(x))L_\text{offset}$, $g_i=L_*+\alpha C_i^\beta$ and $S=\int_0^1\bar\Phi(x)^2\,dx$. Completing the square in $g_i$,
\begin{equation}
    \int_0^1\bigl[y_i(x)-\bar\Phi(x)g_i\bigr]^2\,dx=K_i+S\,(q_i-g_i)^2,
    \qquad
    q_i=\frac{1}{S}\int_0^1\bar\Phi(x)\,y_i(x)\,dx,
    \label{eq:sequential-square}
\end{equation}
where $K_i=\int_0^1y_i(x)^2\,dx-Sq_i^2$ does not depend on $(L_*,\alpha,\beta)$, and $S$ is common to all runs since $\bar\Phi$ is shared. Hence \eqref{eq:sequential-endpoint} is least squares of the projected end-points $q_i$ on $C_i^\beta$. For fixed $\beta$, with $z_i=C_i^\beta$ and $\bar z,\bar q$ the means over $\mathcal I$,
\begin{equation}
    \alpha(\beta)=\frac{\sum_{i\in\mathcal I}(z_i-\bar z)(q_i-\bar q)}{\sum_{i\in\mathcal I}(z_i-\bar z)^2},
    \qquad
    L_*(\beta)=\bar q-\alpha(\beta)\,\bar z,
    \label{eq:sequential-coefficients}
\end{equation}
which are unique when at least two runs have distinct $C_i$. Substituting them back leaves a one-dimensional search,
\begin{equation}
    \min_{\beta}\sum_{i\in\mathcal I}\bigl[q_i-L_*(\beta)-\alpha(\beta)C_i^\beta\bigr]^2.
    \label{eq:sequential-beta}
\end{equation}

\section{Experiments}
\paragraph{Model.}
Our experiments primarily use Qwen3 MoE architecture (Qwen3-30B-A3B) with fixed aspect ratio=128. The intermediate size is $3\times$ hidden size, and MoE intermediate size is $1/4\times$ hidden size. At every scale, we use 16 experts and top-2 routing. We use ``L8W1024'' to denote 8-layer 1024-hidden model.

Notably, our $\TPP_\text{body}/\TPP_\text{active}\approx 4$ at all scales.

\paragraph{Data and Optimization.}
We use Fineweb-Edu dataset to train models under Fully Sharded Data Parallel (FSDP-2). We use a global batch size 256, and sequence length 4096, thus 1M tokens/step. The optimized objective is cross-entropy plus 0.001 auxiliary loss, while the validation loss is purely cross-entropy. We employ Muon-NSGD optimizer similar to \cite{boreiko2025towards} but routing all 2D tensors (including embedding) to Muon component and all 1D tensors to NSGD component:
$$\w_{t+1}=(1-\lambda\eta_t)\w_t-\eta_t \text{Ortho}(\g_t)$$
where $\text{Ortho}$ is orthogonalization with Newton-Schulz iteration for 2D tensors, and normalization $\text{Ortho}(\g_t)=\frac{\g_t}{\|\g_t\|_F}$ for 1D tensors. We use gradient clipping at 1.0, and momentum at 0.95.

Our default learning rate is cosine decay to zero with $\eta_\text{ref}=\lambda_\text{ref}=1$, in order to keep hyperparameters $1/\sqrt{T}$-scaled. Learning rate warmup is $\min(500, 0.02T)$.

\section{Generalized universal scaling}
\label{app:general universal}
\subsection{Evaluation metrics}
\label{app:eval metrics}
In \Cref{fig:lambada hellaswag} and \Cref{fig:eval_others}, we establish the universal scaling on LAMBADA, HellaSwag, WikiText \citep{merity2016pointer}, ARC \citep{clark2018think}, OpenBookQA \citep{mihaylov2018can}, PIQA \citep{bisk2020piqa}, RACE \citep{lai2017race}, SCIQ \citep{welbl2017crowdsourcing}, and NQ \citep{kwiatkowski2019natural}.

\begin{figure}[!htb]
    \centering
    \includegraphics[width=0.7\linewidth]{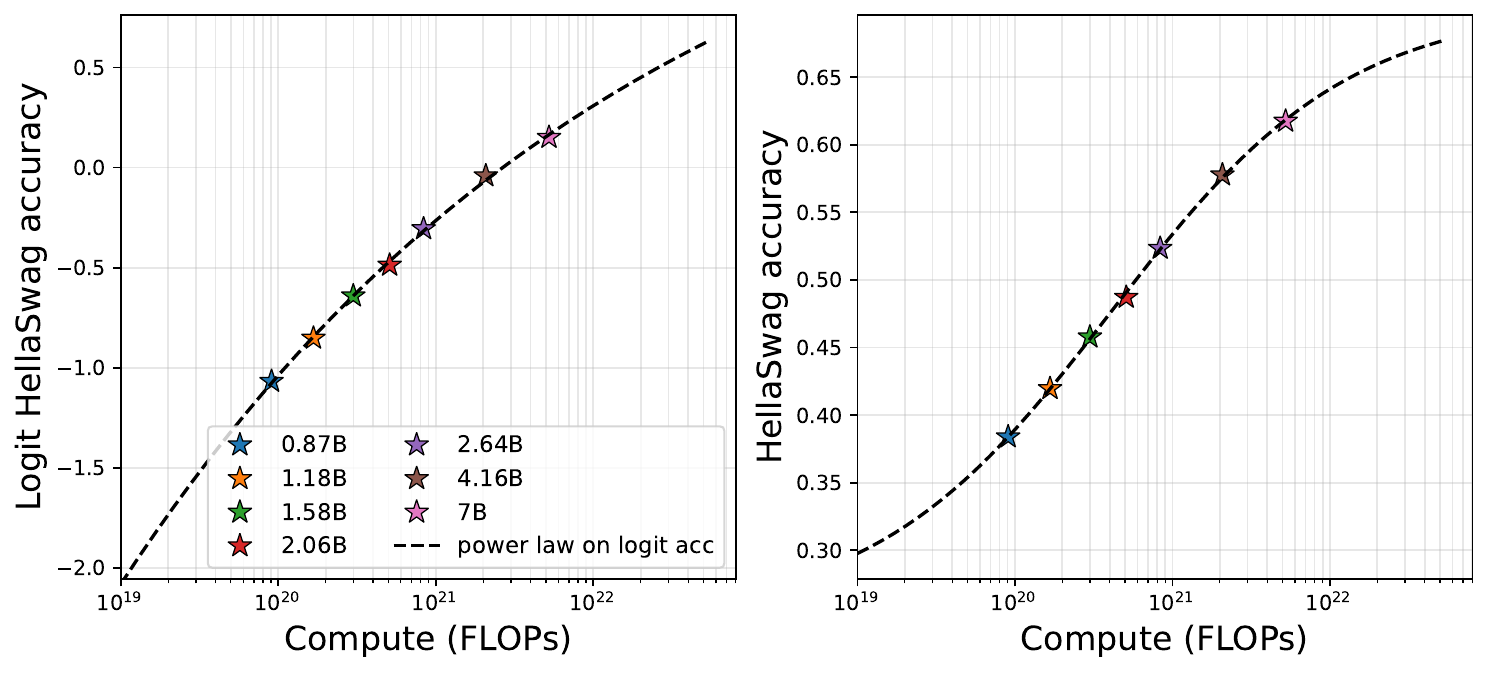}
    \vspace{-0.3cm}
    \caption{End-point laws for HellaSwag accuracy on Qwen3-MoE, under $\TPP=20$. Left: power law on logit accuracy $\ln\frac{\text{acc}-a}{1-\text{acc}}$, identical to the end-point law of HellaSwag in \Cref{fig:lambada hellaswag}. Right: sigmoid law on raw accuracy, $g(N,T)=0.25+(E-0.25)\cdot\text{sigmoid}\left(c_0+c_1\ln(6BNT)\right)$.}
    \label{fig:acc law compare}
\end{figure}

\begin{figure}[!htb]
    \centering
    \includegraphics[width=0.7\linewidth]{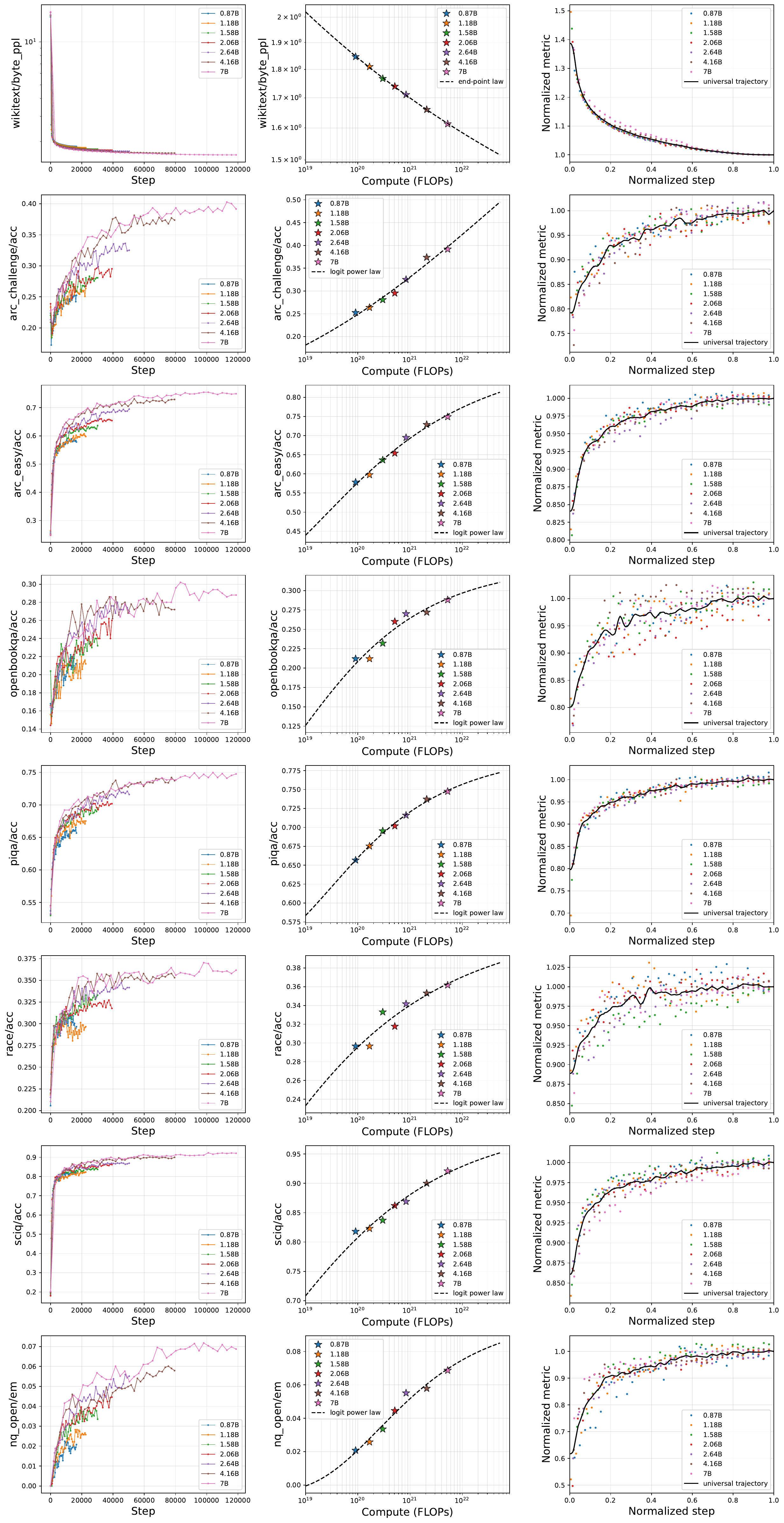}
    \caption{Universal scaling of continuous and discrete evaluation metrics on Qwen3-MoE, under $\TPP=20$. The tasks include WikiText, ARC Easy/Challenge, OpenBookQA, PIQA, RACE, SCIQ, and NQ. Each row shows raw metric, end-point scaling law (a power law on log perplexity or on logit accuracy), and generalized scaling collapse.}
    \label{fig:eval_others}
\end{figure}

For benchmark measured by discrete accuracy, we fit a power law to the logit accuracy $\ell:=\ln\frac{\text{acc}-a}{1-\text{acc}}$, where $a\geq 0$ is the accuracy floor. The power law on logit accuracy is similar to the sigmoid law on raw accuracy (see \Cref{fig:acc law compare}). We note that sigmoid laws relating training scale to benchmark accuracy have been observed both in pretraining \citep{owen2024predictable,grattafiori2024llama,bhagia2024establishing} and in reinforcement learning \citep{khatri2026art}.

\clearpage
\subsection{Parameter norms}
\begin{figure}[H]
    \centering
    \includegraphics[width=0.52\linewidth]{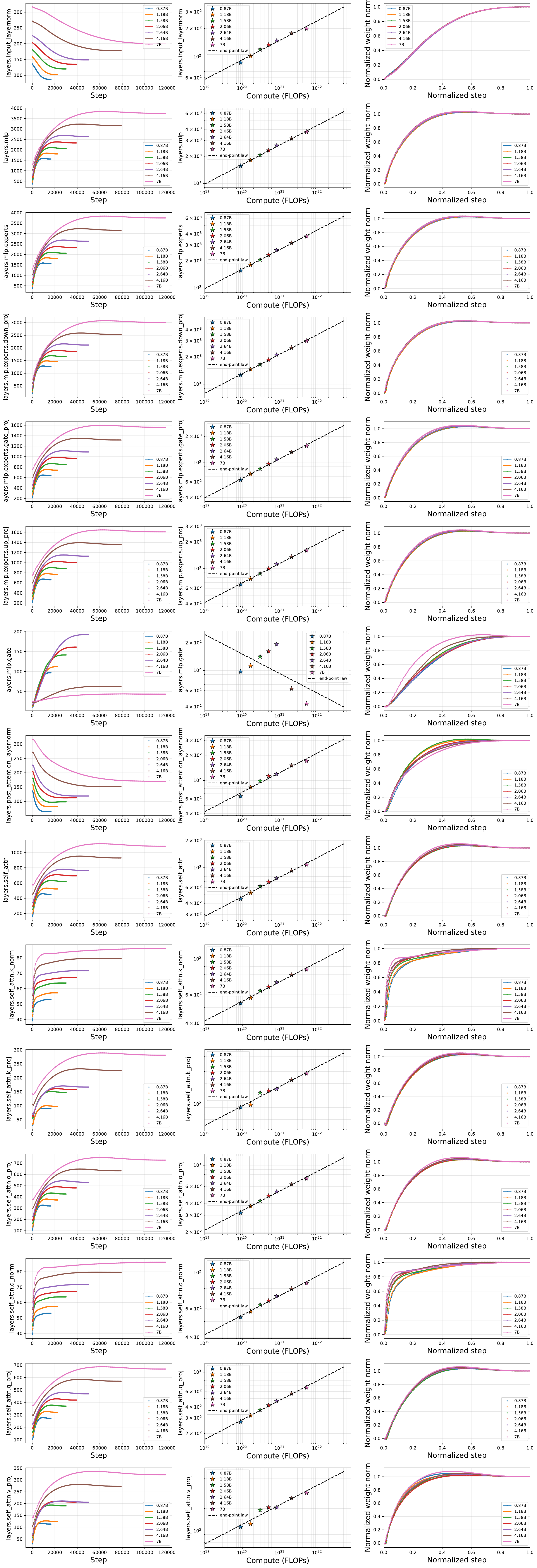}
    \caption{Universal scaling of parameter norms on Qwen3 MoE, under $\TPP=20$. Each group shows raw norm, end-point scaling law, and generalized scaling collapse.}
    \label{fig:norm_others}
\end{figure}

In \Cref{fig:wnorm} and \Cref{fig:norm_others}, we establish the universal scaling on different levels of parameter groups.

\section{Multi-modal training}
\label{app:MMLM}

\subsection{Models}
\begin{figure}[!htb]
    \centering
  \includegraphics[width=\linewidth]{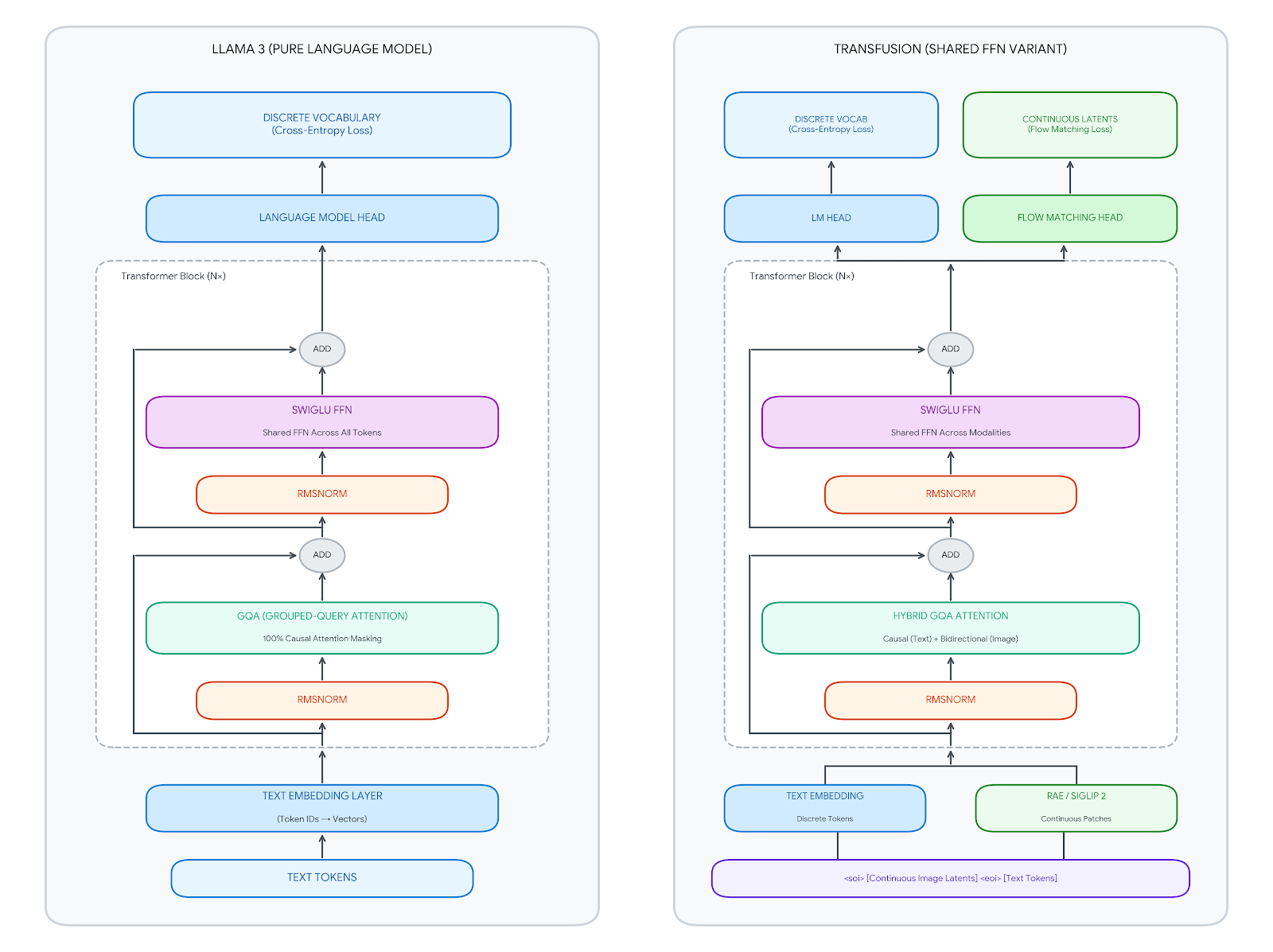}
    \vspace{-0.6cm}
    \caption{Dense MMLM (right) compared to LLaMA-3 (left).}
    \label{fig:dense MMLM}
\end{figure}

\begin{figure}[!htb]
    \centering
    \includegraphics[width=\linewidth]{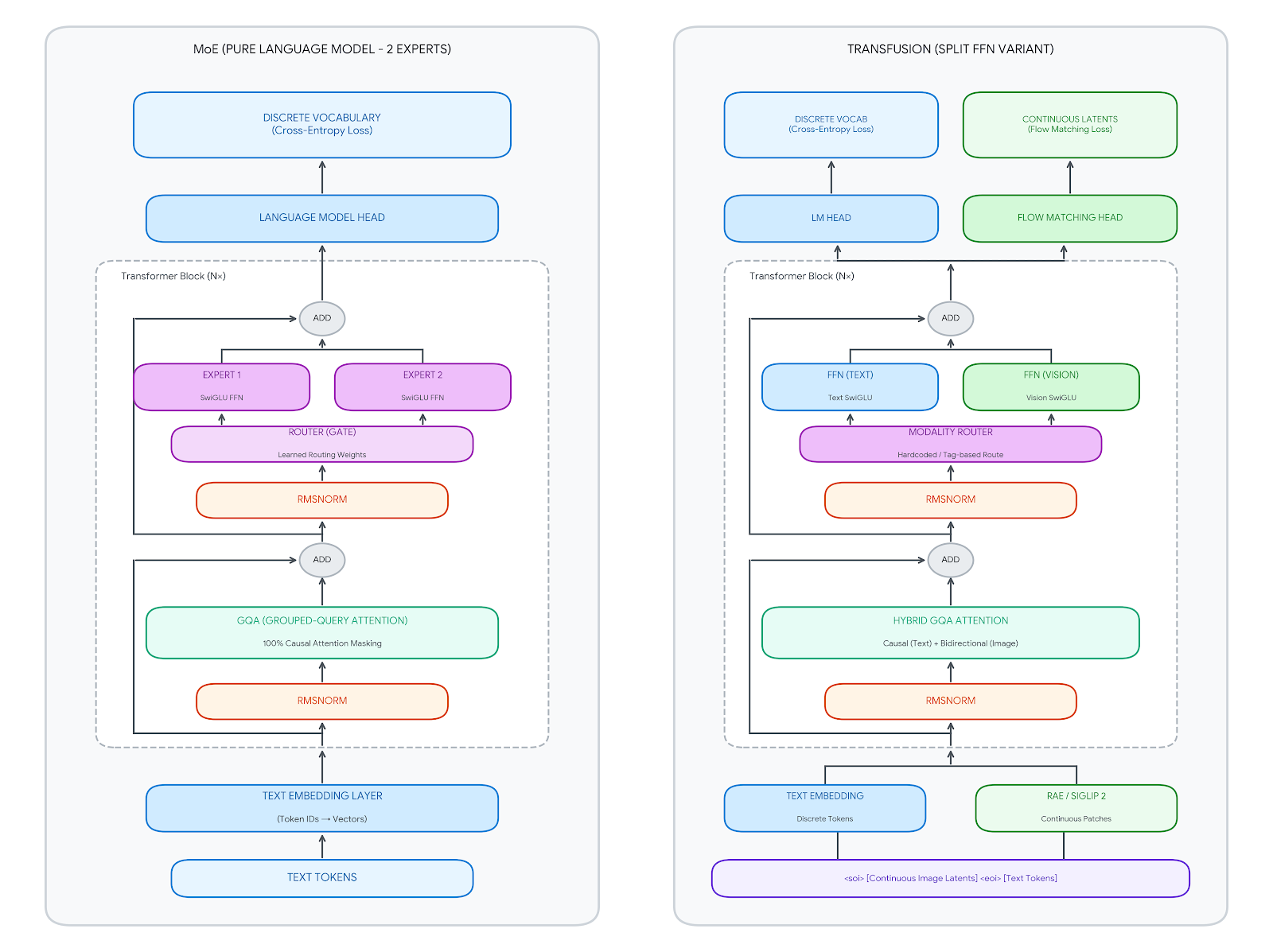}
    \vspace{-0.6cm}
    \caption{Split-FFN MMLM (right) compared to MoE (left).}
    \label{fig:splitffn MMLM}
\end{figure}

We adopt the Transfusion framework to natively unify discrete next-token prediction for text and continuous flow matching for visual generation within a single autoregressive sequence. We use two variants: dense model in \Cref{fig:dense MMLM} and split-FFN in \Cref{fig:splitffn MMLM}.

Model configurations are listed in \Cref{tab:model_configurations}.

\begin{table}[htbp]
\centering
\caption{Model configurations across different scales for split-FFN.}
\label{tab:model_configurations}
\resizebox{\linewidth}{!}{
\begin{tabular}{lcccccc}
\toprule
\textbf{Active Params} & \textbf{Layers} & \textbf{Hidden Dim} & \textbf{Query Heads} & \textbf{KV Heads} & \textbf{Head Dim} & \textbf{Total Params} \\
\midrule
1.5B  & 16 & 2048 & 32 & 8 & 64  & 2.3B  \\
2.5B&  27 &2560 &  20    & 5    &128 & 4.0B  \\
3.6B  & 28 & 3072 & 24 & 8 & 128  & 5.7B  \\
8B    & 32 & 4096 & 32 & 8 & 128  & 14B \\
\bottomrule
\end{tabular}
}
\end{table}
All configurations incorporate SwiGLU activation, Rotary Position Embeddings (RoPE) with a base frequency $\theta = 500,000$, pre-RMSNorm, grouped-query attention (GQA), QK-norm, and FlashAttention.

\subsection{Optimization}
Training is optimized using AdamW with $\beta_1 = 0.9, \beta_2 = 0.95$, and gradient clipping at 1.0.  All models are trained with a context length of 4096 tokens in \texttt{bfloat16} precision using FSDP-2. The global batch size is 512 sequences $\times$ 4096 tokens = 2M tokens/step. We adopt the $x$-prediction formulation where the network outputs the clean sample $x_0$, which is converted on-the-fly to a velocity $v = (x_0 - x_t)/(1 - t)$ for each rectified flow matching step.

\subsection{Tokenization}

\begin{itemize}
    \item \textbf{Text Tokenizer:} We use the LLaMA-3 BPE tokenizer with a vocabulary size of approximately 128,000, augmented with a small set of special multimodal control tokens (e.g., \texttt{<BOI>}, \texttt{<EOI>}).
    \item \textbf{Visual Representation (RAE):} We employ a frozen SigLIP-2 ViT-400m/14 encoder to process $224 \times 224$ images into a $16 \times 16$ grid of 256 semantic tokens. Visual generation is performed via flow matching within the SigLIP latent space and decoded back to pixels by Representation Autoencoder (RAE).
\end{itemize}

\subsection{Metrics}
\begin{figure}[!htb]
    \centering
    \includegraphics[width=\linewidth]{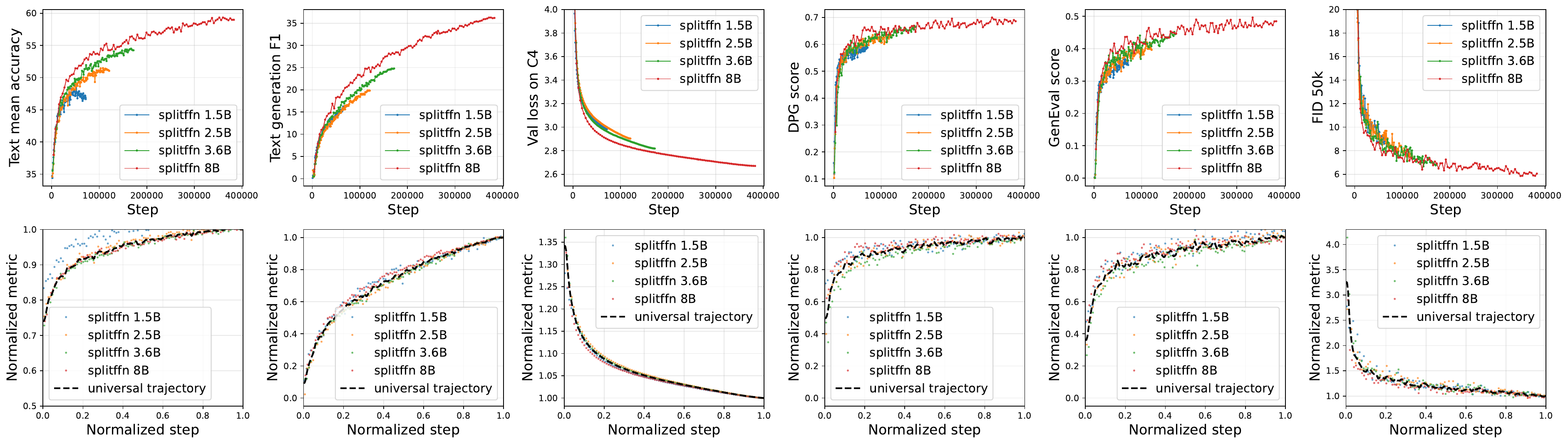}
    \vspace{-0.6cm}
    \caption{Universal scaling for multi-modal training on split-FFN. The pattern is consistent across metric type, scale (1.5-8B), and modality (text and image). The visualization on dense models is \Cref{fig:MMLM dense eval}. For text mean accuracy, the generalized scaling collapse uses logit accuracy (\Cref{app:eval metrics}).}
    \label{fig:MMLM splitffn eval}
\end{figure}

We visualize the universal scaling of split-FFN in \Cref{fig:MMLM splitffn eval}, which is consistent with that of dense models in \Cref{fig:MMLM dense eval}.

\begin{figure}[!htb]
    \centering
    \includegraphics[width=0.7\linewidth]{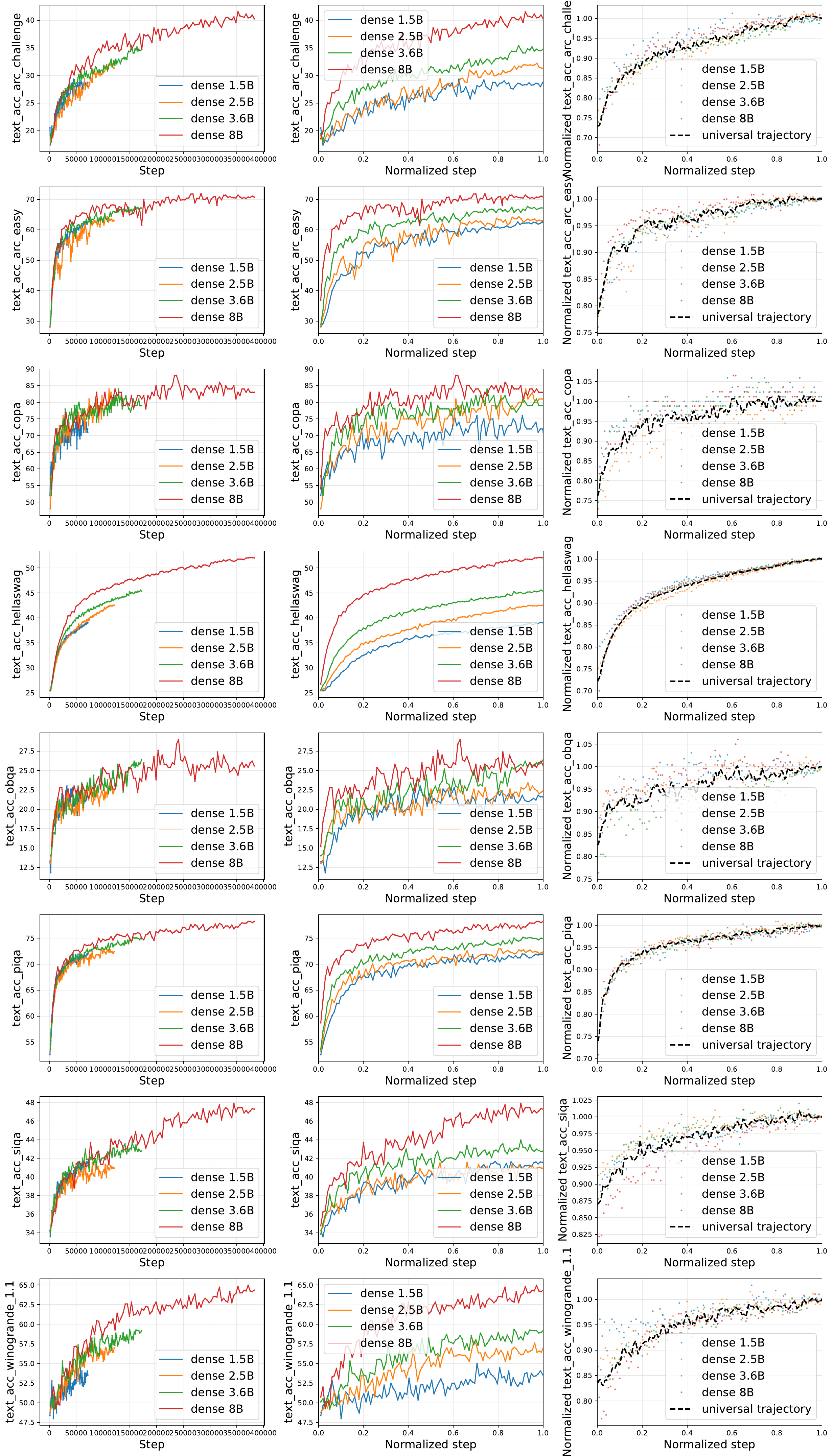}
    \caption{Individual metrics of text mean accuracy for multi-modal training on dense models. The generalized scaling collapse (right column) uses logit accuracy (\Cref{app:eval metrics}).}
    \label{fig:dense individual metric}
\end{figure}

\begin{figure}[!htb]
    \centering
    \includegraphics[width=0.7\linewidth]{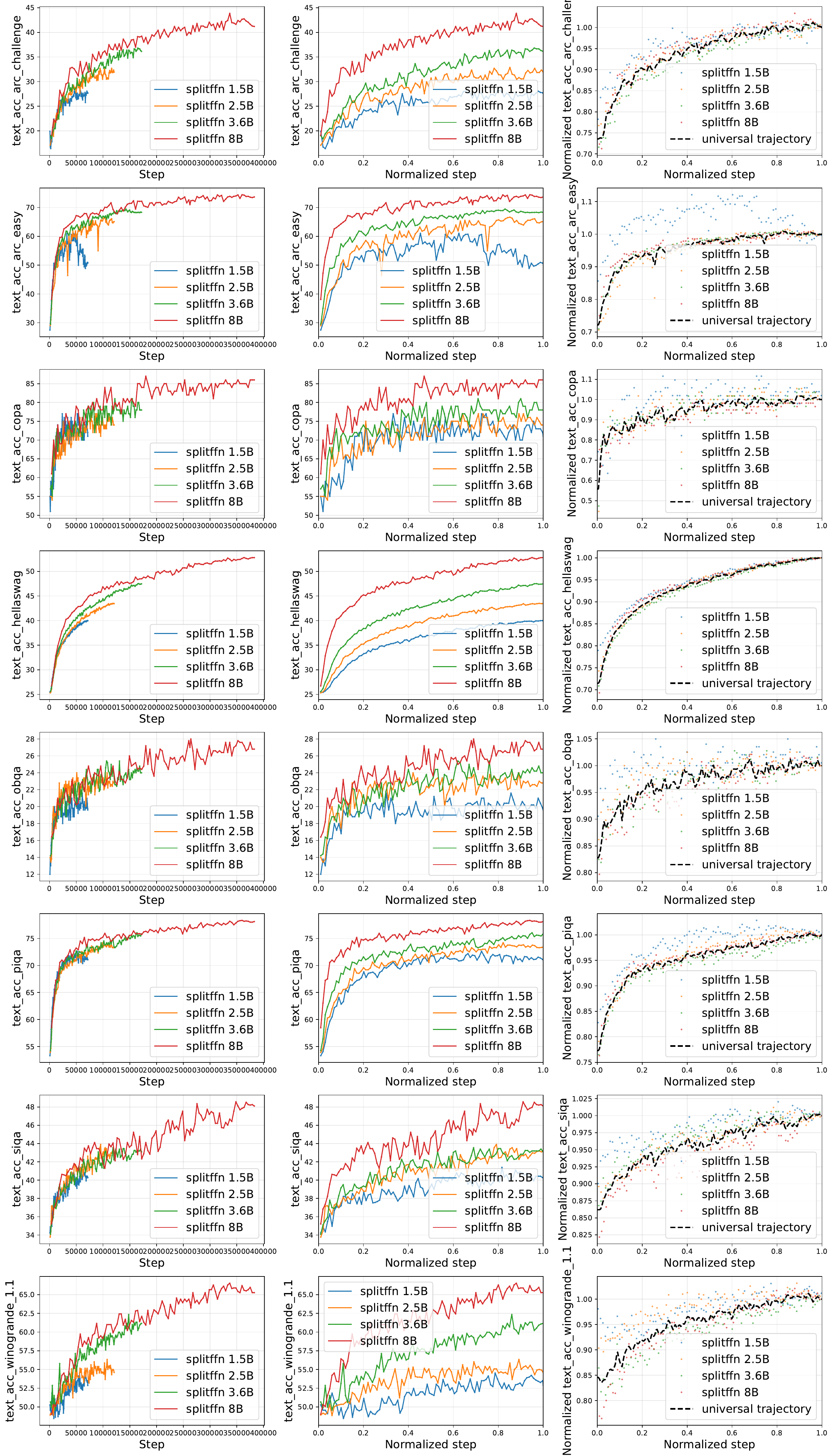}
    \caption{Individual metrics of text mean accuracy for multi-modal training on split-FFN. The generalized scaling collapse (right column) uses logit accuracy (\Cref{app:eval metrics}).}
    \label{fig:splitffn individual metric}
\end{figure}

Specifically, the text mean accuracy is averaged over 8 benchmarks: ARC-Easy/Challenge \citep{clark2018think}, COPA \citep{roemmele2011choice}, HellaSwag \citep{zellers2019hellaswag}, OpenBookQA \citep{mihaylov2018can}, PIQA \citep{bisk2020piqa}, SIQA \citep{sap2019social}, and WinoGrande \citep{sakaguchi2021winogrande}. We observe universal scaling in individual benchmarks for dense MMLM in \Cref{fig:dense individual metric}, and for split-FFN MMLM in \Cref{fig:splitffn individual metric}, with a few outliers such as split-FFN 1.5B on ARC-Easy.

\clearpage
\section{Meta-learning}
\label{app:meta learning}
We repeat \Cref{fig:datadecide ppl} on each of the other 24 DataDecide pretraining datasets, with the same model sizes, metrics, and fitting procedure.

\begin{itemize}
    \item Datasets (25): Dolma 1.7 and its ablations without code, math and code, Reddit, or Flan; Dolma 1.6++; C4; Falcon RefinedWeb; Falcon combined with Common Crawl, unfiltered or under four quality filters; FineWeb-Pro; FineWeb-Edu; DCLM-Baseline and six of its quality-filtered subsets; and DCLM-Baseline mixed with Dolma 1.7 at ratios of 25\%, 50\%, and 75\%.
    \item Models (10): 10M, 14M, 16M, 20M, 60M, 90M, 150M, 300M, 530M, and 1B parameters, trained at $\TPP=100$ (see model card in \url{https://github.com/allenai/DataDecide#hyperparameters}).
    \item Metrics (11): perplexity on the validation sets of C4, the Pile, WikiText-103, M2D2 S2ORC, ICE, and the Dolma subsets Books, Common Crawl, peS2o, Reddit, Stack, and Wiki.
\end{itemize}
\clearpage
\begin{figure}[H]
    \centering
    \includegraphics[width=0.9\linewidth]{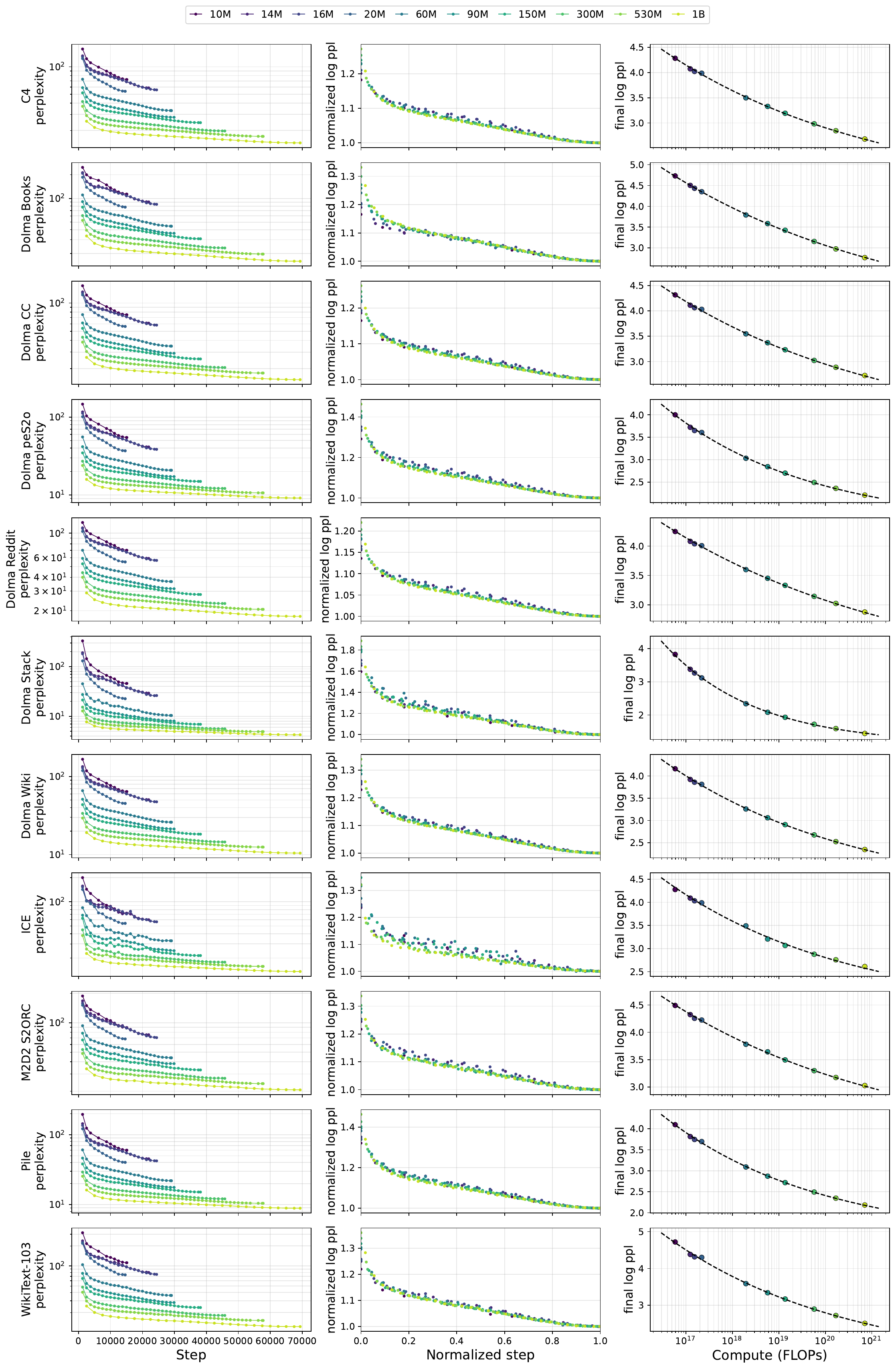}
    \caption{Universal scaling across 10 model sizes (10M--1B) trained on Dolma 1.7 (no code) pretraining dataset, for perplexity on 11 validation sets (rows), as in \Cref{fig:datadecide ppl}.}
    \label{fig:datadecide ppl dolma1_7_no_code}
\end{figure}
\clearpage

\begin{figure}[H]
    \centering
    \includegraphics[width=0.9\linewidth]{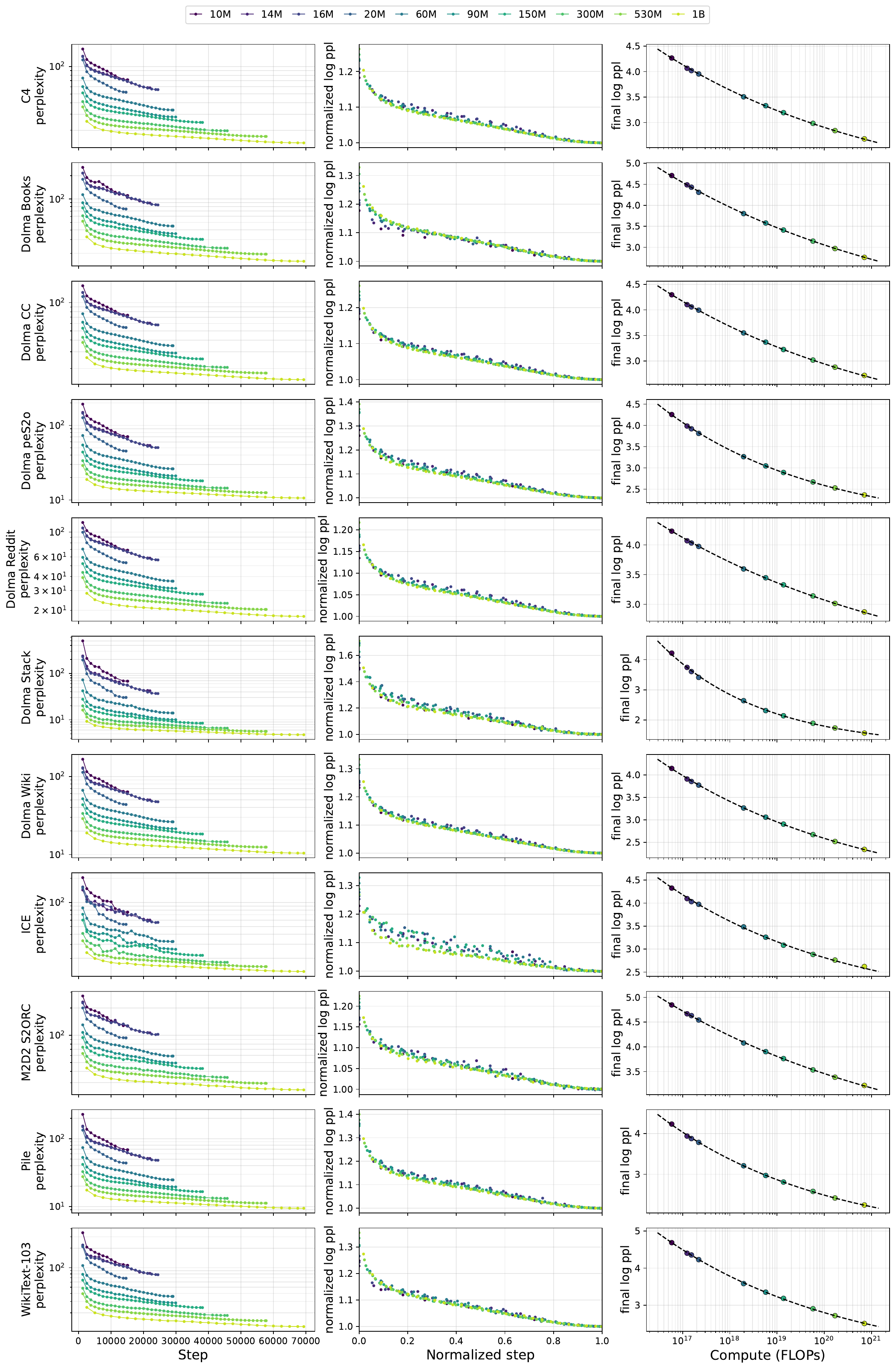}
    \caption{Universal scaling across 10 model sizes (10M--1B) trained on Dolma 1.7 (no math, code) pretraining dataset, for perplexity on 11 validation sets (rows), as in \Cref{fig:datadecide ppl}.}
    \label{fig:datadecide ppl dolma1_7_no_math_code}
\end{figure}
\clearpage

\begin{figure}[H]
    \centering
    \includegraphics[width=0.9\linewidth]{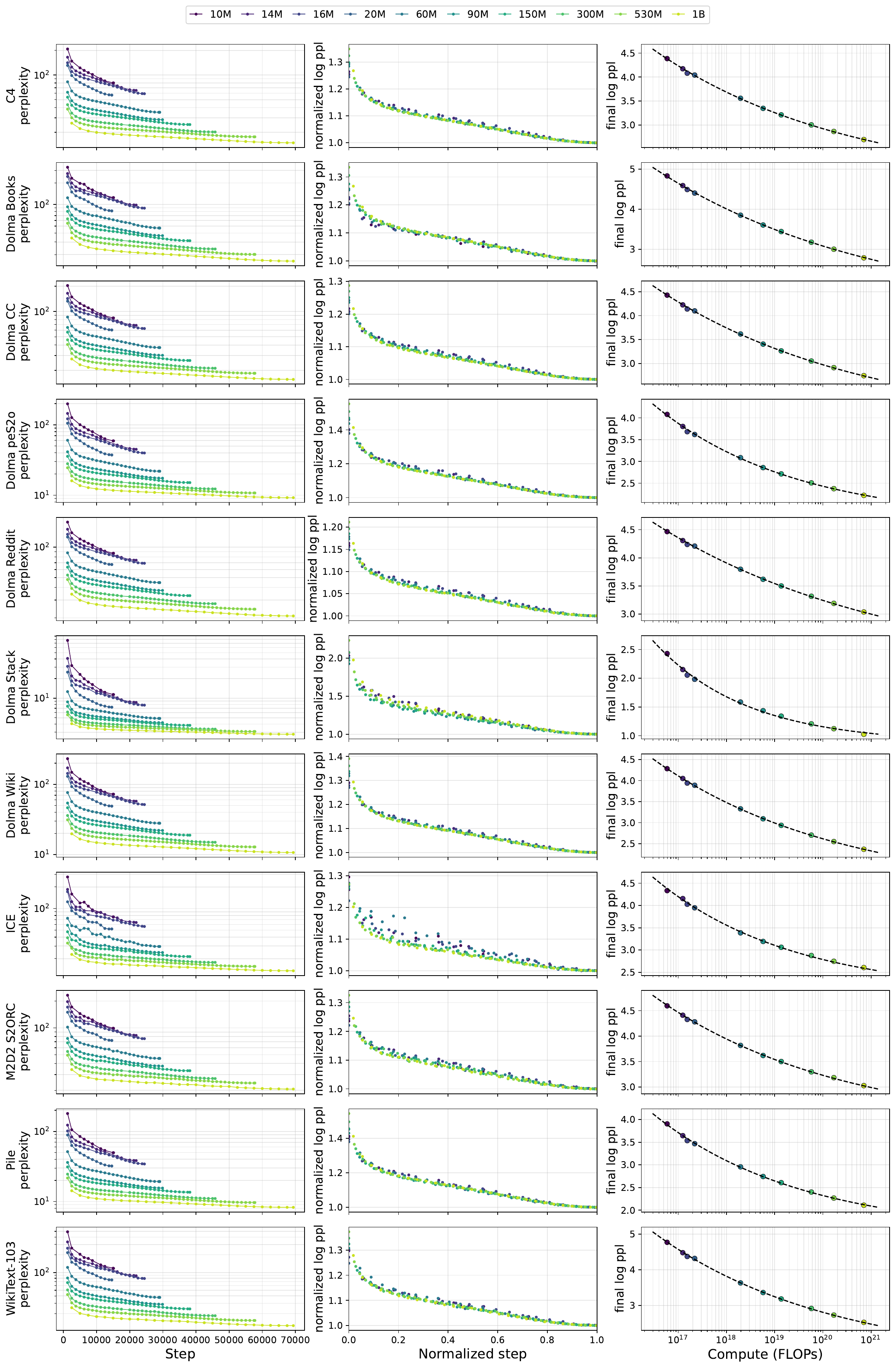}
    \caption{Universal scaling across 10 model sizes (10M--1B) trained on Dolma 1.7 (no Reddit) pretraining dataset, for perplexity on 11 validation sets (rows), as in \Cref{fig:datadecide ppl}.}
    \label{fig:datadecide ppl dolma1_7_no_reddit}
\end{figure}
\clearpage

\begin{figure}[H]
    \centering
    \includegraphics[width=0.9\linewidth]{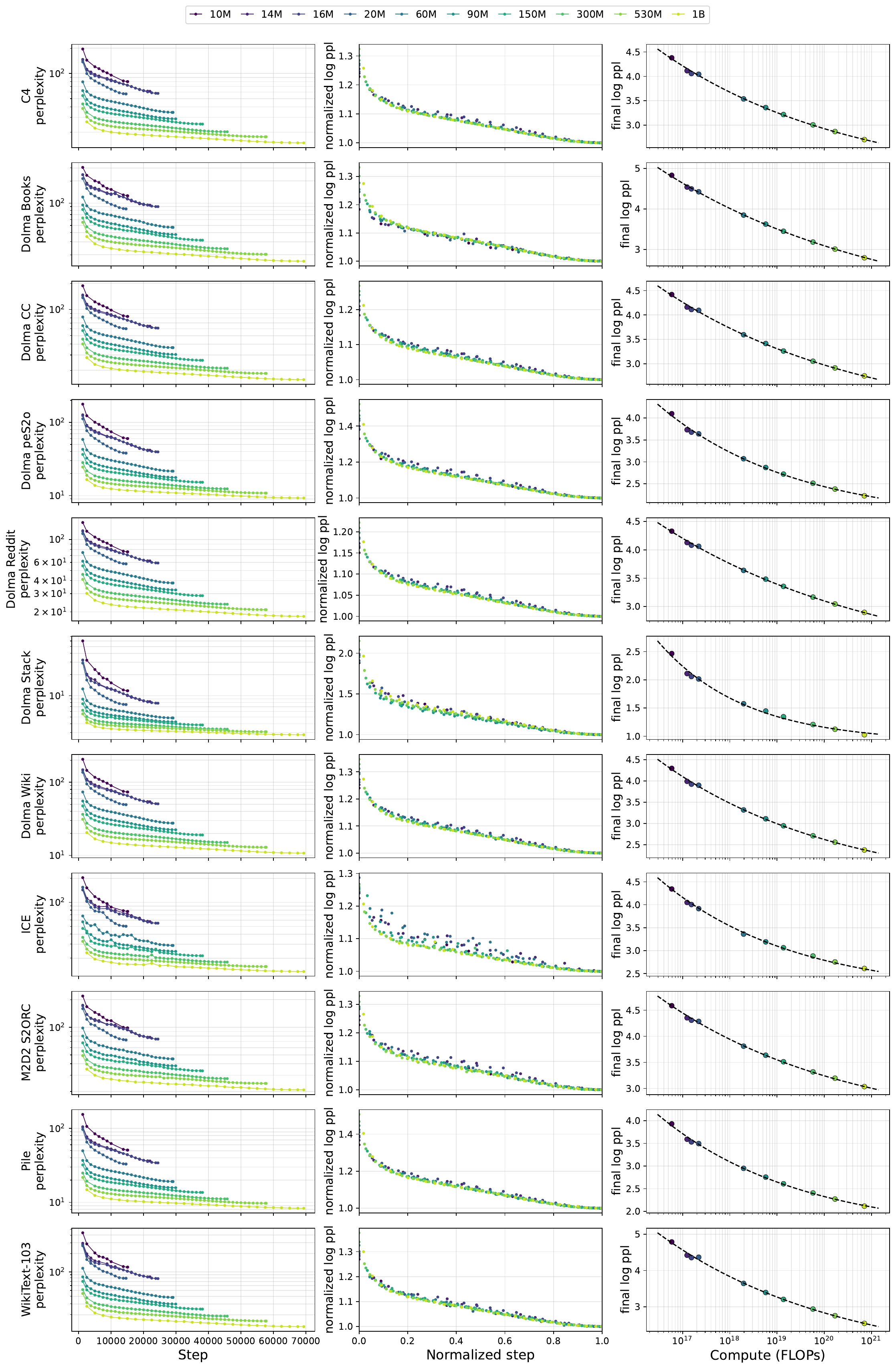}
    \caption{Universal scaling across 10 model sizes (10M--1B) trained on Dolma 1.7 (no Flan) pretraining dataset, for perplexity on 11 validation sets (rows), as in \Cref{fig:datadecide ppl}.}
    \label{fig:datadecide ppl dolma1_7_no_flan}
\end{figure}
\clearpage

\begin{figure}[H]
    \centering
    \includegraphics[width=0.9\linewidth]{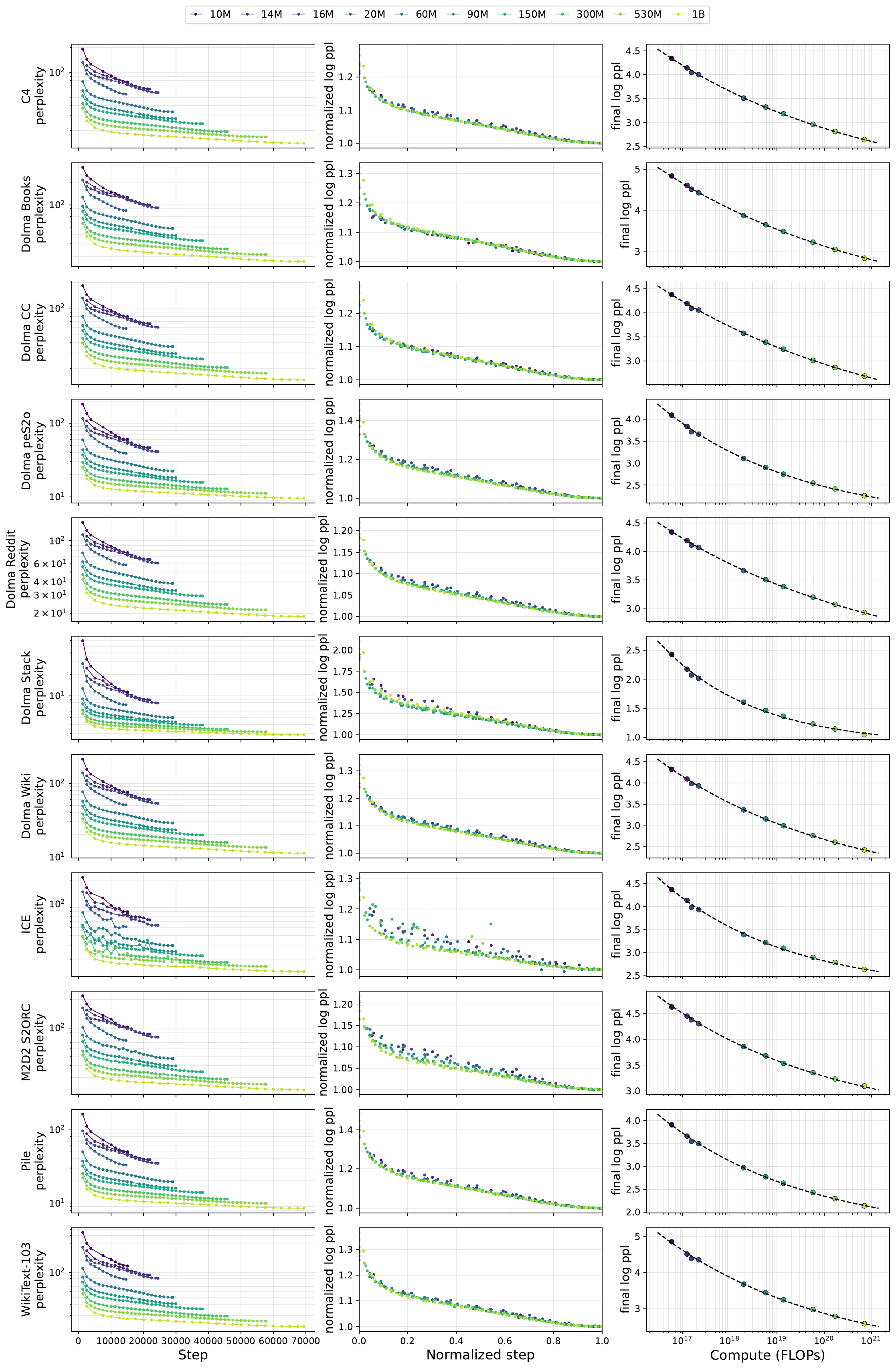}
    \caption{Universal scaling across 10 model sizes (10M--1B) trained on Dolma 1.6++ pretraining dataset, for perplexity on 11 validation sets (rows), as in \Cref{fig:datadecide ppl}.}
    \label{fig:datadecide ppl dolma1_6pp}
\end{figure}
\clearpage

\begin{figure}[H]
    \centering
    \includegraphics[width=0.9\linewidth]{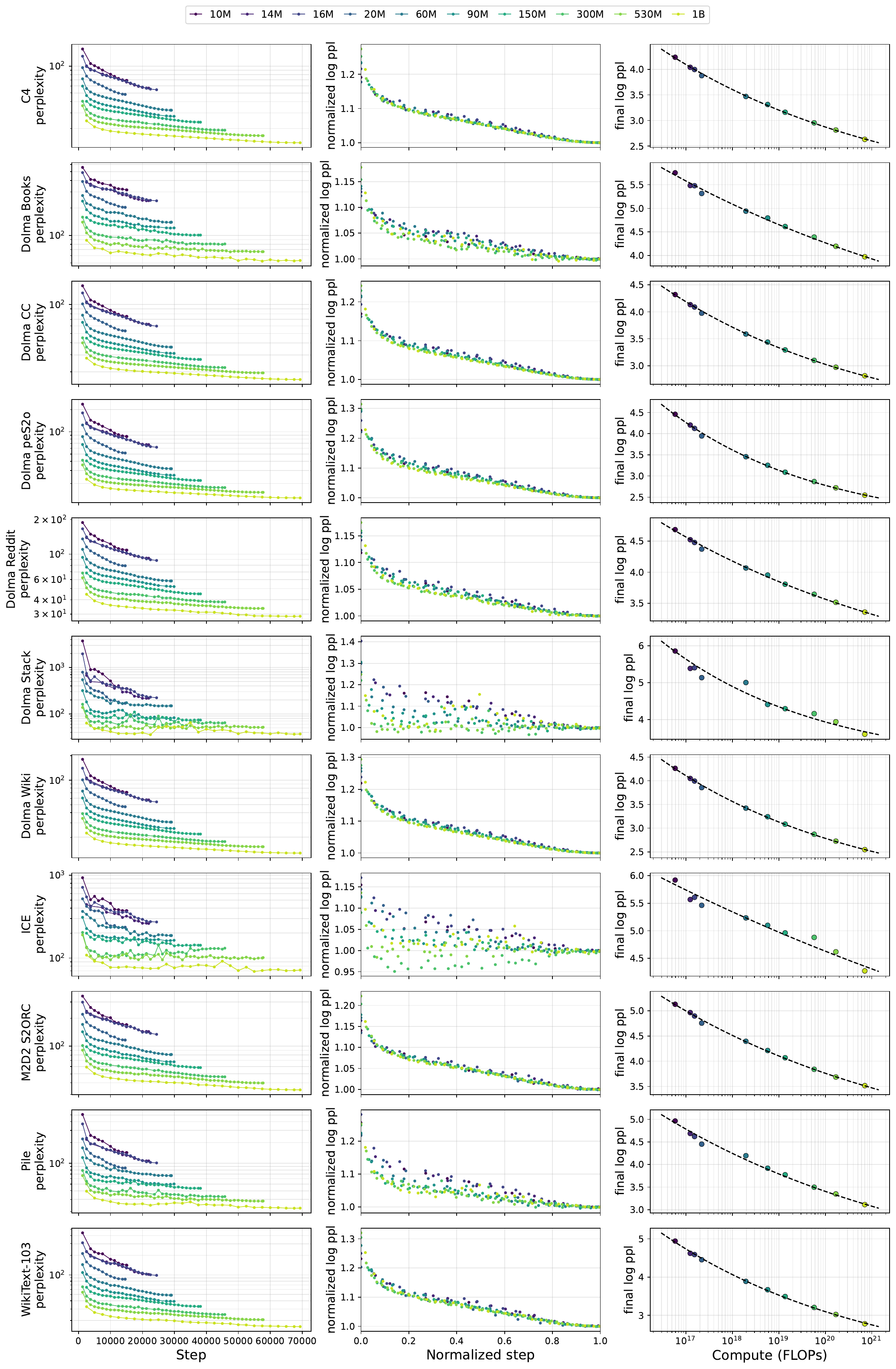}
    \caption{Universal scaling across 10 model sizes (10M--1B) trained on C4 pretraining dataset, for perplexity on 11 validation sets (rows), as in \Cref{fig:datadecide ppl}.}
    \label{fig:datadecide ppl c4}
\end{figure}
\clearpage

\begin{figure}[H]
    \centering
    \includegraphics[width=0.9\linewidth]{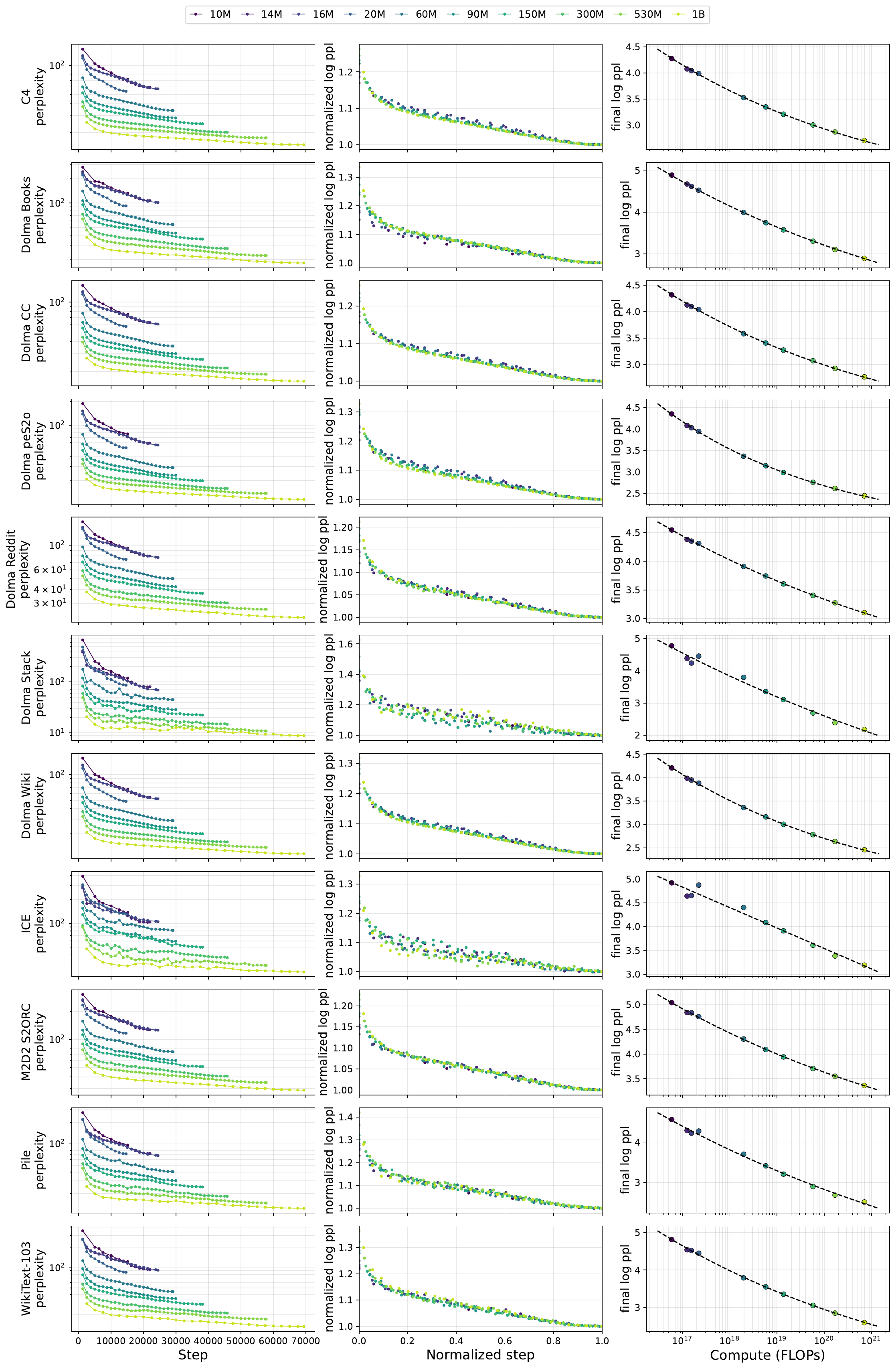}
    \caption{Universal scaling across 10 model sizes (10M--1B) trained on Falcon pretraining dataset, for perplexity on 11 validation sets (rows), as in \Cref{fig:datadecide ppl}.}
    \label{fig:datadecide ppl falcon}
\end{figure}
\clearpage

\begin{figure}[H]
    \centering
    \includegraphics[width=0.9\linewidth]{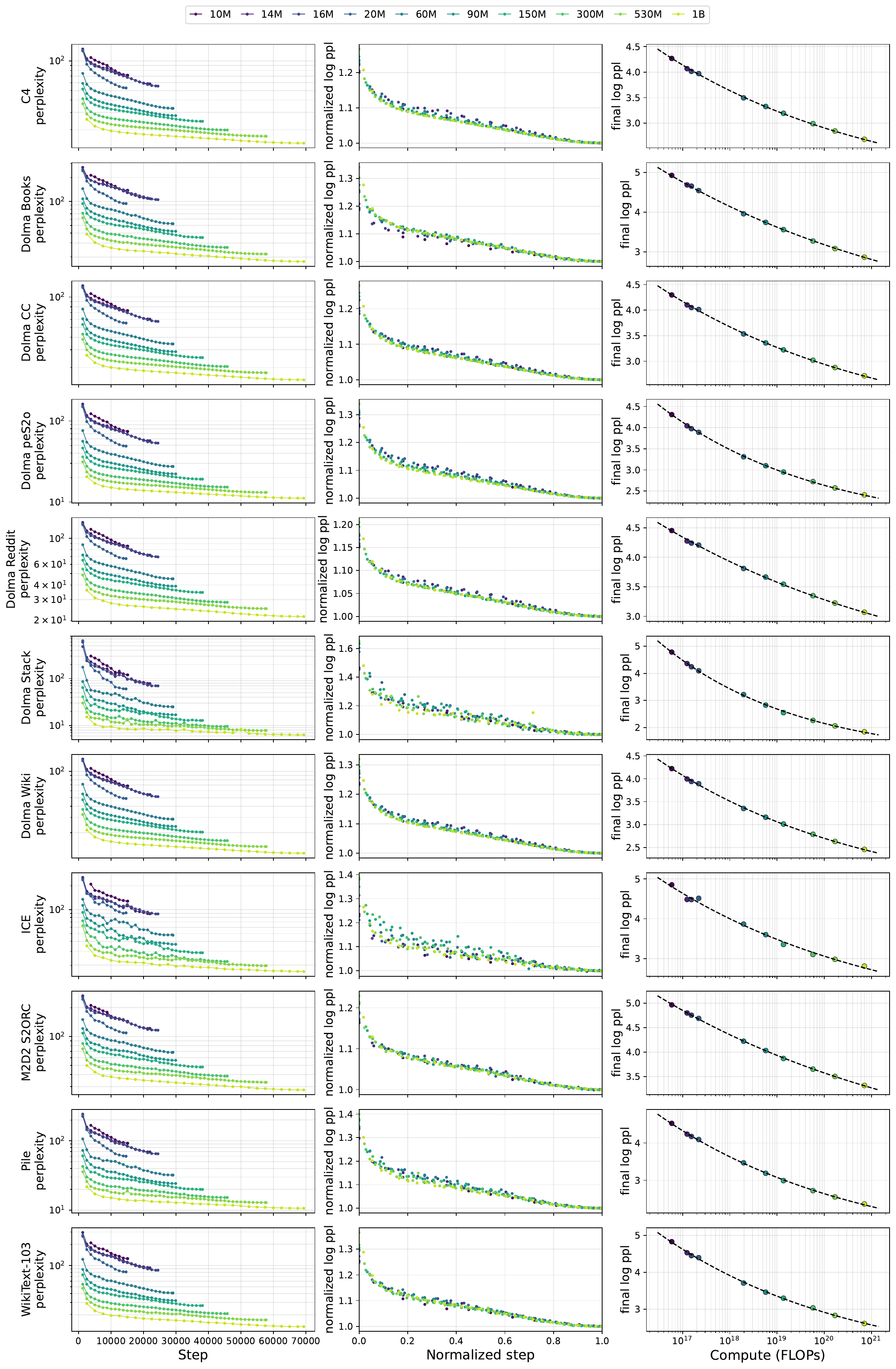}
    \caption{Universal scaling across 10 model sizes (10M--1B) trained on Falcon+CC pretraining dataset, for perplexity on 11 validation sets (rows), as in \Cref{fig:datadecide ppl}.}
    \label{fig:datadecide ppl falcon_cc}
\end{figure}
\clearpage

\begin{figure}[H]
    \centering
    \includegraphics[width=0.9\linewidth]{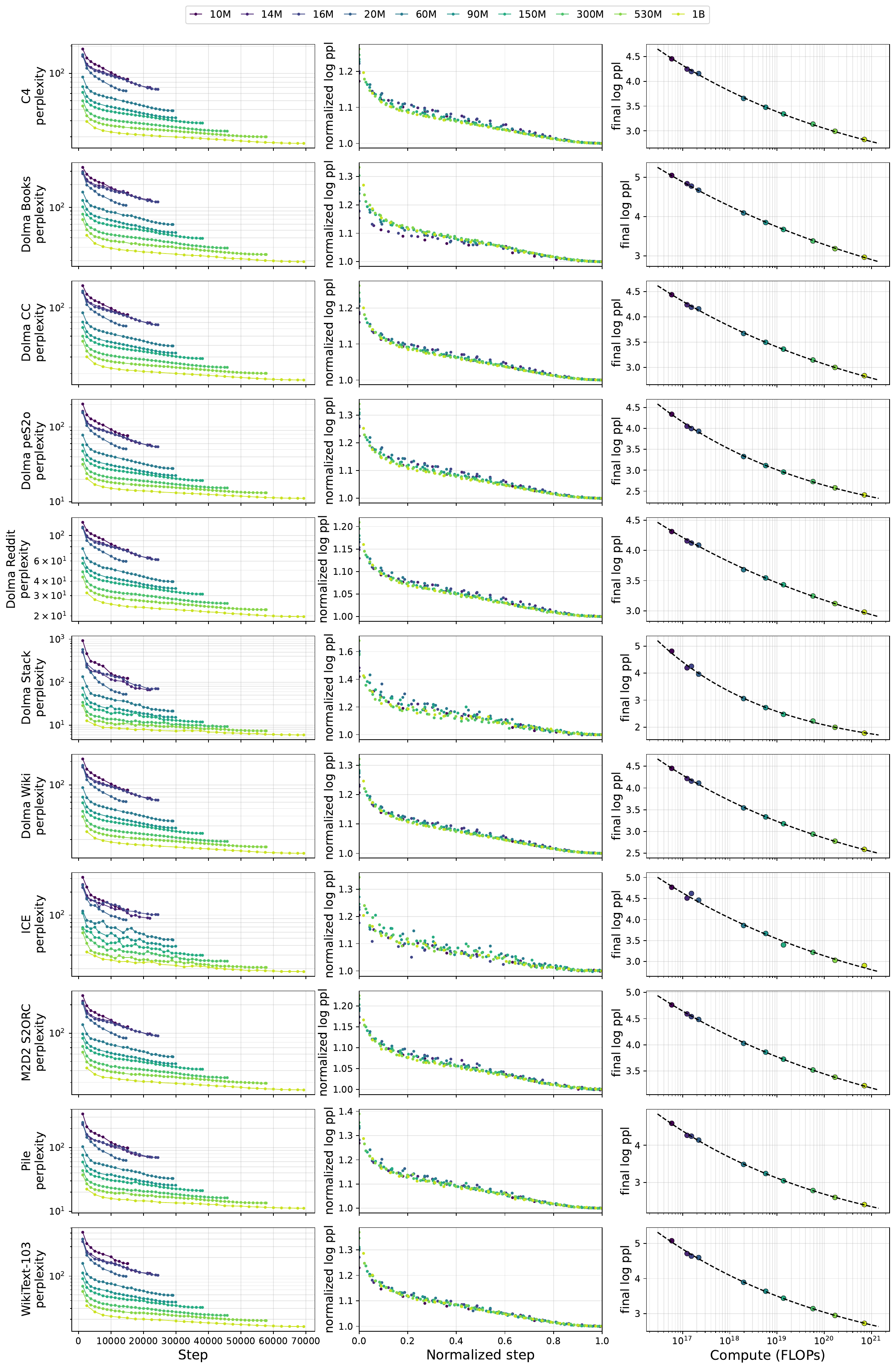}
    \caption{Universal scaling across 10 model sizes (10M--1B) trained on Falcon+CC (QC 10\%) pretraining dataset, for perplexity on 11 validation sets (rows), as in \Cref{fig:datadecide ppl}.}
    \label{fig:datadecide ppl falcon_cc_qc_10}
\end{figure}
\clearpage

\begin{figure}[H]
    \centering
    \includegraphics[width=0.9\linewidth]{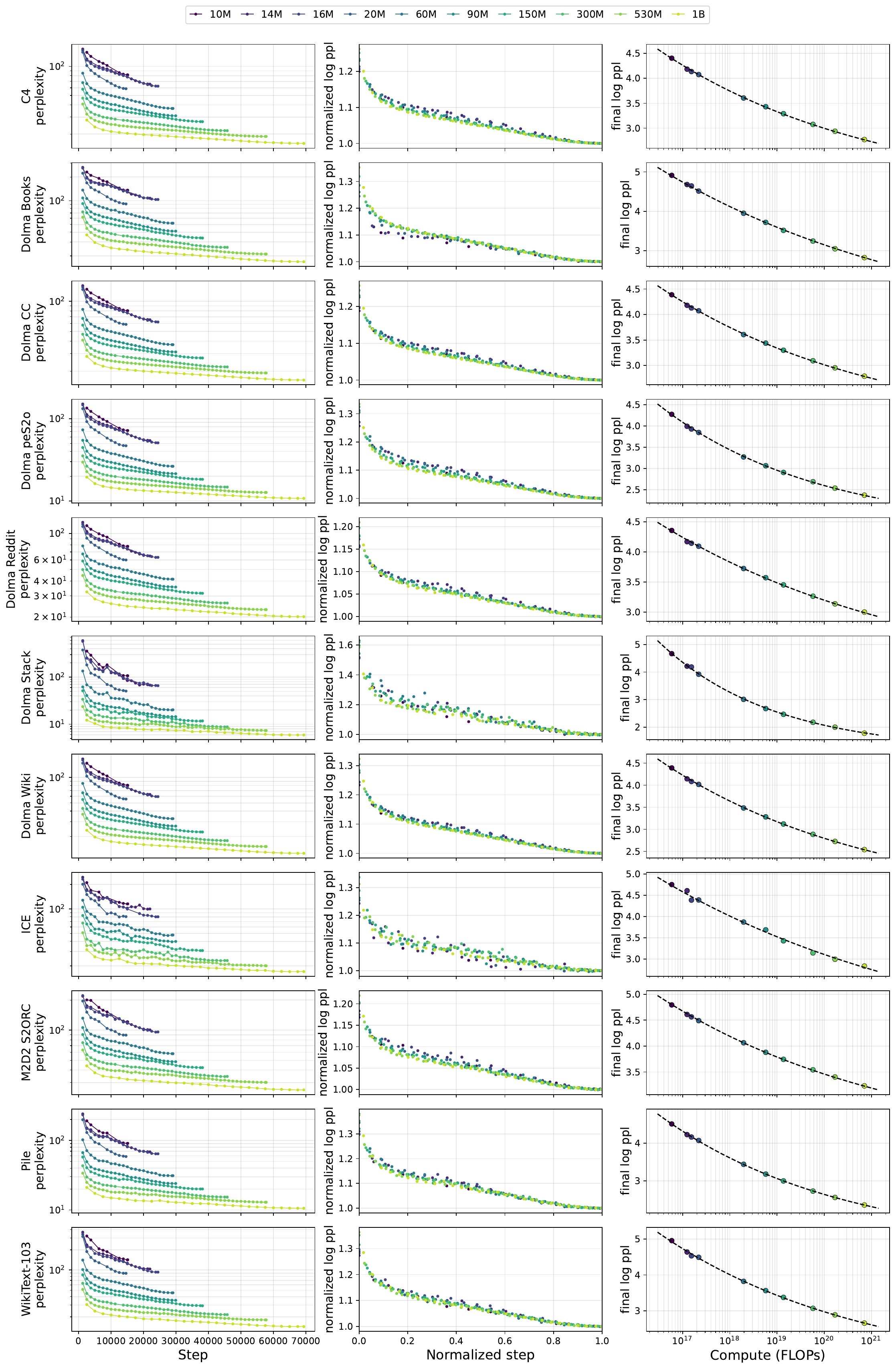}
    \caption{Universal scaling across 10 model sizes (10M--1B) trained on Falcon+CC (QC 20\%) pretraining dataset, for perplexity on 11 validation sets (rows), as in \Cref{fig:datadecide ppl}.}
    \label{fig:datadecide ppl falcon_cc_qc_20}
\end{figure}
\clearpage

\begin{figure}[H]
    \centering
    \includegraphics[width=0.9\linewidth]{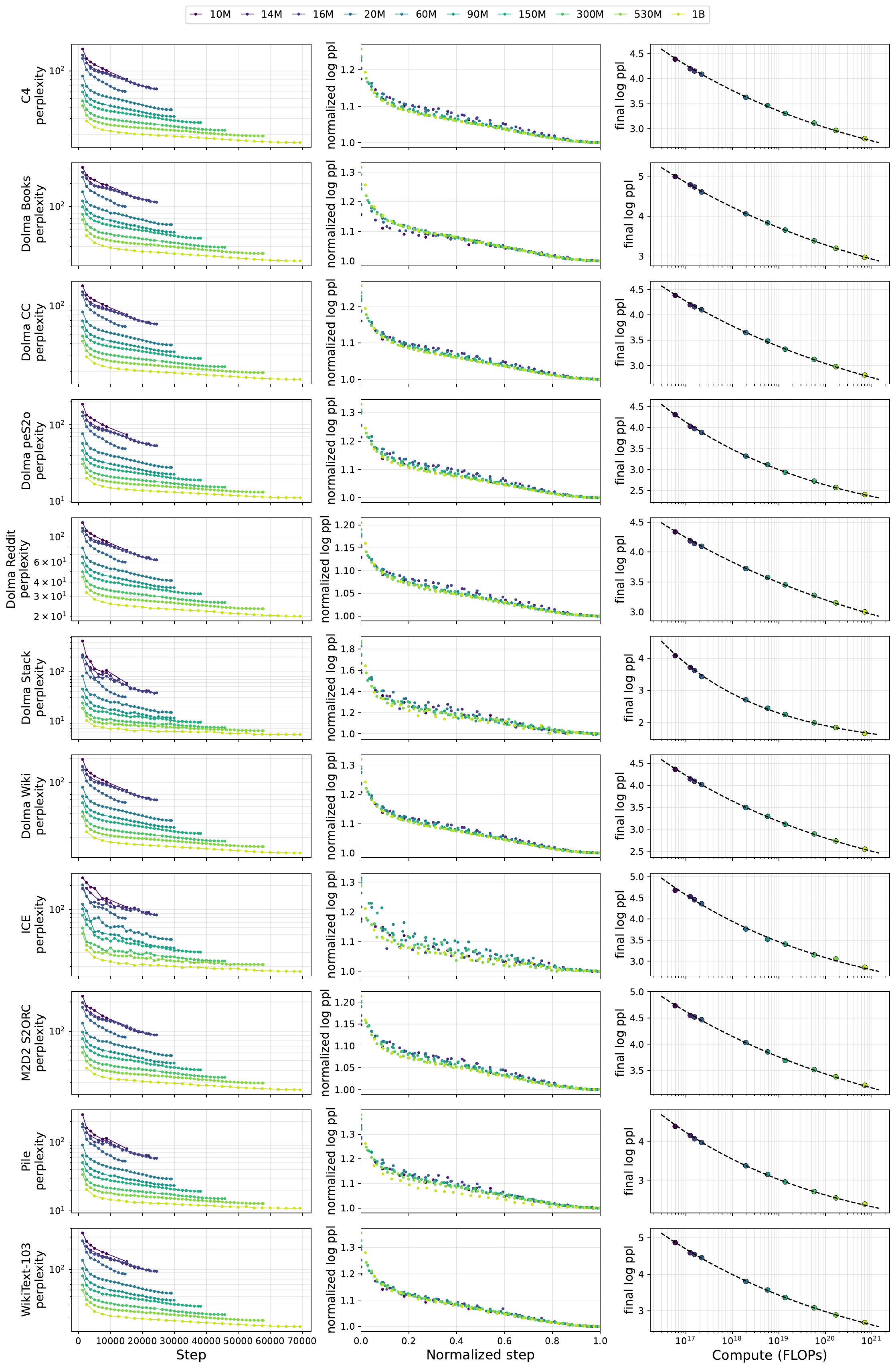}
    \caption{Universal scaling across 10 model sizes (10M--1B) trained on Falcon+CC (QC Orig 10\%) pretraining dataset, for perplexity on 11 validation sets (rows), as in \Cref{fig:datadecide ppl}.}
    \label{fig:datadecide ppl falcon_cc_qc_orig_10}
\end{figure}
\clearpage

\begin{figure}[H]
    \centering
    \includegraphics[width=0.9\linewidth]{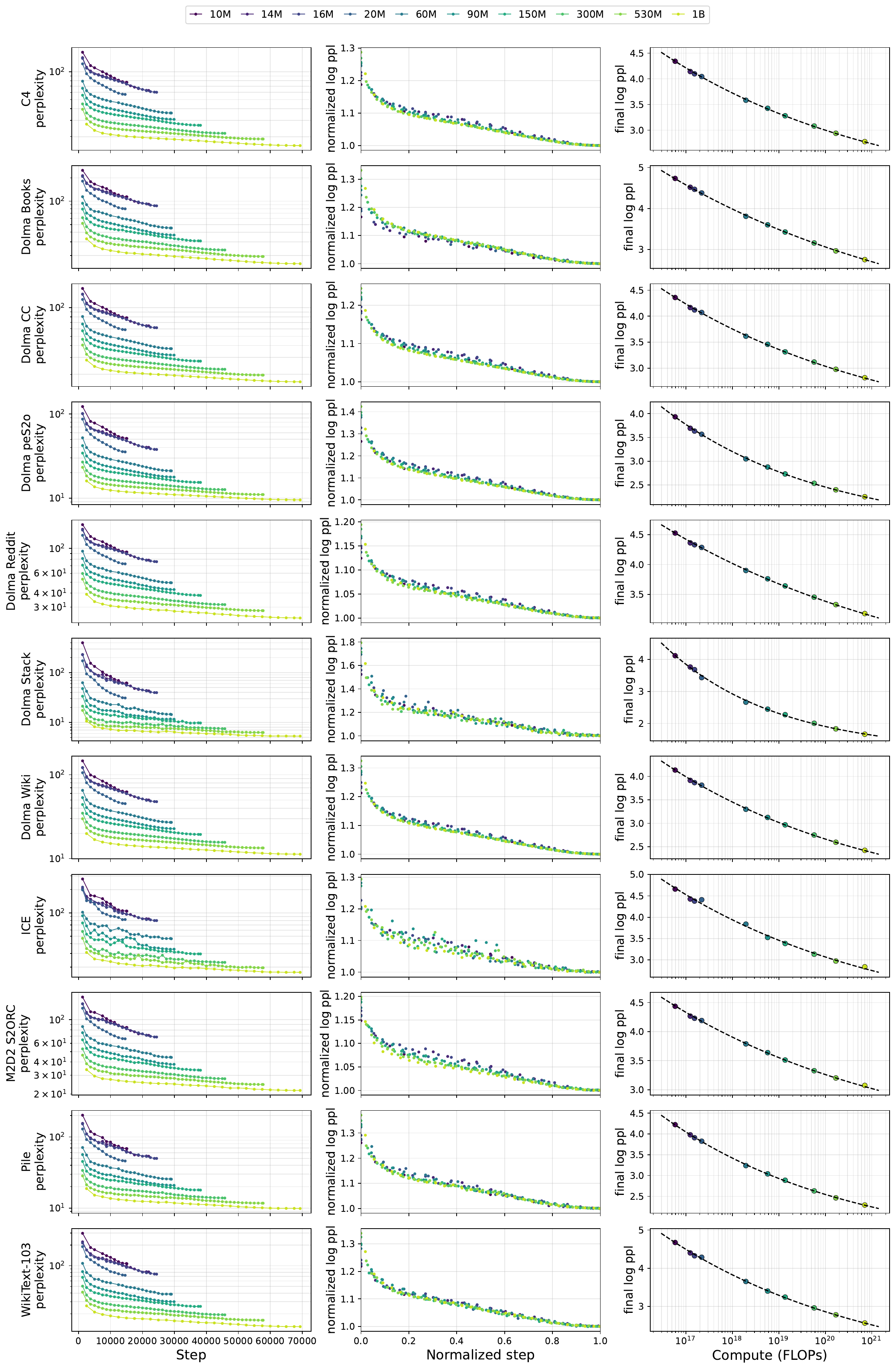}
    \caption{Universal scaling across 10 model sizes (10M--1B) trained on Falcon+CC (QC Tulu 10\%) pretraining dataset, for perplexity on 11 validation sets (rows), as in \Cref{fig:datadecide ppl}.}
    \label{fig:datadecide ppl falcon_cc_qc_tulu_10}
\end{figure}
\clearpage

\begin{figure}[H]
    \centering
    \includegraphics[width=0.9\linewidth]{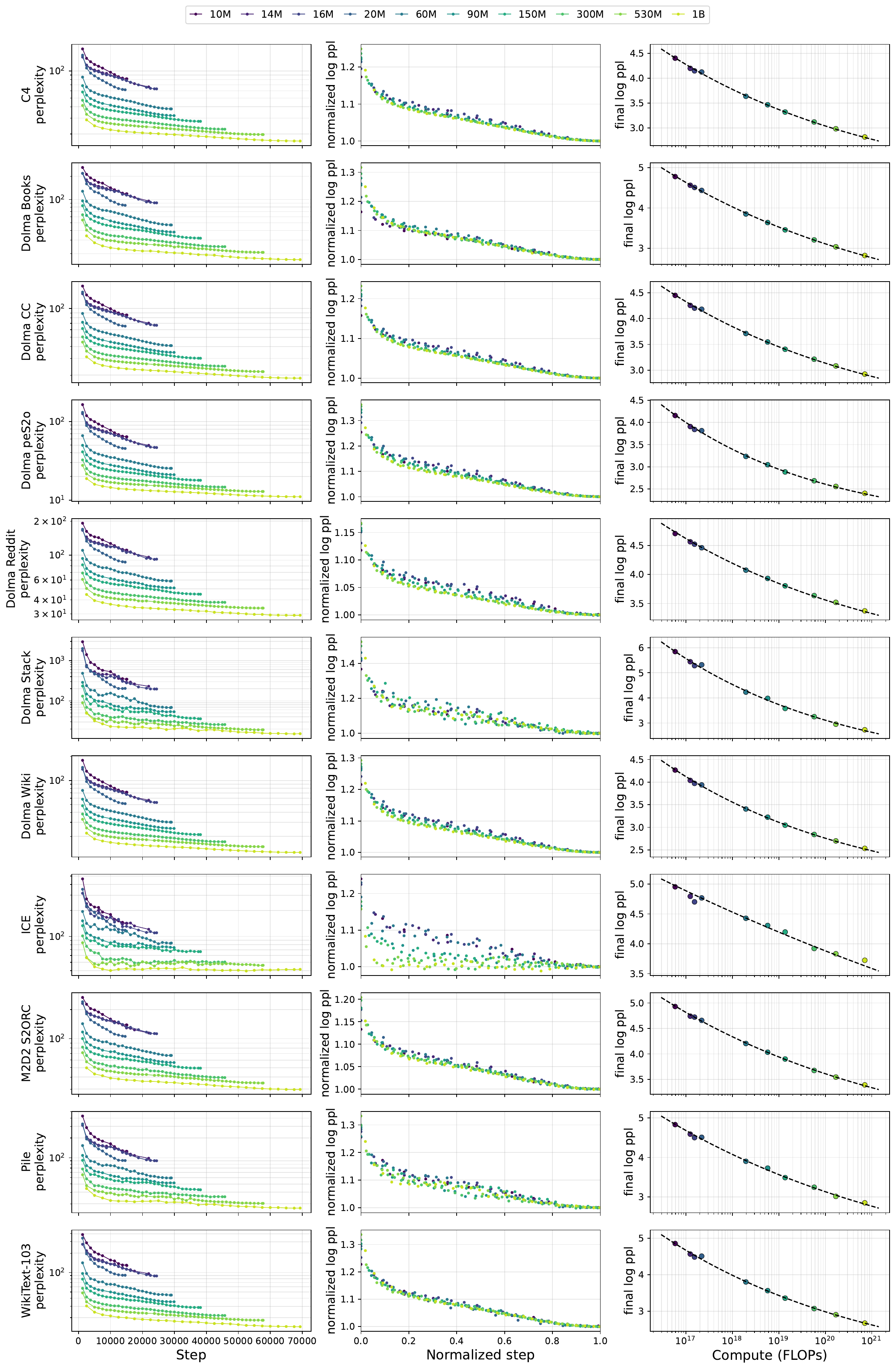}
    \caption{Universal scaling across 10 model sizes (10M--1B) trained on FineWeb-Pro pretraining dataset, for perplexity on 11 validation sets (rows), as in \Cref{fig:datadecide ppl}.}
    \label{fig:datadecide ppl fineweb_pro}
\end{figure}
\clearpage

\begin{figure}[H]
    \centering
    \includegraphics[width=0.9\linewidth]{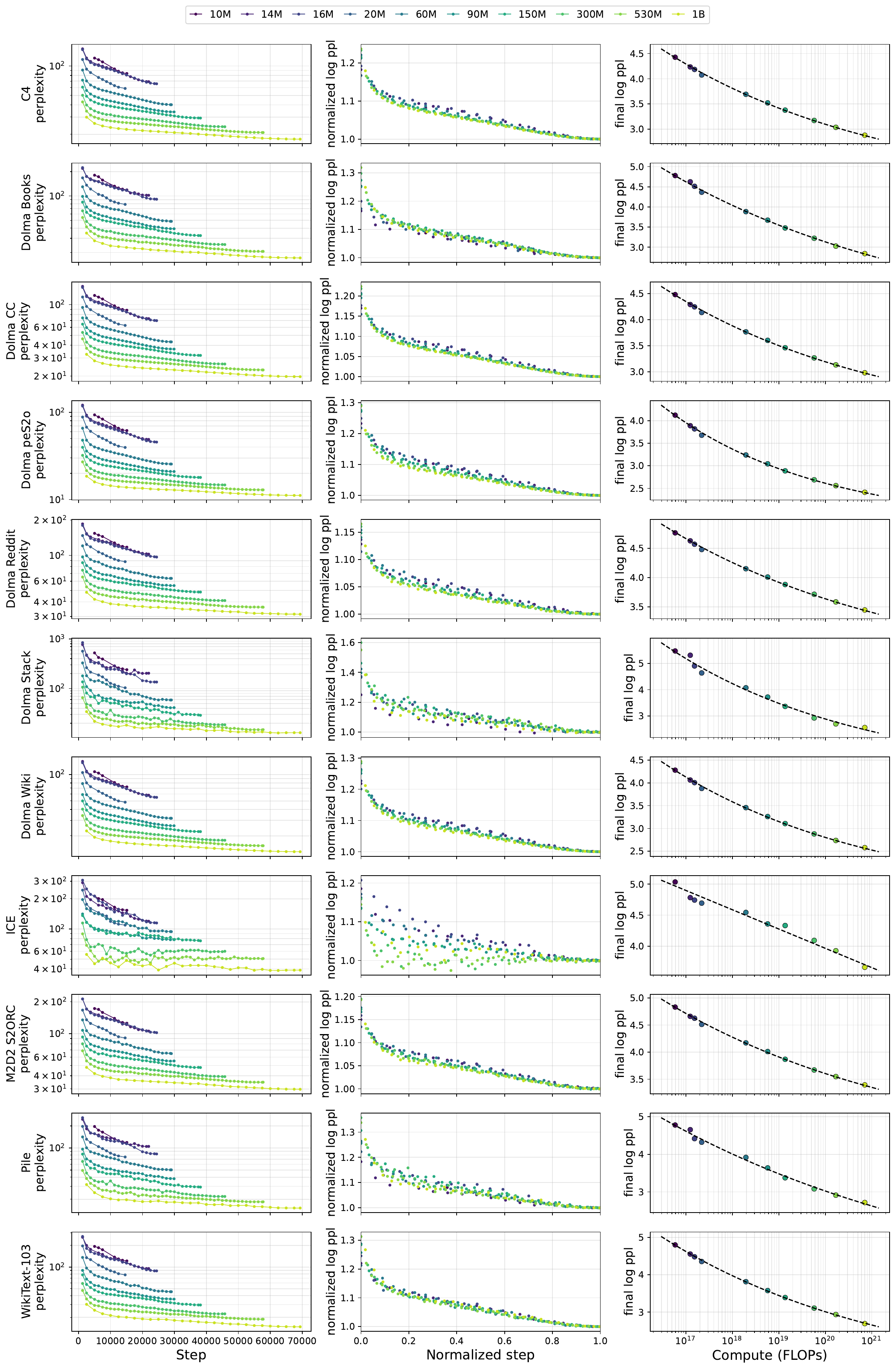}
    \caption{Universal scaling across 10 model sizes (10M--1B) trained on FineWeb-Edu pretraining dataset, for perplexity on 11 validation sets (rows), as in \Cref{fig:datadecide ppl}.}
    \label{fig:datadecide ppl fineweb_edu}
\end{figure}
\clearpage

\begin{figure}[H]
    \centering
    \includegraphics[width=0.9\linewidth]{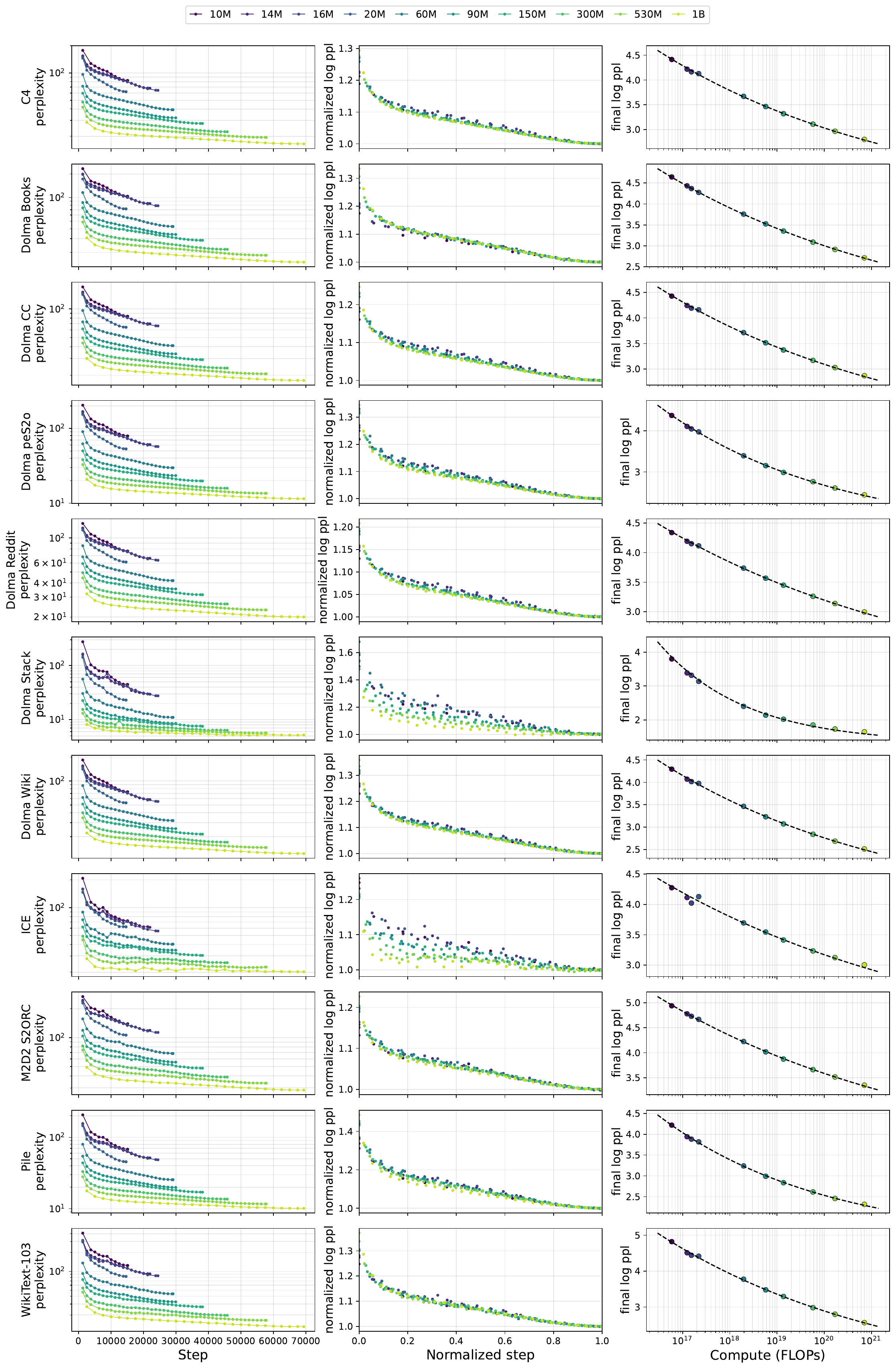}
    \caption{Universal scaling across 10 model sizes (10M--1B) trained on DCLM-Baseline pretraining dataset, for perplexity on 11 validation sets (rows), as in \Cref{fig:datadecide ppl}.}
    \label{fig:datadecide ppl dclm_baseline}
\end{figure}
\clearpage

\begin{figure}[H]
    \centering
    \includegraphics[width=0.9\linewidth]{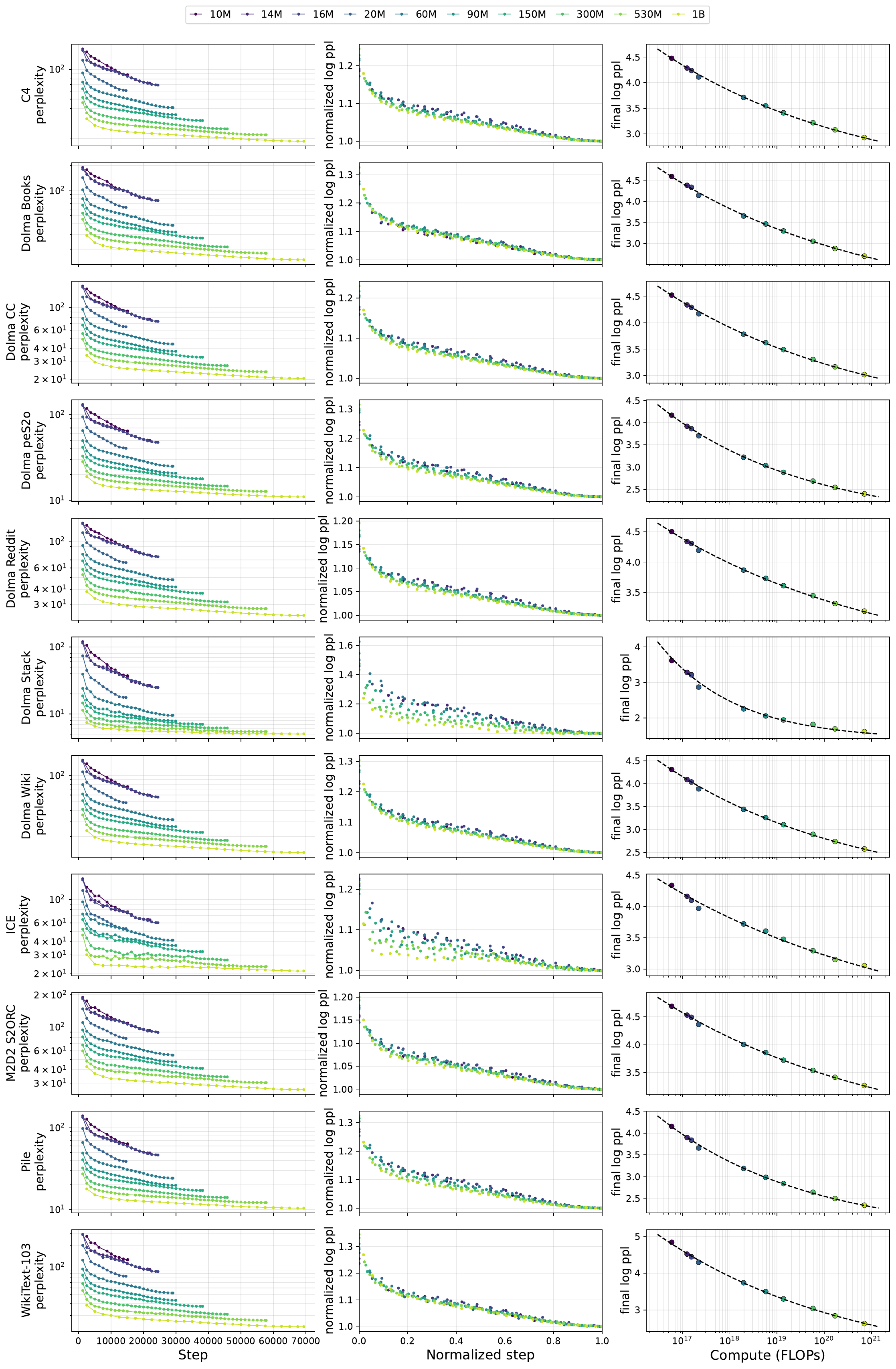}
    \caption{Universal scaling across 10 model sizes (10M--1B) trained on DCLM-Baseline (QC 7\%, FW2) pretraining dataset, for perplexity on 11 validation sets (rows), as in \Cref{fig:datadecide ppl}.}
    \label{fig:datadecide ppl dclm_baseline_qc_7_fw2}
\end{figure}
\clearpage

\begin{figure}[H]
    \centering
    \includegraphics[width=0.9\linewidth]{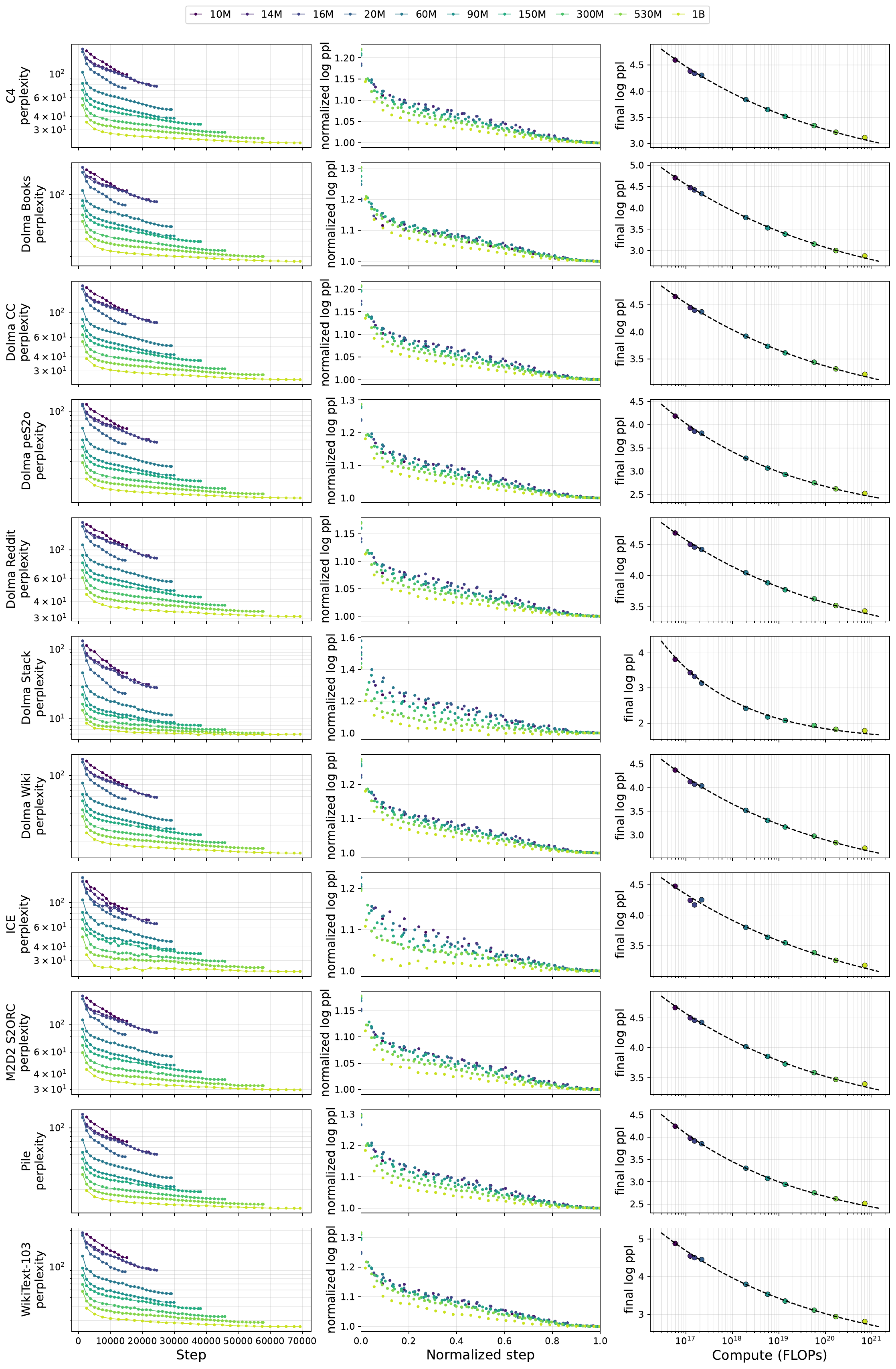}
    \caption{Universal scaling across 10 model sizes (10M--1B) trained on DCLM-Baseline (QC 7\%, FW3) pretraining dataset, for perplexity on 11 validation sets (rows), as in \Cref{fig:datadecide ppl}.}
    \label{fig:datadecide ppl dclm_baseline_qc_7_fw3}
\end{figure}
\clearpage

\begin{figure}[H]
    \centering
    \includegraphics[width=0.9\linewidth]{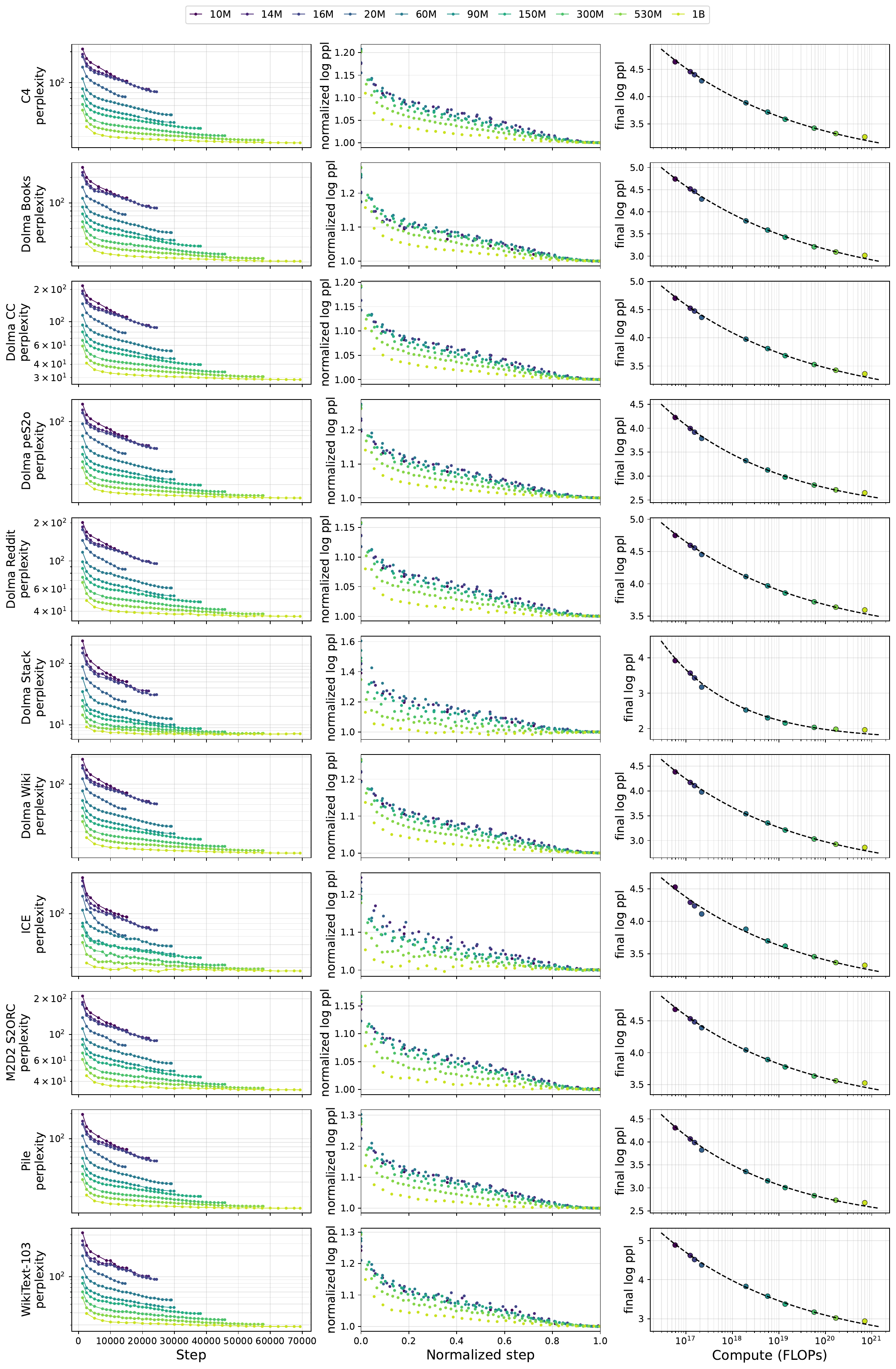}
    \caption{Universal scaling across 10 model sizes (10M--1B) trained on DCLM-Baseline (QC FW 3\%) pretraining dataset, for perplexity on 11 validation sets (rows), as in \Cref{fig:datadecide ppl}.}
    \label{fig:datadecide ppl dclm_baseline_qc_fw_3}
\end{figure}
\clearpage

\begin{figure}[H]
    \centering
    \includegraphics[width=0.9\linewidth]{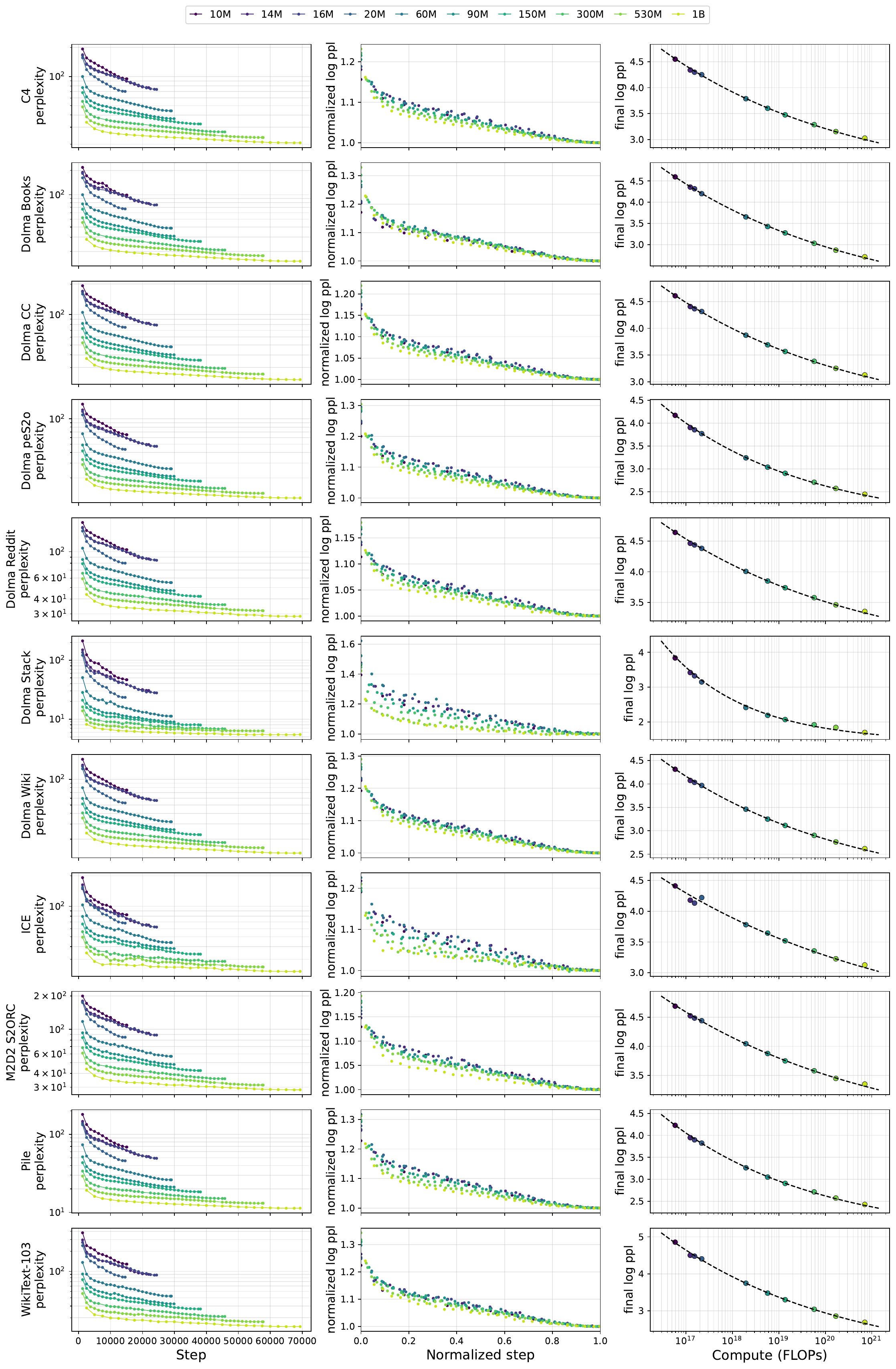}
    \caption{Universal scaling across 10 model sizes (10M--1B) trained on DCLM-Baseline (QC FW 10\%) pretraining dataset, for perplexity on 11 validation sets (rows), as in \Cref{fig:datadecide ppl}.}
    \label{fig:datadecide ppl dclm_baseline_qc_fw_10}
\end{figure}
\clearpage

\begin{figure}[H]
    \centering
    \includegraphics[width=0.9\linewidth]{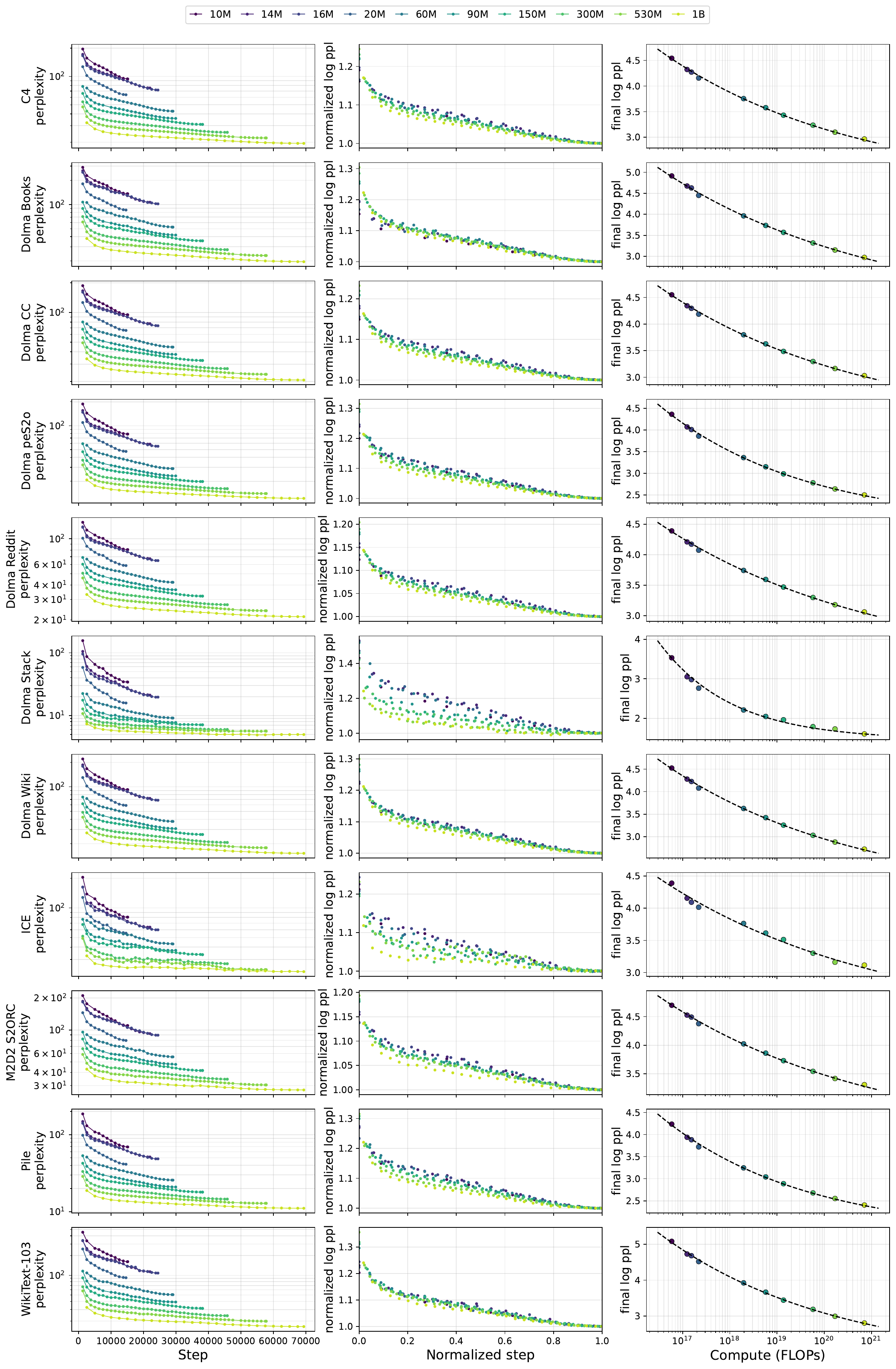}
    \caption{Universal scaling across 10 model sizes (10M--1B) trained on DCLM-Baseline (QC 10\%) pretraining dataset, for perplexity on 11 validation sets (rows), as in \Cref{fig:datadecide ppl}.}
    \label{fig:datadecide ppl dclm_baseline_qc_10}
\end{figure}
\clearpage

\begin{figure}[H]
    \centering
    \includegraphics[width=0.9\linewidth]{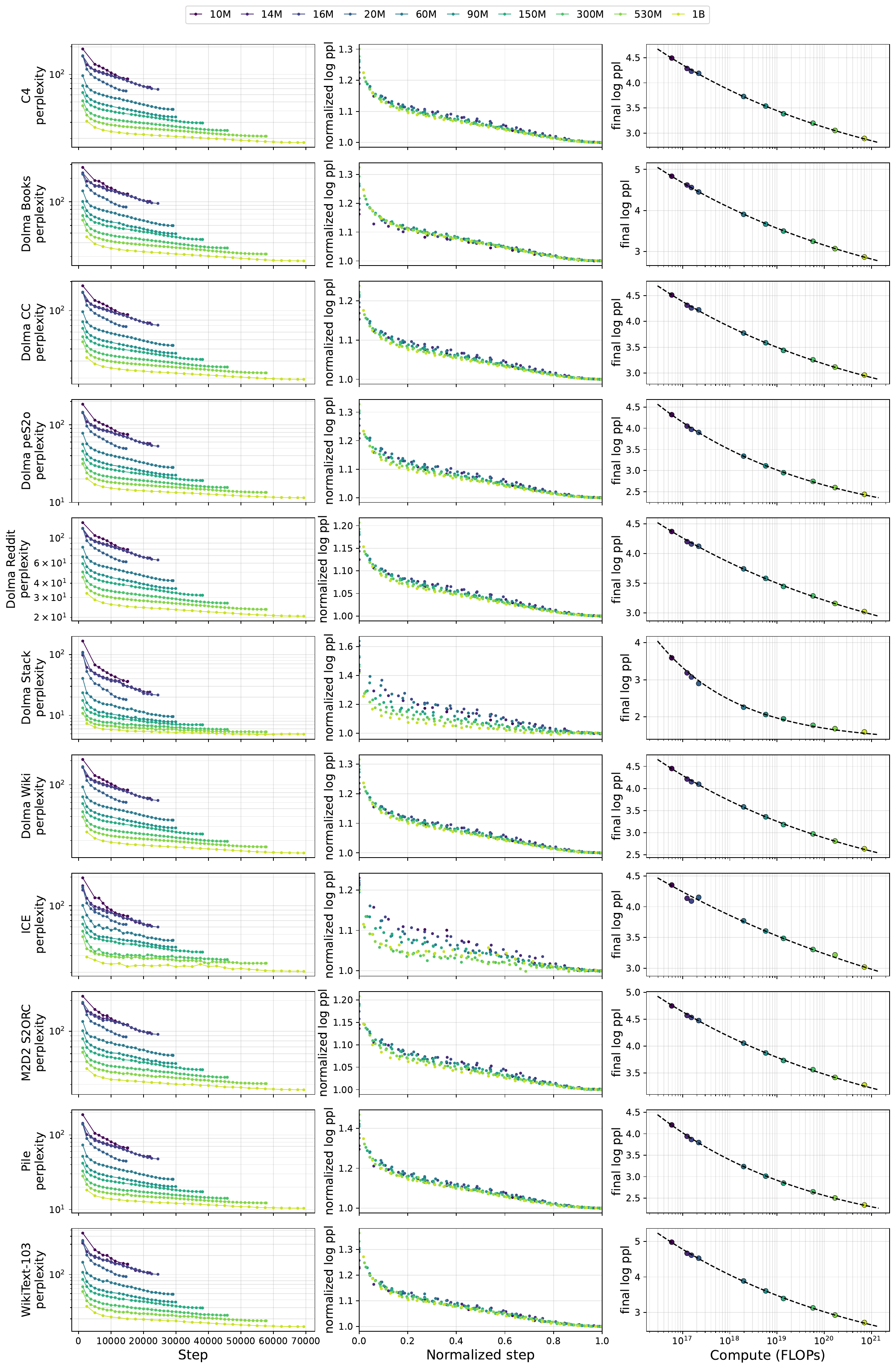}
    \caption{Universal scaling across 10 model sizes (10M--1B) trained on DCLM-Baseline (QC 20\%) pretraining dataset, for perplexity on 11 validation sets (rows), as in \Cref{fig:datadecide ppl}.}
    \label{fig:datadecide ppl dclm_baseline_qc_20}
\end{figure}
\clearpage

\begin{figure}[H]
    \centering
    \includegraphics[width=0.9\linewidth]{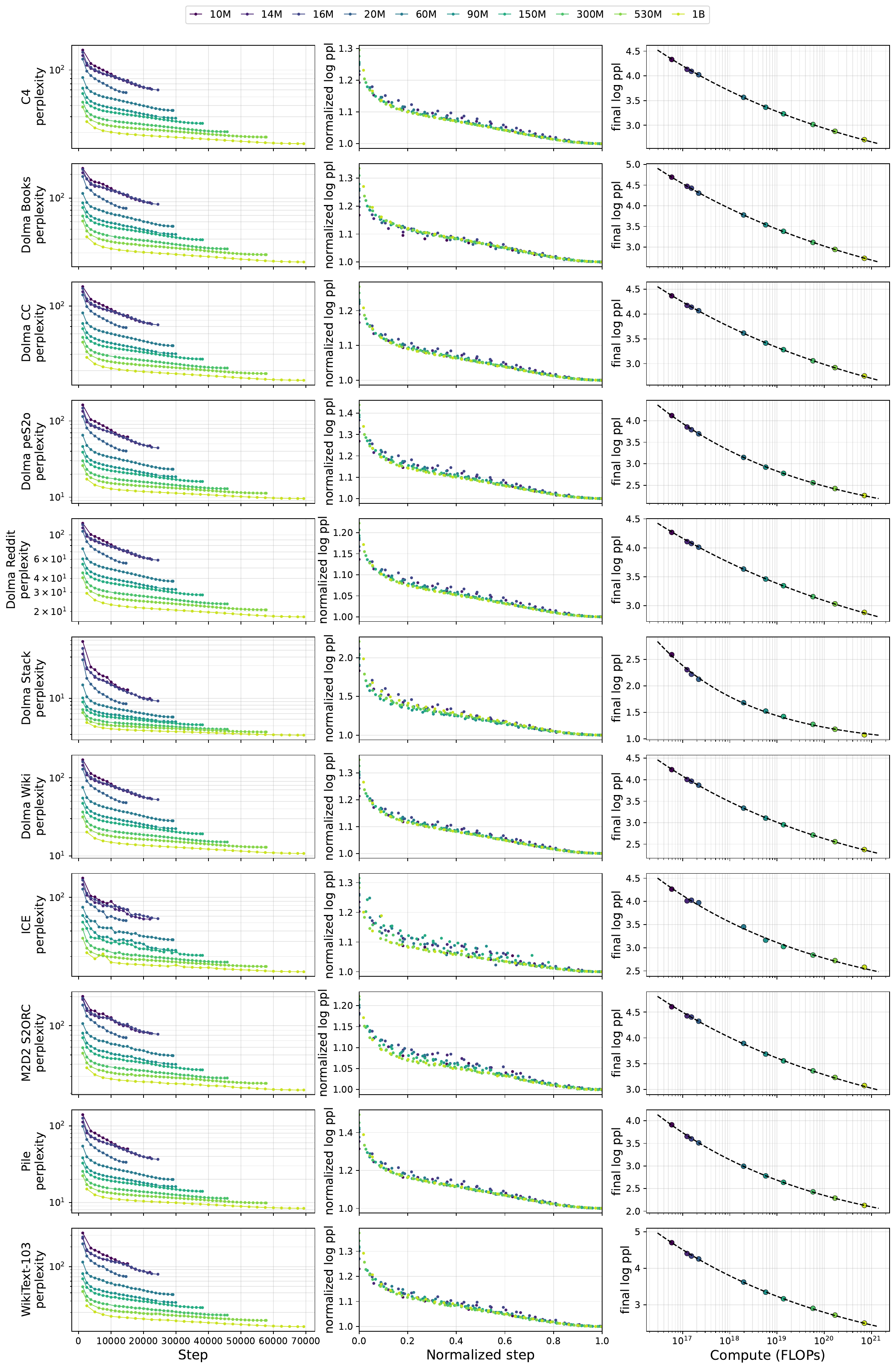}
    \caption{Universal scaling across 10 model sizes (10M--1B) trained on DCLM-Baseline 25\% / Dolma 75\% pretraining dataset, for perplexity on 11 validation sets (rows), as in \Cref{fig:datadecide ppl}.}
    \label{fig:datadecide ppl dclm_baseline_25_dolma_75}
\end{figure}
\clearpage

\begin{figure}[H]
    \centering
    \includegraphics[width=0.9\linewidth]{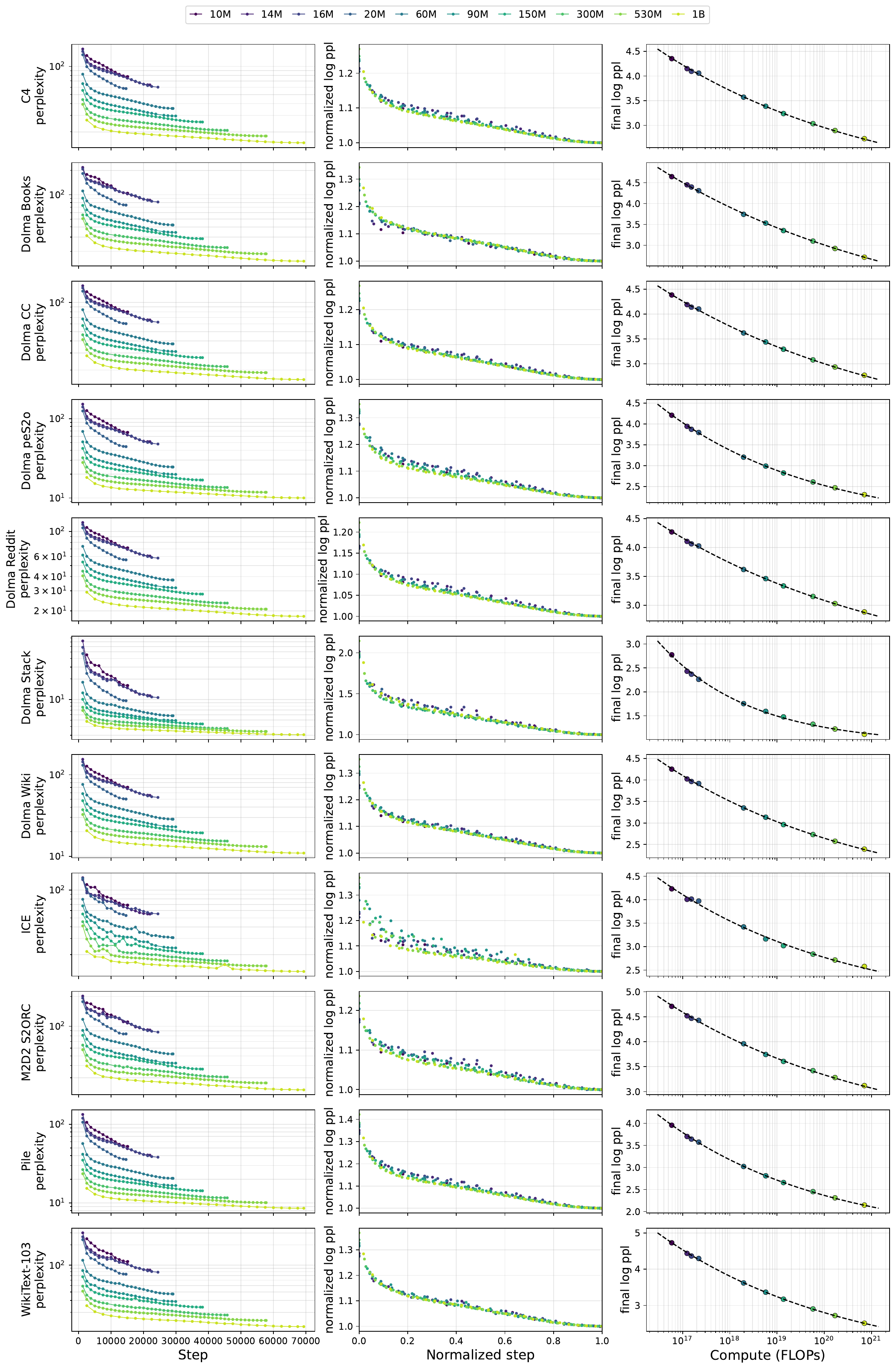}
    \caption{Universal scaling across 10 model sizes (10M--1B) trained on DCLM-Baseline 50\% / Dolma 50\% pretraining dataset, for perplexity on 11 validation sets (rows), as in \Cref{fig:datadecide ppl}.}
    \label{fig:datadecide ppl dclm_baseline_50_dolma_50}
\end{figure}
\clearpage

\begin{figure}[H]
    \centering
    \includegraphics[width=0.9\linewidth]{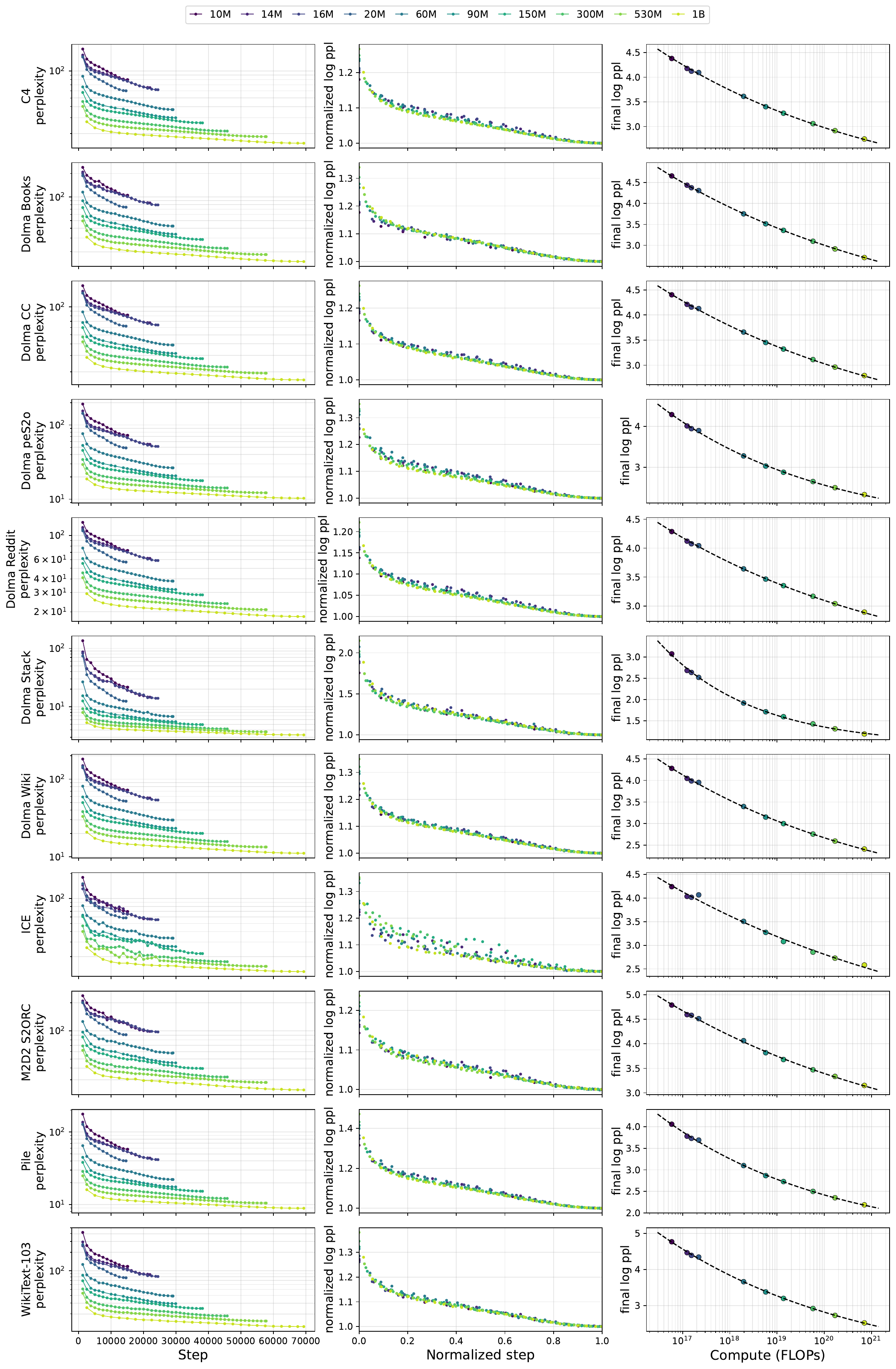}
    \caption{Universal scaling across 10 model sizes (10M--1B) trained on DCLM-Baseline 75\% / Dolma 25\% pretraining dataset, for perplexity on 11 validation sets (rows), as in \Cref{fig:datadecide ppl}.}
    \label{fig:datadecide ppl dclm_baseline_75_dolma_25}
\end{figure}
\clearpage

\end{document}